\documentclass[times,twocolumn,final]{elsarticle}

\usepackage{styles/journalpreprint}
\usepackage{framed,multirow}

\usepackage{amssymb}
\usepackage{latexsym}

\usepackage{url}
\usepackage{xcolor}

\usepackage[breaklinks=true]{hyperref}

\usepackage{graphicx}
\usepackage{epsfig}
\usepackage{amsmath}
\usepackage{enumitem}
\usepackage{booktabs}

\usepackage{bm}
\usepackage{upgreek}
\usepackage{algorithmic}
\usepackage[vlined,ruled]{algorithm2e}
\usepackage{soul,xcolor}
\usepackage[bottom]{footmisc}
\usepackage{mathtools}

\definecolor{newcolor}{rgb}{.8,.349,.1}

\journal{Author Preprint}

\DeclareMathOperator*{\argmin}{argmin}
\DeclareMathOperator*{\argmax}{argmax}
\DeclareMathOperator{\logit}{logit}

\newcommand{\ra}[1]{\renewcommand{\arraystretch}{#1}}

\newcommand{\R}{\mathbb{R}}

\newcommand{\calA}{\mathcal{A}}
\newcommand{\calB}{\mathcal{B}}

\newcommand{\calD}{\mathcal{D}}
\newcommand{\calE}{\mathcal{E}}

\newcommand{\calL}{\mathcal{L}}

\newcommand{\calN}{\mathcal{N}}

\newcommand{\calR}{\mathcal{R}}

\newcommand{\calX}{\mathcal{X}}
\newcommand{\calY}{\mathcal{Y}}

\newcommand{\rmA}{\mathrm{A}}

\newcommand{\eg}{\textit{e.g.}\xspace}
\newcommand{\ie}{\textit{i.e.}\xspace}

\newcommand{\eq}{\!=\!}
\newcommand{\ssim}{\!\sim\!}
\newcommand{\ccoloneqq}{\!\coloneqq\!}
\newcommand{\ttriangleq}{\!\triangleq\!}

\newcommand{\pperp}{\perp\!\!\!\perp}

\allowdisplaybreaks

\begin{document}

\verso{Lo\"ic {Le Folgoc} \textit{et~al.}}

\begin{frontmatter}

\title{Bayesian analysis of the prevalence bias: learning and predicting from imbalanced 
data}

\author[1]{Lo\"ic \snm{Le Folgoc}\corref{cor1}}
\cortext[cor1]{Corresponding author: Department of Computing, Huxley Building, 180 Queen's Gate, London SW7 2RH, United Kingdom}
\ead{l.le-folgoc@imperial.ac.uk}
\author[2]{Vasileios \snm{Baltatzis}}
\author[1]{Amir \snm{Alansary}}
\author[1,3]{Sujal \snm{Desai}}
\author[3]{Anand \snm{Devaraj}}
\author[3]{Sam \snm{Ellis}}
\author[2]{Octavio E. \snm{Martinez Manzanera}}
\author[1]{Fahdi \snm{Kanavati}}
\author[4]{Arjun \snm{Nair}}
\author[2]{Julia \snm{Schnabel}}
\author[1]{Ben \snm{Glocker}}

\address[1]{BioMedIA, Imperial College London, United Kingdom}
\address[2]{Biomedical Engineering and Imaging Sciences, King's College London, UK}
\address[3]{The Royal Brompton \& Harefield NHS Foundation Trust, London, UK}
\address[4]{Department of Radiology, University College London, UK}

\availableonline{July 2021}

\begin{abstract}
Datasets are rarely a realistic approximation of the target population. Say, prevalence is misrepresented, image quality is above clinical standards, etc. This mismatch is known as sampling bias. Sampling biases are a major hindrance for machine learning models. They cause significant gaps between model performance in the lab and in the real world.
Our work is a solution to prevalence bias. 
Prevalence bias is the discrepancy between the prevalence of a pathology 
and its sampling rate in the training dataset, introduced upon collecting data or due to the practioner rebalancing the training batches.
This paper lays the theoretical and computational framework for training models, and for prediction, in the presence of prevalence bias. 
Concretely a bias-corrected loss function, as well as bias-corrected predictive rules, are derived under the principles of Bayesian risk minimization. The loss exhibits a direct connection to the information gain. 
It offers a principled alternative to heuristic training losses and complements test-time procedures based on selecting an operating point from summary curves.
It integrates seamlessly in the current paradigm of (deep) learning using stochastic backpropagation and naturally with Bayesian models.

\end{abstract}


\end{frontmatter}

\section{Introduction}

\begin{figure*}[t]
\includegraphics[width=\textwidth]{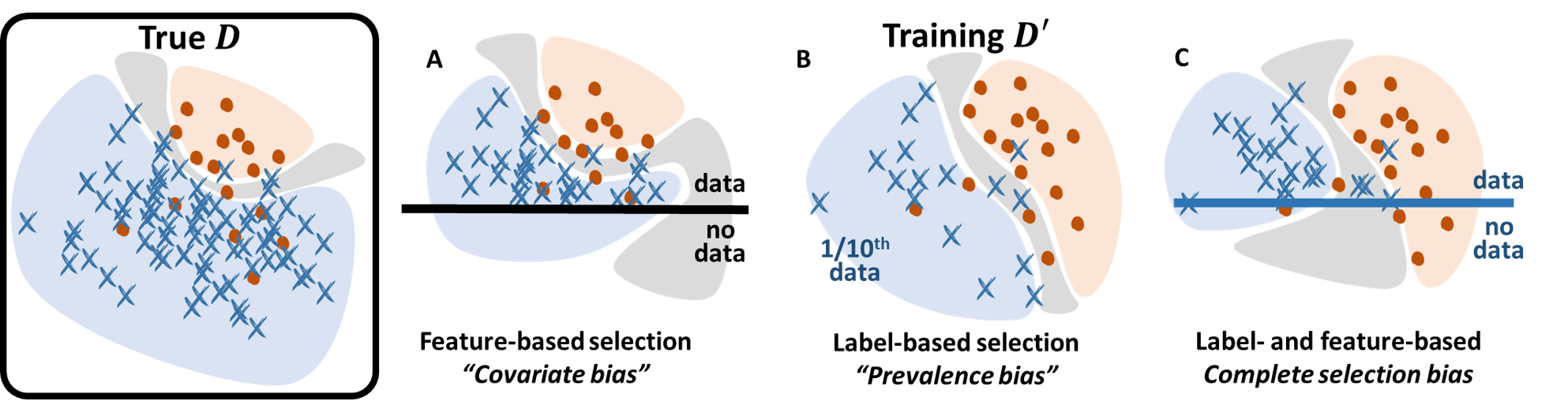}
\caption{Sampling bias types. Three ways in which the true 
population $\calD$ can differ from the training distribution $\calD'$ of the dataset $X,Y$; and its effect on a model trained from such data. Positive samples are shown as red dots and negatives as blue crosses. Regions confidently assigned to either label have matching blue/red colours, while uncertain regions are greyed. (a) Feature-based sampling \eg, no data collected for children or elderly people; (b) Label-based sampling \eg, dataset evenly split between positives and negatives with no regard for the true prevalence. (c) General case: selection mechanism unknown, or based both on outcome and features \eg, volunteer/control data (unlike patient data) may be unavailable for subsets of the population.} \label{fig: sampling biases}
\end{figure*}

\subsection{Motivation}

We consider supervised machine learning tasks, in which a model is trained from data $X\eq\{x_n\}_{n=1}^N,Y\eq\{y_n\}_{n=1}^N$ to predict a dependent variable $y\!\in\!\calY$ (\eg the classification label) from inputs $x\!\in\!\calX$ (a.k.a., \textit{covariates} or features), optimally for a target population $x_\ast,y_\ast\!\sim\!\calD$~\cite{hastie2009elements}. It is often assumed that the training data $X,Y$ is a representative sample from this target population. The present work explores a situation where that basic assumption is violated.

It is helpful to recognize that a model optimal for say, mostly healthy controls, or for some distribution of phenotypes, may no longer perform well for a pathological population or for another distribution of phenotypes. 
The predictive power depends on the statistics of the population $\calD$~\cite{powers2011evaluation}. For instance the predictive value of a diagnostic test is not intrinsic to the test: it depends on the prevalence of the disease~\cite{altman1994statistics}. 
Sampling bias (or sample selection bias) refers to discrepancies between the distribution $\calD'$ of the training dataset and the true distribution $\calD$. 

Machine learning models are subject to sampling bias. It affects the quality of predictions \eg, the classification accuracy for the population $\calD$, and the validity of statistical findings \eg, the strength of association between exposure and outcome. Section~\ref{sec: example} illustrates in concrete terms the significance of prevalence bias and its pitfalls for statistical inference and prediction. 

Sampling bias is 
pervasive in the medical imaging literature. 
An overwhelming body of work relies on data from observational 
studies. The training set $X,Y\sim\calD'$ is the result of a 
potentially less controlled process than for experimental studies. There are numerous sampling protocols for data collection (\eg random, stratified, clustered, subjective)~\cite{panacek2007sampling,hulley2007choosing,etikan2017sampling}, with data possibly aggregated from composite sources.
Morever the current machine learning (ML) paradigm, with its reliance on large data availability, encourages to repurpose retrospective data. This may be done in a way that mismatches the original study, or unaware of inclusion/exclusion criteria specific to that study. 
An automated screening model may be trained from incidental data or from purpose-made data collected in specialized units. Incidence rates would differ between these populations and the general population. 
Statistics may further be biased by the acquisition site \eg, by country, hospital; and by practical choices. Say, clinical partners may handcraft a balanced dataset with equal amounts of healthy and pathological cases; relying on their expertise to judge the value and usefulness of a sample (\textit{subjective} sampling) \eg, discarding trivial or ambiguous cases, or based on quality control criteria (\eg, image quality).

At the other end the dataset is often adjusted by the ML practitioner upon training models. 
Sampling heuristics are generally introduced with performance w.r.t.~set quantitative benchmarks in mind, disregarding population statistics. Of course, such performance gains may not transfer to the real world.

\subsection{Related work}

Heckman~\cite{heckman1979sample} provides in Nobel Prize winning econometrics work a comprehensive discussion of, and methods for analyzing selective samples. The typology is adopted in sociology~\cite{berk1983introduction}, machine learning~\cite{zadrozny2004learning,cortes2008sample} and for statistical tests in genomics~\cite{young2010gene} and medical communities~\cite{stukel2007analysis}. Selection biases are discussed from the broader scope of structural biases in sociology~\cite{winship1999estimation} and epidemiology~\cite{hernan2004structural}. 
A naive yet useful dichotomy from a practical machine learning viewpoint is to ask whether the selection mechanism conditions on covariates $x$ or outcome variables $y$ (Fig.~\ref{fig: sampling biases}). The worst cases are when the mechanism underlying the bias is unknown or conditions both on covariates and outcome. Early work in this setting is for bias correction in linear regression models with fully parametric or semi-nonparametric selection models~\cite{vella1998estimating}. We focus instead on a well-posed practical scenario of \textit{prevalence} bias, but allowing for arbitrary nonlinear relashionships between $x$ and $y$. 

\setlength{\parskip}{0pt}
Of course much insight into model identifiability, transportability of results, and whether correct inference and prediction are possible at all, can be gained from a more thorough structural characterization of selection biases~\cite{geneletti2009adjusting,bareinboim2012controlling,cooper2013bayesian}. The present work contributes to bridge a computational gap in this literature when dealing with large non-linear models.  The applicability to arbitrary model architectures (\eg deep learning) poses additional computational challenges that motivate the key technical contributions of the paper. 

\setlength{\parskip}{0pt}
This paper focuses on prevalence bias, or \textit{label}-based sampling bias (Fig.~\ref{fig: sampling biases}(b)), as in~\cite{elkan2001foundations,lin2002support}, but unlike~\cite{shimodaira2000improving,zadrozny2004learning,huang2007correcting,cortes2014domain} who address \textit{feature}-based sampling bias, also known as \textit{covariate} shift (Fig.~\ref{fig: sampling biases}(a)).
The proposed approach is derived from Bayesian principles through which prevalence bias, unlike covariate shift, necessitates a different probabilistic treatment than the bias-free case. 
Concretely in a fully Bayesian treatment, undersampling parts of the input space $\calX$ mostly results in higher uncertainty, whereas undersampling a specific label invalidates the (probabilistic) decision boundary.

\setlength{\parskip}{0pt}
Much of the machine learning literature~\cite{shimodaira2000improving,elkan2001foundations,lin2002support,zadrozny2004learning,huang2007correcting} adopts a strategy of importance weighting, whereby the cost of training sample errors is weighted to more
closely reflect that of the test distribution. 
Importance weighting is rooted in a frequentist analysis and regularized risk minimization~\cite{huang2007correcting}, that is maximizing the expected log-likelihood $\mathbb{E}_\calD[\log{p(y|x,w)}]$ plus a regularizer $-\lambda\calR(w)\!\equiv\!\log{p(w)}$, w.r.t. model parameters $w$. The present analysis departs from importance weighting. 
In section~\ref{sec: bias-corrected posterior} it leads instead to a modified training likelihood, which we refer to as the Bayesian Information Gain by analogy to the information-theoretic concept. Information gain and mutual information have found much use in a variety of medical imaging tasks~\cite{wells1996multi,zikic2014encoding,lefolgoc2016lifted} as well as in ML and deep learning~\cite{chen2016infogan,belghazi2018mutual,hjelm2019learning}, but to our knowledge have not appeared in the context of Bayesian posteriors and prevalence bias.

\setlength{\parskip}{0pt}
Related work also appears in the literature on transfer learning~\cite{pan2009survey,weiss2016survey,shin2016deep} and domain adaptation~\cite{cortes2014domain,kamnitsas2017unsupervised,frid2018gan} driven by NLP, speech and image processing applications. The aim is to cope with generally ill-posed shifts of the distribution of the input $x$. In that sense the present paper is orthogonal to, and can be combined with this body of work. Besides the problem of class imbalance is central in medical image segmentation where a class (\eg, the background) is often over-represented in the dataset. It brings about a number of resampling (class rebalancing) strategies, see for instance a discussion of their effect on various metrics in~\cite{kamnitsas2017efficient}, as well as a review, benchmark and informative look into various empirical corrections in~\cite{li2019overfitting}.  

\setlength{\parskip}{0pt}
When 
dealing with miscalibrated probabilistic models, we often see the cut-off threshold for predicting a given label as the work-around to fix prediction performance. The search for a suitable operating point can be formalized via sensitivity-specificity plots, a.k.a.~ROC curves~\cite{fawcett2006introduction,akobeng2007understanding,cook2007use}. 
The resulting prediction is no longer probabilistic. 
Instead we explicitly account for prevalence 
in inference and prediction to derive optimal bias-corrected probabilistic decision rules. 
ROC curves are nonetheless useful as a prevalence-agnostic summary, in the post-hoc analysis of predictive performance.

Finally, the (log-)odds ratio~\cite{bland2000odds, szumilas2010explaining}, commonly used in case-control studies, is relevant as a prevalence-agnostic measure of association between exposure and outcome. We show that models trained under the proposed methodology place higher probability in parts of the parameter space that lead to odd ratios consistent with the empirical data.

\subsection{Contributions}

Section~\ref{sec: Bayesian IG in practice} presents the main result of the paper from a practical standpoint. 
It establishes the form of the Bayesian posterior under prevalence bias (section~\ref{sec: bias-corrected posterior}) and the resulting loss function (\eg, for neural network training) in section~\ref{sec: IG loss}. It also covers the key algorithmic elements (section~\ref{sec: marginal computations}) that underline our implementation. One technical contribution of wider scope is an efficient, unbiased and backpropable approximation of marginal distributions $p(y|w)$ of the outcome $y$ conditioned on the model parameters $w$, in neural networks and other arbitrary probabilistic models $p(y|x,w)$. The approach integrates seamlessly with the predominant paradigm of stochastic (minibatch) backpropagation. 

Section~\ref{sec: Bayesian analysis} lays out the formal Bayesian analysis. Section~\ref{sec: Generative model} presents the generative model. Section~\ref{sec: Bayesian risk} derives the posterior based on the principle of Bayesian risk minimization. This principle is the rationale from which training-time inference and test-time prediction rules are derived. Section~\ref{sec: test-time} discusses prevalence-adjusted test-time predictions, as a counterpart to section~\ref{sec: Bayesian IG in practice} for training.

\section{What is prevalence bias, and why does it matter?}
\label{sec: example}

\begin{figure*}[t]
\includegraphics[width=\textwidth]{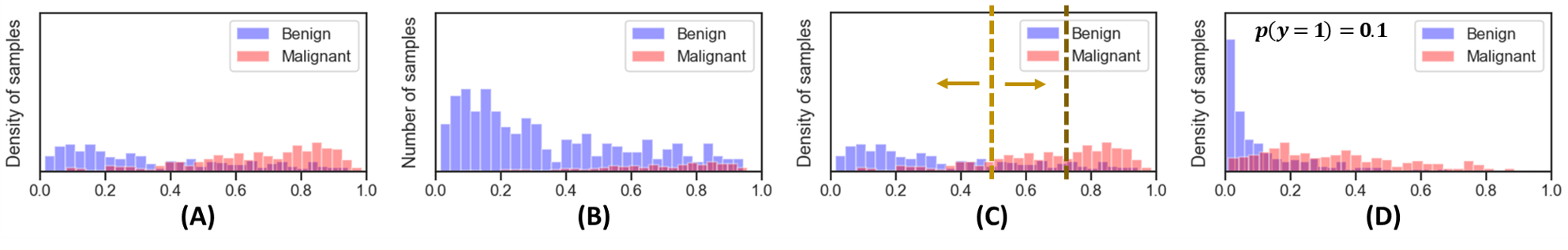}
\caption{Impact of prevalence, lung nodule example. (A) Predicted malignancy probabilities when trained on a balanced dataset. For malignant nodules, higher is better. For benign nodules, lower is better. Density histograms of true benign and malignant nodules are plotted separately and integrate to $1$ over $[0,1]$. The natural threshold $p_{thresh}\coloneqq 0.5$ is adequate for a balanced target population. This is not so for a population heavily skewed towards benign nodules, as shown in (B) for a prevalence $p(y\eq 1)$. (B) illustrates the corresponding relative class counts at various predicted malignancy probabilities. (C) The standard non-probabilistic solution is to adjust the operating point. Alternatively, (D) the paper shows how to incorporate knowledge of the prevalence at training time, 
so that predictions are by default optimal on the true population (illustrated for $p(y\eq 1)\coloneqq 0.1$). 
} 
\label{fig: lead example}
\end{figure*}

\textbf{The problem.} Consider the following screening scenario. The task is to predict nodule malignancy from metadata including subject demographics (age, sex, smoking habits), subject condition (emphysema) and high-level nodule appearance (diameter, opacity, location \ie lobe).
We are given a dataset ($679$ nodules) that represents positive ($357$ malignant nodules) and negative ($322$ benign nodules) classes almost equally. 
Suppose we train a neural network architecture $NN_w$ 
to output a predicted probability of malignancy 
from the aforementioned $7$-dimensional input metadata $x$. 
Fig.~\ref{fig: lead example}(A) shows the resulting histogram of predicted probability of malignancy 
for an example $1$-layer linear NN (similar results hold across a range of deep architectures – multi-layer, attention-based, etc. – and inference methods – VB, MAP, MLE, etc.). The accuracy on the 
dataset is around $75\%$. 
The misclassification rate among malignant and benign nodules is similar (Fig.~\ref{fig: lead example}(A)). 
This relative success is unlikely to translate well into the real-world.

Across a range of practical applications such as screening, the true prevalence of malignant nodules is drastically lower than in this balanced $50/50\%$ scenario -- what if one nodule in a hundred or in ten thousands is malignant in the target population? Since one benign nodule in four is misclassified, and given a population skewed towards healthy subjects, almost one person in four would be recalled for further examination. 

To contrast, a trivial classifier that outputs a constant benign prediction regardless of the input has closer to $0.5$ accuracy on the balanced training data, yet a much higher predictive accuracy on the imbalanced target population ($0.99$ accuracy for a prevalence of $1$ in $100$). Therefore~(1) predictive performance should be thought of as prevalence dependent;~(2) performance ``in the lab'' can be a misleading surrogate for performance ``in the wide'';~(3) accuracy as a sole metric poorly reflects predictive performance (it may rank higher a trivial classifier that is clearly not informed by the data).  \\

\textbf{Modelling sampling bias.} Skewing the prevalence of the training dataset compared to the general population is at times the only viable design choice (\eg, for rare diseases). The intended target population may also differ from the general population. Suppose the diagnostic happens after a referral (effectively a population filter), then the prevalence in the target subgroup is neither the true prevalence nor the apparent training prevalence. 
Hence our approach reasons with respect to an abstract ``true'' population in which the base model holds, that explanatory features $x$ cause (or precede) the outcome $y$. It is an idealization of the general population. Training and target populations are then defined as sub-populations obtained via selection mechanisms that shift prevalence. \\

\textbf{Reasoning under prevalence bias.} The goal is: (a) to infer the association between $x$ and $y$ in the true population in a training phase, possibly from a biased training sample; and (b) in the test phase for a target population (possibly also skewed), to inform accordingly the prediction. Hence two distinct questions: how to account for a shift in prevalence between true population and training set (affecting \textit{inference})? how to account for a shift between the true and target populations (affecting \textit{prediction})? 

A prevalence \textit{shift} (\textit{label shift}) can be either a causal byproduct of a distribution shift in \textit{input} $x$; 
or it can result from anticausal selection mechanisms directly acting on the distribution of \textit{labels} $y$. In the former case (covariate bias),
Bayesian principles eventually lead to the same formal solution 
as for the bias-free case both for inference (for a training distribution with covariate bias) and for prediction (for a target distribution with covariate bias). Hence the present focus is on the latter case, specifically referred to as \textit{prevalence bias}. 

Prevalence bias in the \textit{training} set has to be accounted for during \textit{inference}. Prevalence bias in the \textit{test} or \textit{target} population has to be accounted for during \textit{prediction}. 
Put together, one can \textit{infer} during training, from arbitrarily balanced observational data, an association $y\eq f(x,w)$ between input $x$ and label $y$ 
respecting prior knowledge that a disease is uncommon; then \textit{predict} the outcome optimally on various target populations (biased or not), with dedicated \textit{probabilistic} decision rules. 
Note that the case of biased test-time prevalence, known or unknown,
occurs commonly \eg, when performing hold-out validation, or if participating in a medical challenge where prevalence can be somewhat artificial.

To contrast, prevalence bias has historically been addressed via non-probabilistic \textit{test-time} heuristics. 
Traditionally one adjusts an operating point (the cut-off probability for the benign vs.~malignancy decision) to optimize a decision-theoretic cost, say of false positives vs.~false negatives. For instance in Fig.~\ref{fig: lead example}(A), shifting the probabilistic threshold for malignancy from $p_{thresh}\!\coloneqq\! 0.5$ to a higher value yields fewer false positives and higher accuracy. 
When the bias is inputable to the training data, it is misguided to ``fix'' prediction rather than inference. Trained under erroneous assumptions about the prevalence of the condition, the model is likely to learn erroneous associations between explanatory factors $x$ and outcome $y$.

\section{Training under prevalence bias: the Bayesian information gain}
\label{sec: Bayesian IG in practice}

The formal Bayesian analysis is conducted in section~\ref{sec: Bayesian analysis}. We anticipate here on the main result and its implementation. 
We consider a learning task in which a probabilistic model with parameters $w$ (\eg, a neural network and its weights) is trained from  data $X\eq\{x_n\}_{n=1}^N,Y\eq\{y_n\}_{n=1}^N$ to predict a dependent variable $y\!\in\!\calY$ (\eg the classification label) from inputs $x\!\in\!\calX$ (a.k.a., \textit{covariates} or features), optimally for a target population $x_\ast,y_\ast\!\sim\!\calD$.
We take the association between covariates $x$ and outcome $y$ 
to write in the form of a likelihood $p(y|x,w)$.

In binary classification for instance, the archetypal model is that the label $y\!\sim\!\calB(\sigma[\text{NN}_w(x)])$ results from a Bernoulli draw with probability conditioned on the input $x$. Namely the probability of $y\eq 1$ is obtained by squashing the output of a neural network architecture $\text{NN}_w(x)$ through a logistic link function $\sigma$. 

The main change in presence of prevalence bias in the training data is in the expression for the Bayesian parameter posterior $p(w|X,Y)$, reflected in the training loss.

\subsection{The bias-corrected posterior}
\label{sec: bias-corrected posterior}

Given a training set $X,Y$ with label-based sampling bias, the posterior on model parameters $w$ is expressed as:
\begin{equation}
{p(w|X,Y)} \enspace \propto \enspace 
\underbrace{
\vphantom{\frac{p(y_n|x_n,w)}{p(y_n|w)} }
p(w)
}_{\text{prior}} 
\; \cdot \; 
\prod_{n\leq N} \underbrace{
\frac{p(y_n|x_n,w)}{p(y_n|w)}
}_{\substack{\text{surrogate training}\\ \text{likelihood}}}\, ,
\label{eq: bias-corrected posterior}
\end{equation}
indexing samples by $n\eq 1\cdots N$, where the product runs over the full training batch. 
Contrary to the bias-free case, the bias-corrected posterior includes a normalizing factor in the denominator of the surrogate likelihood, the marginal $p(y_n|w)$. 
Thus the bias-corrected posterior captures the relative \textit{information gain} when conditioning on $x_n$ compared to an educated ``random guess'' based on marginal statistics.

Consider a high class imbalance setting where the probability of $y\eq1$ is small compared to that of $y\eq 0$. If the values $x$ are not known, one would by default place their bet on $y\eq 0$. In the prevalence bias-free case, it makes sense for the model to learn to predict $y\eq 0$ more often if the training data suggests so. In the prevalence bias scenario, the training statistics are a design artefact, and do not reflect those of the true population $\calD$. Hence the marginal in the denominator accounts for the fact that it is comparatively easier to predict $y\eq0$.

The posterior $p(w|X,Y)$ can be approximated using any standard strategy from Maximum Likelihood (ML) or Maximum A Posteriori (MAP) estimates to Variational Inference (VI)~\cite{blundell2015weight}, Expectation Propagation (EP)~\cite{minka2013expectation}, MCMC~\cite{chen2016bridging}. Next the presentation focuses on the case of MAP/ML estimates, which are most commonly used under the deep learning paradigm (it is easily adapted to the remaining inference techniques).

\subsection{The Bayesian IG loss}
\label{sec: IG loss}

From a practical deep learning standpoint, inference usually comes down to optimizing a loss function $\calL_{X,Y}(w)$ that writes as a sum of individual sample contributions, plus a regularizer, as in Eq.~\eqref{eq: batch loss function}:
\begin{equation}
\calL_{X,Y}(w) \triangleq -\log{p(w)} + \sum_{n\leq N} \calL_w(x_n,y_n)\, . 
\label{eq: batch loss function}
\end{equation}
The sum is over the training dataset (a.k.a., the full batch), and the sample loss $\calL_w(x,y)\!\triangleq\! -\log{p(y|x,w)}$ is the negative log-likelihood of the sample $x,y$ (output by the network). For a sigmoid or softmax likelihood, one retrieves the log-loss. For computational reasons, the minimizer $\hat{w}_{MAP}$ of Eq.~\eqref{eq: batch loss function} is often obtained by stochastic backpropagation. Given  a minibatch $B\eq\{n_1,\cdots,n_B\}$, one replaces the full-batch gradient $\nabla_w \calL_{X,Y}(w)$ by an unbiased minibatch estimate $\nabla_w \tilde{\calL}_{X,Y}(w)$, where $\tilde{\calL}_{X,Y}(w)$ is defined as per Eq.~\eqref{eq: unbiased minibatch estimate}:
\begin{equation}
\tilde{\calL}_{X,Y}(w) \triangleq -\log{p(w)} + \frac{N}{n_B}\sum_{n\in B} \calL_w(x_n,y_n)\, .
\label{eq: unbiased minibatch estimate}
\end{equation}
This estimate assumes that the minibatch is sampled i.i.d.~from the training data. Modified estimates given in~\ref{sec: generalized minibatch estimates} hold without restriction. 

The prevalence bias scenario directly mirrors the bias-free scenario. Taking the logarithm of Eq.~\eqref{eq: bias-corrected posterior} leads to the bias-corrected counterpart of Eq.~\eqref{eq: batch loss function}:
\begin{equation}
\calL_{X,Y}^{BC}(w) \triangleq -\log{p(w)} + \sum_{n\leq N} \calL_w^{BC}(x_n,y_n)\, .
\label{eq: batch loss function -- bias corrected}
\end{equation}
The contribution of a sample $(x,y)$ to the loss of Eq.~\eqref{eq: batch loss function -- bias corrected},
\begin{equation}
\calL_w^{BC}(x,y) 
\triangleq \log{p(y|w)} - \log{p(y|x,w)} \, ,
\label{eq: sample loss contribution}
\end{equation} 
is the negative information gain for the sample, \ie the log-ratio of the sample likelihood by the marginal. For computational reasons, the full-batch gradient $\nabla_w \calL_{X,Y}^{BC}(w)$ is replaced by a minibatch estimate $\nabla_w \tilde{\calL}_{X,Y}^{BC}(w)$. $\tilde{\calL}_{X,Y}^{BC}(w)$ is given by Eq.~\eqref{eq: unbiased minibatch estimate -- bias corrected} as a counterpart to Eq.~\eqref{eq: unbiased minibatch estimate}:
\begin{equation}
\tilde{\calL}_{X,Y}^{BC}(w) \triangleq -\log{p(w)} + \frac{N}{n_B}\sum_{n\in B} \calL_w^{BC}(x_n,y_n)\, .
\label{eq: unbiased minibatch estimate -- bias corrected}
\end{equation}
The only question is how to compute the marginal $p(y|w)$, as it is not a standard output of the network.

\subsection{Computing the marginal $p(y|w)$}
\label{sec: marginal computations}

$p(y|w)$ expands as the analytically intractable integral of Eq.~\eqref{eq: model marginal}:
\begin{equation}
p(y|w)=\int_\calX p(y|x,w)p_\calX(x)dx \, ,
\label{eq: model marginal}
\end{equation} 
whose computation requires evaluating and summing the network outputs $p(y|x,w)$ over the whole input space $\calX$. We demonstrate that efficient unbiased estimates of this quantity can be computed, and backpropagated through. Let $B\eq\{n_1,\cdots,n_B\}$ be a minibatch 
and $n_B(y)$ be the number of samples with label $y$ in $B$. \ref{sec: Appendix -- minibatch estimates} shows that the following empirical estimates 
based on the minibatch data are unbiased (the LHS and RHS are equal in expectation over the sample):
\begin{equation}
p(y|w)\simeq \hat{p}_B(y;w) \triangleq \frac{1}{n_B}\sum_{n\in B} \beta(y_n) \cdot p(y|x_n,w)\, ,
\label{eq: empirical marginal estimate}
\end{equation}
where the corrective weights $\beta(y)$ can be set to either one of the two values $\beta^{(1)}(y)$ or $\beta^{(2)}(y)$ from Eq.~\eqref{eq: corrective weights}:
\begin{equation}
\beta^{(1)}(y) \coloneqq \frac{p_\calY(y)}{\tilde{p}(y)}\, , \quad 
\beta^{(2)}(y) \coloneqq \frac{n_B}{n_B(y)} p_\calY(y) \, .
\label{eq: corrective weights}
\end{equation}
$p_\calY(y)$ 
stands for the probability of label $y$ in the true population, a.k.a. the true prevalence, and is assumed to be known. $\tilde{p}(y)$ in $\beta^{(1)}(y)$ refers to the expected distribution of labels in the minibatch. $\beta^{(1)}$ is a corrective factor based on \textit{expected} label counts for the minibatch, whereas $\beta^{(2)}$ is based on the actual (\textit{empirical}) label counts $n_B(y)$. Both choices lead to unbiased estimators, but their properties (\eg, variance) differ (section~\ref{sec: variance of estimators}). 

In general the variance of $\hat{p}_B(y;w)$ as a minibatch estimator, and the presence of a $\log$-nonlinearity, discourage the use of Eq.~\eqref{eq: empirical marginal estimate} as a direct plug-in replacement into $\calL_w^{BC}(x,y)$ of Eq.~\eqref{eq: sample loss contribution} \footnote{unless the full batch fits in memory, $B\!\coloneqq\!{1\cdots N}$, $n_B\!\coloneqq\! N$; such a batch implementation is hardly relevant for imaging data, but is suitable for less memory-intensive data and as a sanity-check for other implementations}. 
Instead, we approximate the marginal $p(y|w)$ via an auxiliary neural network $q_\psi: w\mapsto q_\psi(w)$ with trainable parameters $\psi$, where $q_\psi(w): y\in\calY \mapsto q_\psi(w)(y)$ assigns a probability to every outcome $y\!\in\!\calY$\footnote{In the practical implementation, numerical stability suggests for the auxiliary network to output $\log{q_\psi(w)}$ instead, and to exponentiate only if needed.}. 
The auxiliary network is trained (jointly with the main model) by minimizing the Kullbach-Leibler divergence, which takes a friendly form in this context (\ref{sec: KL auxiliary training}).

It is again tempting to plug $q_\psi(w)(y)\!\simeq\! p(y|w)$ directly into $\calL_w^{BC}(x,y)$ of Eq.~\eqref{eq: sample loss contribution}, and to train the main model $NN_w(x)$ by backpropagating through this approximation. This implies backpropagating through $q_\psi(w)(y)$ w.r.t. $w$. Our second insight is to avoid this.
This allows to use a generic, lightweight $q_\psi(w)$. Indeed in all experiments we use either a simple linear layer followed by a softmax activation (a.k.a. a logistic regressor), or an even simpler $|\calY|$-dimensional bias vector. 

To that aim we derive the gradient of $\log{p(y|w)}$ in closed form (\ref{sec: Appendix -- minibatch estimates}), and an unbiased estimate as per Eq.~\eqref{eq: log marginal gradient}:
\begin{equation}
\nabla_w \log{p(y|w)}\simeq \frac{\hat{p}_B(y;w)}{p(y|w)}\cdot\nabla_w \log{\hat{p}_B(y;w)}\, . 
\label{eq: log marginal gradient}
\end{equation}
All quantities involved are available, except the exact marginal $p(y|w)$. Finally we plug the approximate marginal $q_\psi(w)$, yielding the final minibatch estimate of the gradient:
\begin{equation}
\nabla_w \log{p(y|w)}\simeq \frac{\hat{p}_B(y;w)}{q_\psi(w)(y)}\cdot\nabla_w \log{\hat{p}_B(y;w)}\, . 
\label{eq: log marginal gradient -- approximation}
\end{equation}

To sum up, $\nabla_w \calL_w^{BC}(x,y)\eq \nabla_w \log{p(y|x,y)} - \nabla_w \log{p(y|w)}$ is evaluated by replacing the last term with its approximation from Eq.~\eqref{eq: log marginal gradient -- approximation}. Contrast with a solution that computes the gradient of the approximate marginal: $\nabla_w \log{p(y|w)}\!\simeq\! \nabla_w \log{q_\psi(w)}$.
In the former case, $\nabla_w \log{q_\psi(w)}$ is never used nor computed, so that the auxiliary network $q_\psi(w)$ only needs provably accurate $0$th order approximation, rather than accurate $1$st order gradients approximation. This is crucial for practical applications, where model parameters $w$ are high dimensional and $p(y|w)$ has intricate dependencies w.r.t. variations of $w$. Autodifferentiation libraries such as \textit{pytorch} allow for a straightforward implementation of the minibatch estimates from Eq.~\eqref{eq: empirical marginal estimate},\eqref{eq: log marginal gradient -- approximation} 
by implementing a custom backward routine. The computational logic is clarified in~\ref{sec: Computational logic with automatic differentiation}.

\subsection{High-level pseudo-code}

\begin{algorithm}
	\For{minibatch $X_B,Y_B$ in data loader}{
		forward $X_B$ through the main model $NN_w$ for sample log-likelihoods\;
		forward $w$ through auxiliary $q_\psi$ for log-marginal estimates~\ref{sec: auxiliary network architectures}\;
		compute main (section~\ref{sec: IG loss},~\ref{sec: Computational logic with automatic differentiation}) and auxiliary (\ref{sec: KL auxiliary training}) losses from $NN_w(X_B)$, $Y_B$, and $q_\psi(w)$\;
		backward on losses and step: update $w$, $\psi$
	}
 \caption{Training epoch overview}
\end{algorithm}

The implementation has small overhead, with an additional forward pass through a minimalistic auxiliary network, the computation of the auxiliary loss to train this network, and the additional backward step to update its parameters.  
\section{Case study: prevalence-bias, Bayesian posterior and log odds ratio}
\label{sec: logodds}

Consider the example data of the contingency table~\ref{table: contingency table}. Through this example we will peek into the behaviour of the prevalence-bias corrected model, contrasting its behaviour with established approaches. Taking the table at face value, it would seem that $X=1$ potentially increases the probability of having the condition. The frequentist estimate of $p(Y=1|X=0)$ is $44/91\simeq 0.48$, that of $p(Y=1|X=1)$ is $6/9\simeq 0.66$. Of course, given the scarcity of data for $X=1$, one expects high uncertainty attached to this claim (we will account for uncertainty within the Bayesian paradigm, although frequentist confidence intervals are straightforward to derive here).

Upon closer look, the apparent frequency of $Y$ in the contingency table is identical across the two outcomes (condition / no condition). Clearly such a prevalence of one in two is unreasonably high in a medical scenario, for most conditions. It is reasonable to assume that the sample was artificially balanced to have equal counts along columns $Y\eq 0$ and $Y\eq 1$. Assume the true prevalence known (for the sake of illustration, $1$ in $100$). What can we say about the association between $X$ and $Y$? What confidence is attached to the statement?

\begin{table}[h!]
  \begin{center}
    \caption{Example data \eg, association of a phenotype $X$ with condition $Y$.}
    \label{table: contingency table}
    \begin{tabular}{c@{\hskip .5em}|@{\hskip 2em}c@{\hskip 2em} c}
      \, & $Y=0$ & $Y=1$\\
      \toprule
      $X=0$ & $47$ & $44$ \\ 
	  $X=1$ & $3$ & $6$ \\            	
      \midrule
      Total counts & $50$ & $50$ \\ 
	  True prevalence & $0.99$ & $0.01$ \\ 
    \end{tabular}
  \end{center}
\end{table}

A useful statistics in this context is the so-called (log-)odds ratio, commonly used in the medical literature~\cite{bland2000odds, szumilas2010explaining}, given by Eq.~\eqref{eq: odds ratio}: 
\begin{equation}
OR\triangleq \frac{p(Y=1|X=1)/p(Y=0|X=1)}{p(Y=1|X=0)/p(Y=0|X=0)}
\label{eq: odds ratio}
\end{equation}
Notice how the odds ratio is unchanged if replacing $p(Y\eq y|X\eq x)$ with unnormalized quantities. One then verifies that the roles of $X$ and $Y$ can be switched in the odds ratio computation, \ie Eq.~\eqref{eq: odds ratio bis} holds:
\begin{equation}
OR=\frac{p(X=1|Y=1)/p(X=0|Y=1)}{p(X=1|Y=0)/p(X=0|Y=0)}\, .
\label{eq: odds ratio bis}
\end{equation}
Hence the odds ratio is unchanged regardless of whether sampling is bias-free or prevalence-biased. 
The frequentist estimate of $OR$ on the data is $2.14$. The log odds ratio $\log{OR}$ (estimate $0.76$, st.d.~$\simeq 0.74$) symmetrizes the roles of $X\eq 0$ and $X\eq 1$, so that switching labels simply switches the sign of $\log{OR}$. The log odds ratio naturally arises in logistic regression. Suppose then the following model:
\begin{equation}
p(Y= 1|X= 0)\triangleq \sigma(\eta_0)\, , \quad p(Y= 1|X= 1)\triangleq \sigma(\eta_1)\, ,
\label{eq: logistic model}
\end{equation}
where $\sigma(\eta)\triangleq 1/(1+\exp{-\eta})$ is the sigmoid function and $\eta_0$, $\eta_1$ so-called logits. It follows from Eq.~\eqref{eq: logistic model} that $\log{OR}=\eta_1 - \eta_0$. In Bayesian inference, the goal is to infer the value of logits $\eta_0, \eta_1$, from which to build predictive rules for various statistics or probability values. From the previous remarks, one would then expect the posterior distribution to place more probability mass in regions compatible with log odds ratio values close to $0.76$, up to some uncertainty due to the scarcity of data. Denote $p_0\triangleq p(Y= 1|X= 0)$ and $p_1\triangleq p(Y= 1|X= 1)$ the probability of positives when $X$ takes value $0$ or $1$. When visualizing the posterior probability in the $(p_0,p_1)$ plane, the locus of points such that $\log{OR}=\alpha$ is a curve of equation $\logit{p_1} - \logit{p_0} = \alpha$, \textit{viz.} $p_1=\sigma(\alpha+\logit{p_0})$.

\begin{figure}
\includegraphics[width=\columnwidth]{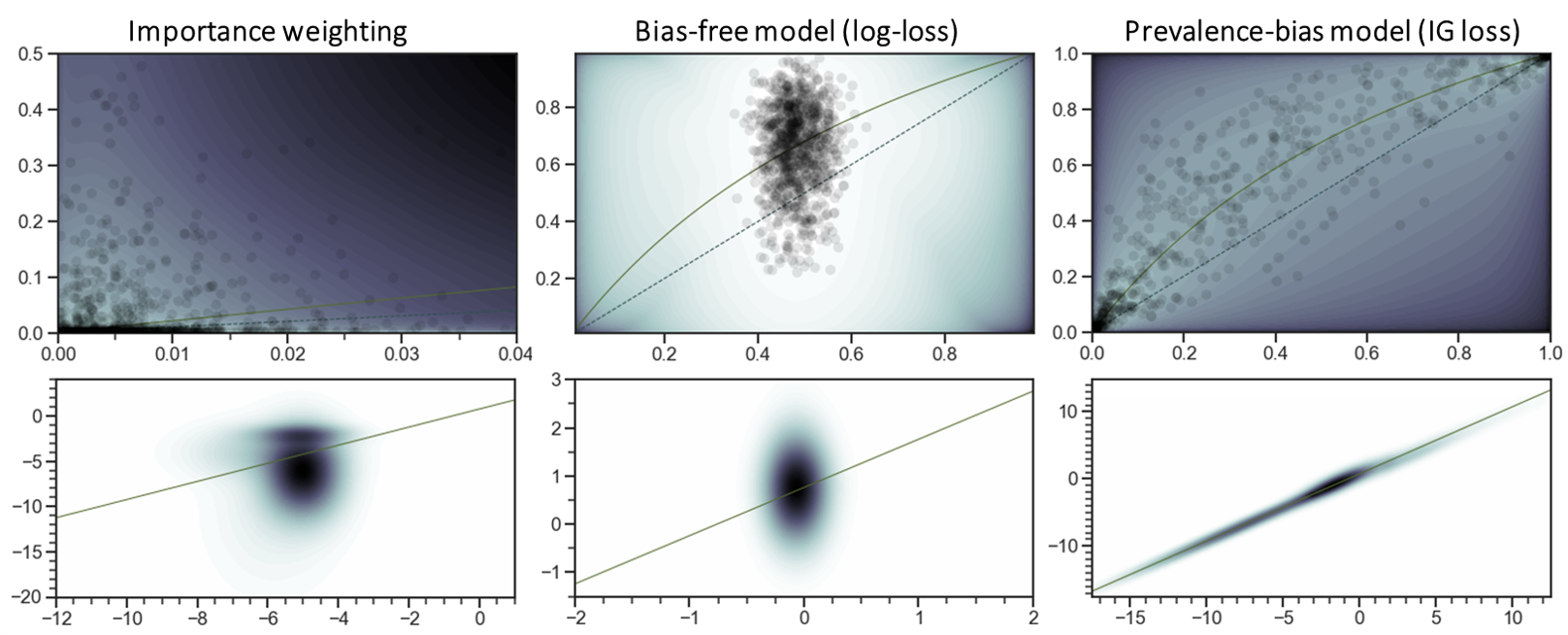}
\caption{Summary of Bayesian inference for the data of Table~\ref{table: contingency table}, for importance weighting (first column), a standard bias-free model (second column) and the proposed prevalence-bias model (third). Denote $p_0\triangleq p(Y= 1|X= 0)=\sigma(\eta_0)$ and $p_1\triangleq p(Y= 1|X= 1)=\sigma(\eta_1)$. Top row: probability distribution in $(p_0,p_1)$ space. Colour encodes the approximate posterior log-probability density (higher is brighter). As an alternative visualization, $1000$ samples (semi-transparent black dots) from the approximate posterior are overlaid. The dotted black line is the $p_0\eq p_1$ line. The green curve is the $0.76$-isocurve of log odds ratio (cf.~text). Bottom row: approximate posterior log-probability density in $(\eta_0,\eta_1)$ logit space (darker is higher). The green line $\eta_1-\eta_0=0.76$ is the $0.76$-isocurve of log odds ratio.} 
\label{fig: contingency 2d comparison}
\end{figure}

\begin{figure}
\includegraphics[width=\columnwidth]{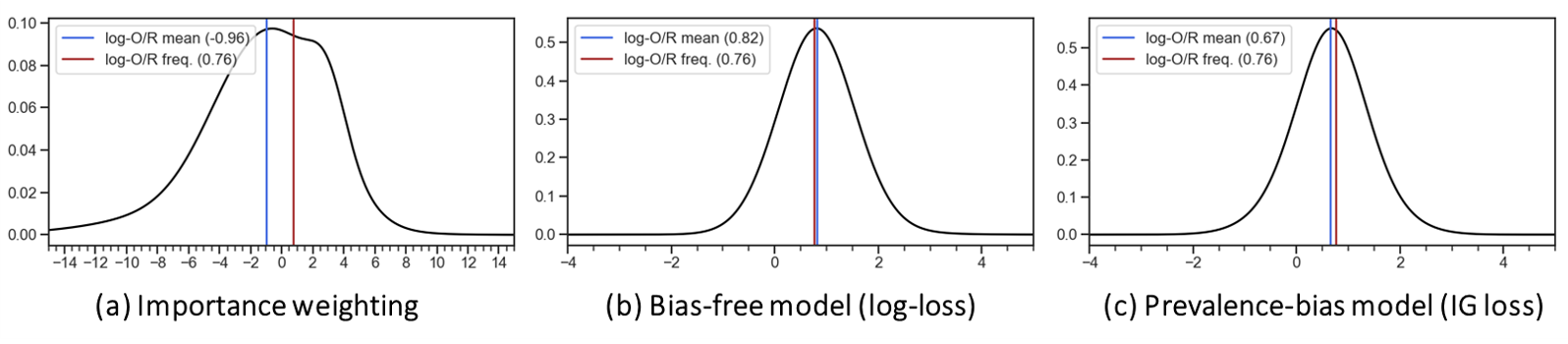}
\caption{Approximate posterior distribution of the log odds ratio for the data of Table~\ref{table: contingency table}, for: (a) importance weighting, (b) the standard bias-free model, and (c) the proposed prevalence-bias model. The brown line is the frequentist estimate at $0.76$, the blue line the mean posterior value.} 
\label{fig: contingency 1d comparison}
\end{figure}

We assume (independent) normal distributions $\eta_x\ssim \calN(0, 10^2)$ as priors on $\eta_0$ and $\eta_1$. Inference is done and compared across three (meta-)models: a standard bias-free Bayesian model $X\to Y$ (therefore using a cross-entropy loss, a.k.a.~log-loss), an importance-weighted log-loss (anticipating on section~\ref{sec: Frequentist estimator}), and finally the proposed Bayesian model of prevalence-based sampling bias (using the loss of section~\ref{sec: Bayesian IG in practice}). We optimize a variational evidence lower bound (ELBO) with the general purpose Adam optimizer~\cite{kingma2014adam}, using a variational family $q_{{\eta}}({\eta})\triangleq \text{GMM}({\eta}; \{\mu_k, \Sigma_k\}_{k\leq K})$ of multivariate Gaussian Mixture Models with $K\eq 8$ components as an approximate joint posterior on $\eta=(\eta_0,\eta_1)$. Results are summarized in Fig.~\ref{fig: contingency 2d comparison} and Fig.~\ref{fig: contingency 1d comparison}. Several comments are in order.

In Fig.~\ref{fig: contingency 1d comparison}, the mass of the posterior probability lies around the frequentist estimate, for both the bias-free and prevalence-bias Bayesian models. There is significant uncertainty in the estimate, making a probabilistic Bayesian approach \textit{a fortiori} valuable (in this textbook example it is also consistent with available frequentist confidence intervals). The output of the importance-weighted log-loss is unconvincing. The reweighting of samples makes this approach fundamentally lack a Bayesian interpretation and ill-suited to variational inference\footnote{weights may sum to any chosen value $N_w\in\R_+^\ast$; here, $N_w\ccoloneqq 100$ the total number of observations, so that on average one data point $\equiv$ one observation. 
In reality penalized maximum likelihood inference is better suited to this approach (setting the strength of the regularizer by trial-and-error); this comes at the cost of pointwise inference, but we do so in subsequent experiments.}.

Fig.~\ref{fig: contingency 2d comparison} highlights how probabilistic estimates qualitatively differ depending on whether the prevalence bias is accounted for. The standard model (second column) miscalibrates the probability $p(Y\eq 1|X\eq x)$ and estimates that the uncertainty about $\eta_1$ (resp. $p(Y\eq 1|X\eq 1)$) is uncorrelated with that about $\eta_0$ (resp. $p(Y\eq 1| X\eq 0)$). It is easily seen upon inspection of the loss function, that under this model inference proceeds independently over the \textit{rows} of Table~\ref{table: contingency table}, regardless of the fact that data was sampled i.i.d. \textit{column-wise}. The prevalence-bias model accounts for this dynamics, with the probability mass here again clearly lying along log odds ratio isocurves, \textit{viz.}~there is low uncertainty in the value $\eta_1-\eta_0$ relative to the uncertainty in the orthogonal direction. \\

\textbf{Encoding prevalence.} In the present case the sampling-bias corrected model does not accurately pinpoint the true prevalence either. This illustrates that the the bias-corrected model is about accounting for the \textit{mechanism} of prevalence-based \textit{sample selection} (\ie, ``anticausal'' column-wise i.i.d.~sampling), rather than injecting knowledge of the actual prevalence. Notice in fact that the true prevalence does not explicitly enter Eq.~\eqref{eq: bias-corrected posterior}\footnote{The prior $p(w)$ and the likelihood terms $p(y|x,w)$ do not depend on the prevalence. The marginal terms $p(y|w)=\int_x p(y|x,w)p(x)\, dx$ have an implicit dependence on the true prevalence through $p(x)$ (see~\ref{sec: Appendix -- minibatch estimates}), but this dependence can be non-specific when the association between $X$ and $Y$ is weak.
}. 

\begin{figure}
\includegraphics[width=\columnwidth]{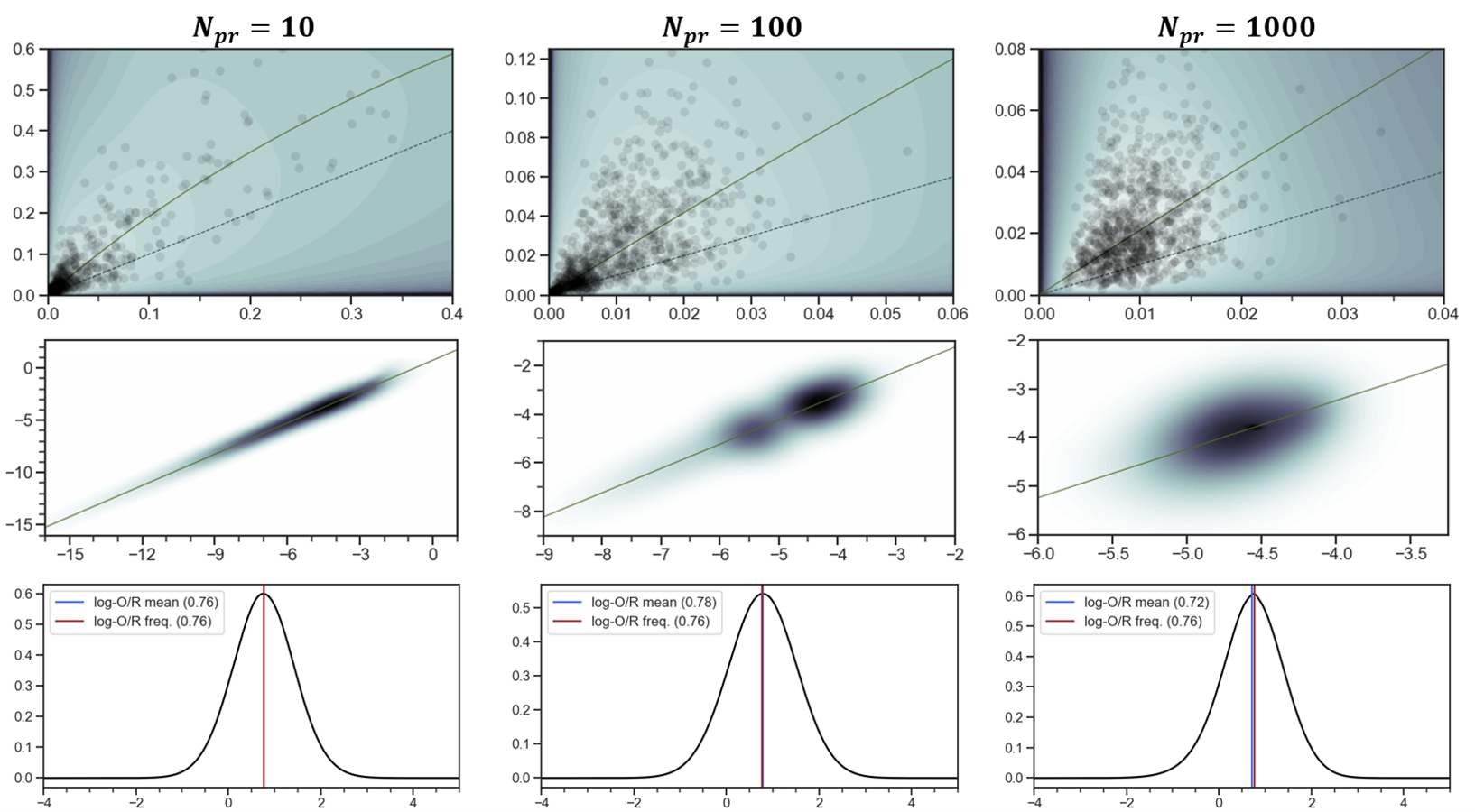}
\caption{Approximate posterior distribution for the model of prevalence-based sampling bias, with a prevalence prior. As in Fig.~\ref{fig: contingency 2d comparison}, the top row is in $(p_0,p_1)$-space and the middle row in $(\eta_0,\eta_1)$-space. Bottom row is in log odds ratio space as in Fig.~\ref{fig: contingency 1d comparison}. From left to right: adjusting the number of prior observations $N_{pr}$.} 
\label{fig: contingency -- varying prevalence prior}
\end{figure}

The knowledge of the prevalence can instead naturally be added as a Bayesian prior. Indeed, one can reasonably assume this knowledge to stem from prior experience, having observed a certain number of samples $Y=y$ (but not necessarily of $X$), a fraction of which were positive samples $Y\eq 1$. In other words, across a total of $N_{pr}$ prior observations one has observed $Y\eq y$ with a frequency $\hat{p}_\calY(y)$. These prior observations (of $Y$ alone) induce an additional factor $\prod_{y\in\calY} p(y|w)^{N_{pr}\hat{p}_\calY(y)}$ in the posterior distribution, which translates to the complementary term of Eq.~\eqref{eq: loss -- prevalence term} entering additively the loss function of Eq.~\eqref{eq: batch loss function -- bias corrected}: 
\begin{equation}
\calL_{prev}(w) = - N_{pr}\sum_{y\in\calY} \hat{p}_\calY(y) \log{p(y|w)}\, .
\label{eq: loss -- prevalence term}
\end{equation} 
Equivalently this reads, up to additive constant, as a Kullback-Leibler (KL) divergence penalty term $-N_{pr}\, KL[\hat{p}_\calY(y)\Vert p(y|w)]$. $N_{pr}$ regulates the strength of the prior. Fig.~\ref{fig: contingency -- varying prevalence prior} illustrates its effect on the inferred posterior. Notice how the prediction more and more confidently focuses around the specified prevalence, while displaying a consistent general behaviour and uncertainty w.r.t.~the log odds ratio.

\section{Bayesian analysis of the prevalence bias}
\label{sec: Bayesian analysis}
 
Bayesian decision theory provides the theoretical backbone for the previous sections. Section~\ref{sec: Generative model} specifies the generative model. 
Section~\ref{sec: Bayesian risk} derives the model posterior and predictive posterior from decision theoretic arguments, via Bayes' utility.

\subsection{Generative model.} 
\label{sec: Generative model}

\begin{figure}[b]
\centering
\includegraphics[width=\columnwidth]{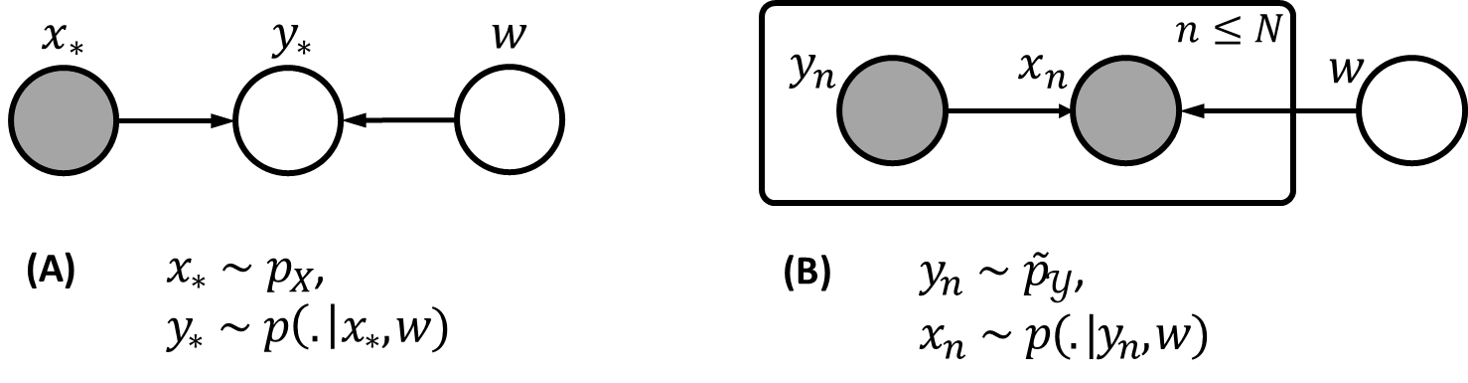}
\caption{Generative model (A) of the true population $(x_\ast,y_\ast)\!\sim\!\calD$ and (B) of the training dataset $(x_n,y_n)\!\sim\!\calD'$ under the label-based sampling bias of Fig.~\ref{fig: sampling biases}(B). Model parameters $w$ are implicitly shared.}
\label{fig: graphical models}
\end{figure}

The core property of the scenario of prevalence bias is that the apparent prevalence $\tilde{p}(y)$ during training is an artefact of the experimental design. Contrast with the true prevalence $p_\calY(y)$ in a population, which reflects facts about the real world. The causal models are different. Fig.~\ref{fig: graphical models} emphasizes structural differences in generative mechanisms for the true population and for the training data. \ref{sec: prevalence bias as sample selection} provides an alternative (but equivalent) viewpoint with a single generative model, from the angle of sample selection.\\

\textbf{True population} $(x_\ast,y_\ast)\!\sim\!\calD_w$. 
Let $y_\ast\!\sim\! p(y_\ast|x_\ast,w)$ depend on the causal variables $x_\ast$ according to a probabilistic model 
with parameters $w$ (Fig.~\ref{fig: graphical models}(a)). \textit{E.g.}, age, sex and life habits ($x$) may condition the probability of developing cancer ($y$). Image data ($x$) might condition the patient management ($y$). The sampling process $(x_\ast,y_\ast)\!\sim\!\calD_w$ expands as sampling $x_\ast\!\sim\!p_\calX$, then an outcome
$y_\ast\!\sim\!p(\cdot |x_\ast,w)$. \\

\textbf{Training dataset} $X,Y\!\sim\!\calD_w'$. By assumption of prevalence bias, labels $Y\!\sim\! \tilde{p}(Y)$ are sampled first (Fig.~\ref{fig: graphical models}(b)). For instance the training set is deliberately balanced, regardless of the true prevalence. Then $x_n\!\sim\!p(x_n|y_n,w)$ is drawn uniformly according to the true conditional distribution $\calD_{x|y_n,w}$. \\

\textbf{Structural implications.} In the \textit{true population}: the outcome $y$ depends on $w,x$. But $x\pperp w$. This is the familiar setting, and the marginal distribution $p(y|w) \eq \int_\calX p(y|x,w)p(x)dx$ gives the true population prevalence for the true association $w\!\coloneqq\! w^\ast$, \ie $p_\calY(y)\ccoloneqq p(y|w^\ast)$. More generally the marginal depends on the strength $p(y|x,w)$ of association of label $y$ with explanatory features $x$ (described by the parameters $w$), and on the population distribution $p_\calX$. The strength of association $p(y_\ast|x_\ast,w)$ between cause and outcome itself does not depend on $p_\calX$, and would remain unaffected by population drift ($\triangleq$ change of $p_\calX$).

At \textit{training time}, data comes from biased sampling. The notation $\tilde{p}$ emphasizes that the training marginal for $y$ is an artefact of the dataset design, unrelated to $p(y|w)$ nor $p(y)\eq\int_w p(y|w)p(w)dw$\footnote{in fact we never refer to $p(y)\eq\int_w p(y|w)p(w)dw$ in the present work}. In particular $y\!\pperp\! w$. To sample $x$ (``if $y\eq1$, sample $x$ uniformly among malignant cases''), one implicitly inverts $p(y|x,w)$ via Bayes' rule. Thus $x$ depends on $w,y$.

\subsection{Bayesian risk minimization, prediction and inference} 
\label{sec: Bayesian risk} 

Bayesian inference is grounded in Bayesian decision theory via the concept of Bayesian utility~\cite{berger1985statistical} or Bayesian risk. Since the present setting has the somewhat unusual property that training and test generative models structurally differ, we briefly restate the principle and rederive the predictive posterior as a risk-minimizing prediction rule.

Define a (supervised) predictive task as that of finding an optimal probabilistic prediction rule $q_{x_\ast,X,Y}(y)$ for new observations $x_\ast\!\sim\!p_\calX$, given training data $X,Y\!\sim\!\calD'_w$. 
A probabilistic prediction rule $q_{x_\ast,X,Y}$ is a probability distribution over $\calY$ that is allowed to depend on $x_\ast,X,Y$. It assigns a probability to all possible outcomes based on the knowledge of the causal variables $x_\ast$ and of known data $X,Y$. 

Let $\calE(q_\calY,y)\triangleq -\log{q_\calY(y)}$ define a probabilistic \textit{loss} incurred by a distribution $q_\calY$ for a given $y\!\in\!\calY$. $\calE$ can then be used to quantify the loss $\calE(q_{x_\ast,X,Y},y_\ast)$ incurred for a sample $(x_\ast,y_\ast)$ by the probabilistic decision rule $q_{x_\ast,X,Y}$.  

If the parameters $w\!\coloneqq\! w^\ast$ involved in the data generation process were known, one could then define a prediction \textit{risk} $\calR_w$ for a given prediction rule, as an average loss w.r.t.~the data distributions $(x_\ast,y_\ast)\sim \calD_w$ and $(X,Y)\sim \calD'_w$, yielding\footnote{the risk takes as input a family of predictors $q_{x_\ast,X,Y}$: one for each possible $(X,Y)$ and $x_\ast$, and we write $\calR_w[q_{x_\ast,X,Y}]$ for short, but as an abuse of notation}:
\begin{equation}
\calR_{w}[q_{x_\ast,X,Y}] \triangleq \mathbb{E}_{(X,Y)\sim\calD'_w}\!\left[
								\mathbb{E}_{(x_\ast,y_\ast)\sim\calD_w}\left[\calE\left(q_{x_\ast,X,Y}, y_\ast\right)\right]
						  \right]\, .
\label{eq: risk for known w}
\end{equation}
In fact it is easy to see that, were $w\!\coloneqq\! w^\ast$ known (and the prediction rule hence allowed to use these known values), the optimal rule $q_{x_\ast,X,Y}\coloneqq q_{x_\ast,w}$ would no longer depend on the observed $X,Y$. Optimizing Eq.~\eqref{eq: risk for known w} is equivalent to optimizing Eq.~\eqref{eq: frequentist risk -- intermediate},~\eqref{eq: frequentist risk}:
\begin{align}
\calR_{F,w}[q_{x_\ast,w}] & = \mathbb{E}_{(x_\ast,y_\ast)\sim\calD_w}\left[\calE\left(q_{x_\ast,w}, y_\ast\right)\right] \, , \label{eq: frequentist risk -- intermediate} \\
\, & = \mathbb{E}_{(x_\ast,y_\ast)\sim\calD_w}\left[-\log{q_{x_\ast,w}(y_\ast)}\right] \, .
\label{eq: frequentist risk}
\end{align}
The optimal prediction is the likelihood $q_{x_\ast,w}(y_\ast)\eq p(y_\ast|x_\ast,w)$, and we recognize in $\calR_{F,w}$ the \textit{frequentist} risk. In practice however, the true value $w^\ast$ of model parameters $w$ is unknown. 

Hence the Bayesian prediction risk is defined as the expectation of the prediction risk $\calR_w$ of Eq.~\eqref{eq: risk for known w} w.r.t.~a prior distribution $p(w)$ of $w$:
\begin{equation}
\calR_{\text{Bayes}}[q_{x_\ast,X,Y}(y)] \triangleq 
\mathbb{E}_{w\sim p(w)}\!\left[
	\calR_{w}[q_{x_\ast,X,Y}]
\right] \, .
\label{eq: Bayes prediction risk}
\end{equation}  
Eq.~\eqref{eq: Bayes prediction risk} states that the optimal rule should remain adequate under all reasonable ``world-generating'' values $w$. The posterior predictive distribution $p(y_\ast|x_\ast,X,Y)$ minimizes $\calR_{\text{Bayes}}$ as usual for the logarithmic loss $\calE$ defined previously (\ref{sec: proofs}). $p(y_\ast|x_\ast,X,Y)$ expands as a weighted sum over the space of model parameters:
\begin{equation}
p(y_\ast|x_\ast,X,Y)=\int_w \underbrace{p(y_\ast|x_\ast,w)}_{\text{likelihood}}\underbrace{p(w|X,Y)}_{\text{posterior}}dw\, .
\label{eq: predictive posterior}
\end{equation}
The point of departure from the bias-free setting is in the exact form of the posterior $p(w|X,Y)$. 
Prevalence bias induces a change in the structural dependencies between model variables \eg, $Y\!\pperp\! w$, leading to Eq.~\eqref{eq: bias-corrected posterior}. The proof reported in~\ref{sec: proofs} relies on this insight. 

\subsection{The posterior predictive and optimal policies}
\label{sec: optimal policies}

The choice of the logarithmic loss $\calE(q_\calY,y)\triangleq -\log{q_\calY(y)}$ gives an optimality result on the posterior predictive distribution, and will facilitate the comparison with the established importance weighting approach in the next section; but it may appear somewhat arbitrary. A much stronger result holds, linking the predictive posterior to all optimal policies w.r.t.~all Bayes risks $\calR_{\calE,\text{Bayes}}$ defined for arbitrary loss functions $\calE$. 

Consider a set of actions $a\in\calA$ (recalling a patient for further examination or not, say). Let $d_{X,Y}:x\in\calX\mapsto d_{X,Y}(x)\in\calA$ define a \textit{decision rule} or \textit{policy}, \ie~a choice of action $d_{X,Y}(x_\ast)$ for a given observation $x_\ast$. Let $\calE:\calA\times\calY\rightarrow \R$ define a user-specified loss $\calE(a,y)$ for policy $a$ when the sample label is $y$. Loss values can be chosen to reflect an asymmetry in the consequences of various actions $a$ in given scenarios $y$. Say $\calA\coloneqq \calY$, the consequence of misclassifying a sample as $d_{X,Y}(x_\ast)\ccoloneqq1$ when $y_\ast=0$ (type I error), may differ from that of mistakenly assigning $d_{X,Y}(x_\ast)\ccoloneqq 0$ when $y_\ast=1$ (type II error). Finally, define the Bayes risk $\calR_{\calE,\text{Bayes}}[d_{X,Y}]$ corresponding to the loss $\calE$ similarly to Eq.~\eqref{eq: Bayes prediction risk}. Say again $\calA\coloneqq \calY$, the argmax rule $d_{X,Y}:x\!\in\!\calX\mapsto \argmax_y p(y|x, X, Y)$ is clearly not optimal under all Bayes risks. 

It still holds nonetheless that to construct an optimal policy $d_{X,Y}$, it is sufficient to know the predictive posterior distribution (\ref{sec: proofs}). The asymmetry in costs $\calE(a,y)$ translates to various thresholds on the predictive posterior distribution.

\subsection{Related work: frequentist risk and importance weighting}
\label{sec: Frequentist estimator}

Frequentist and Bayesian analyses depart in their approach to coping with the unknown model parameters $w$. In the frequentist case, one recognizes that the likelihood $p(y|x,w)$ is the minimizer of Eq.~\eqref{eq: frequentist risk}, which motivates the search for a suitable estimator $\hat{w}=h(X,Y)$. In the bias-free case, plugging $p(y|x,\hat{w})$ in Eq.~\eqref{eq: frequentist risk} and taking the expectation over the empirical data distribution instead, motivates the Maximum Likelihood Estimator (MLE) $\calL_{MLE}$ of $\hat{w}$:
\begin{equation}
-\calL_{MLE}[\hat{w}] \triangleq \mathbb{E}_{(x,y)\sim\hat{\calD}}[\log{p(y|x,\hat{w})}]\, ,
\label{eq: MLE}
\end{equation}
Eq.~\eqref{eq: MLE} is now a finite sum over the empirical distribution $\hat{\calD}$ of the training data, 
with density $\hat{p}(x,y)\!\triangleq\!\sum_{n\leq N} \delta_{x_n,y_n}(x,y) / N$. In presence of prevalence bias, a derivation similar to that of~\ref{sec: proofs} shows that here again, an unbiased estimate can be derived as per Eq.~\eqref{eq: weighted MLE} from the empirical data distribution $\hat{\calD}'$, provided that we introduce corrective weights:
\begin{equation}
-\calL_{MLE}[\hat{w}] \triangleq \mathbb{E}_{(x,y)\sim\hat{\calD}'}[\beta(y)\log{p(y|x,\hat{w})}]\, , \quad \beta(y) \coloneqq \frac{p_\calY(y)}{\tilde{p}(y)}\, . 
\label{eq: weighted MLE}
\end{equation}
It is known from the statistical literature that the MLE estimator is prone to overfitting. It is often discarded in favour of other penalized estimators within the framework of Empirical Risk Minimization (ERM)~\cite{vapnik1992principles}. Typically one adds a regularizer $-\lambda R(\hat{w})\!\equiv\!\log{p(\hat{w})}$ to Eq.~\eqref{eq: weighted MLE}, which plays a role analogous to a Bayesian prior. Henceforth we refer to either Eq.~\eqref{eq: weighted MLE} or its regularized variant as the \textit{weighted (log-)loss} or as \textit{importance weighting}~\cite{huang2007correcting}. The weighted loss is a natural point of comparison for the proposed approach. \\

\textbf{A logical fallacy?} Importance weighting captures the intuition that  
prevalence bias 
can be handled via corrective weights that ``make it look like the data comes from the true distribution''.
In practice the correction can be unreliable when applied to imbalanced classes, which has spurred alternative weighted heuristics~\cite{lin2017focal}. 

We argue that a choice of parameters $w$ for the model is not merely a statement about the likelihood $p(y|x,w)$ of association between $x$ and $y$. It is also a statement about the prevalence $p(y|w)$ of outcomes. For different values of $w$, the models are biased towards different outcomes, so that minimizing an empirical risk of the form $\sum_{X,Y} \log{p(y_n|x_n,w)}$ plays on two chords: capturing the correct association between $x$ and $y$, and biasing the model prevalence towards the apparent prevalence. 

In presence of prevalence bias, the apparent prevalence carries no information. Regardless of the hypothetical variant $w$ of the ``real world'' that is factual, the dataset would have been collected with an artificial prevalence $\tilde{p}(y)$, disregarding the prevalence $p(y|w)$. Therefore 
it is sensible to remove from the empirical risk the contribution of the prevalence for each sample: 
\begin{equation}
-\calL_{MLE}^{BC}[\hat{w}]\triangleq \sum_{X,Y} \log{p(y_n|x_n,\hat{w})} - \log{p(y_n|\hat{w})}\, . 
\label{eq: modified MLE estimator}
\end{equation}
Equivalently Eq.~\eqref{eq: modified MLE estimator} can be introduced as an MLE for the likelihood $p(x_n|y_n,w)$ (inversed via Bayes rule). Either insight recovers the proposed approach (Eq.~\eqref{eq: sample loss contribution}) from a frequentist viewpoint.

\begin{figure*}
\includegraphics[width=\textwidth]{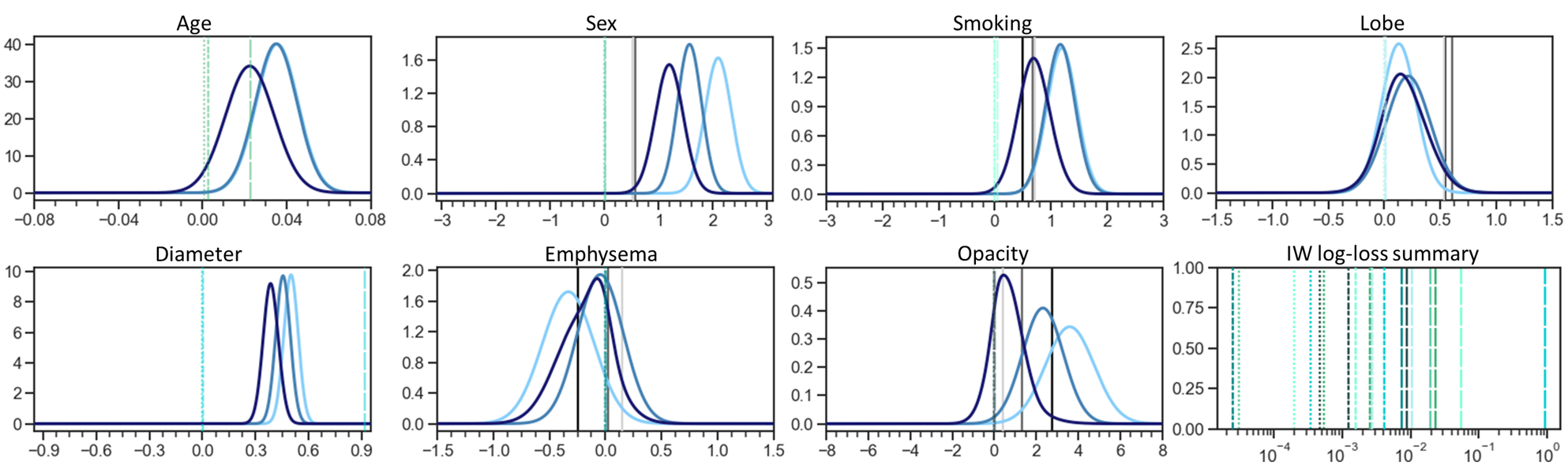}
\caption{Inferred log odds ratios for all seven covariates in lung nodule malignancy prediction, at three presumed malignancy prevalence values ($0.1$, $0.01$, $0.001$), for the Bayesian prevalence-bias (IG) model vs.~the importance weighted (IW) log-loss baseline. IG returns posterior probability density estimates, IW pointwise (penalized ML) estimates. One plot per covariate. First seven plots: IG posterior p.d.f.~shown as dark blue (resp. medium, light blue) curves, IW point estimates as lines (large, small dashes or dotted) at prevalence $0.1$, resp. $0.01$, $0.001$. Also shown as full lines for all categorical covariates (\ie, not age/diameter): frequentist point estimates from empirical marginal distributions on training (light grey), all (medium) and test data (dark grey). Eigth, bottom-right plot: summary of IW log-OR estimates, in log-scale for emphasis (all but $2$ estimated values were positive; the $2$ negative values $-0.000106$ and $-0.000364$ at prev.~$0.001$ for lobe and emphysema are not shown). Colours and patterns match that of corresponding lines on corresponding covariate plots. The magnitude is inconsistent as prevalence changes (hence all dotted lines, say, are clustered; rather than all lines of a given colour/covariate).}
\label{fig: calibration -- lor}
\end{figure*}

\section{Case study: calibration of the probabilistic predictions}
\label{sec: calibration}

Let us turn towards an application to prediction of lung nodule malignancy from seven covariates: age (between $18$ and $90$), sex (M/F), smoking habits (binary, based on frequency and time span of habit), presence of emphysema (binary condition), nodule location (lower or upper lobe) and appearance (diameter, solid or part-solid state). We use data provided by the Royal Brompton \& Harefield NHS Foundation Trust. The dataset contains $679$ nodules ($357$ malignant and $322$ benign) with metadata and malignancy diagnosis. 

We are specifically interested in whether probabilistic predictions are well-calibrated, and how they change under different assumptions on the prevalence of malignant nodules (for the sake of illustration: $0.1$, $0.01$ or $0.001$). We compare predictors built from the Bayesian prevalence bias model, and from the importance weighted log-loss. The likelihood of malignancy $y=1$ is parametrized via a logistic regressor $p(y=1|x,\eta)\triangleq \sigma(z(x))$, where $\sigma(z)\triangleq 1/(1+\exp{-z})$ is the sigmoid function and $z(x)= \sum_{1\leq i\leq 7}\eta_ix_i + \eta_0$ aggregates the bias $\eta_0$ and logits $\eta_i$ for all seven covariates $x_i$. For the Bayesian approach, we place (independent) Student-t priors $\eta_i\ssim t_\nu(\eta_i)$ ($\nu\coloneqq 0.002$) on the logits $\eta_i$, as an approximately scale-invariant non-informative prior. As a variational family $q_{{\eta}}({\eta})\triangleq \text{GMM}({\eta}; \{\mu_k, \Sigma_k\}_{k\leq K})$ jointly on $\eta=(\eta_i)_{0\leq i\leq 7}$, we use multivariate Gaussian Mixture Models with $K= 8$ components. The prior knowledge of prevalence is introduced as in section~\ref{sec: logodds}, setting $N_{pr}=10$. For inference in the weighted log-loss approach, regularized Maximum Likelihood Estimation (MLE) is used. Predictors are trained via stochastic modified gradient descent, through $500$ steps of Adam optimization~\cite{kingma2014adam}, on a training fold of $302$ samples ($150$ for validation). The remaining third of the dataset ($227$ nodules) is used for testing. \\

\textbf{Learnt associations.} $p(y\eq 1|x_i,x_{-i},\eta)/p(y\eq 0|x_i,x_{-i},\eta)$ describes the odds of nodule malignancy, for given values of a variate $x_i$ of interest and covariates $x_{-i}$. The logistic model entails considerable simplifications. The odds write as $\exp{\sum_i \eta_i x_i + \eta_0}$. The change in odds for a change $x_i\to \tilde{x}_i$ is equal to $\exp{\eta_i(\tilde{x}_i-x_i)}$, which does not depend on the value of covariates $x_{-i}$. Taking an incremental change $\Delta x_i\ccoloneqq 1$ if the variate is continuous, or $0\to 1$ for a binary variate, yields an odds ratio $OR_i$, which can be interpreted as the effect of the $i$th variate on $y$ (independent of the marginal distribution of $x$). Its logarithm is exactly $\log{OR_i}=\eta_i$. Appealing to Bayes' rule:
\begin{equation}
\begin{split}
\log{OR_i}=\log&{\frac{p(x_i=1,x_{-i}|y=1, \eta)}{p(x_i=0,x_{-i}|y=1, \eta)}} \\
	&- \log{\frac{p(x_i=1,x_{-i}|y=0, \eta)}{p(x_i=0,x_{-i}|y=0, \eta)}}\, .
\end{split}
\label{eq: calibration -- lor from Bayes rule}
\end{equation} 
From Eq.~\eqref{eq: calibration -- lor from Bayes rule} along with section~\ref{sec: Generative model}, the odds ratio is preserved in the presence of prevalence-based sampling bias. Fig.~\ref{fig: calibration -- lor} reports the inferred associations, across models and assumed malignancy prevalence. 

Unlike for the univariate example of section~\ref{sec: logodds}, direct empirical estimates of $\log{OR_i}$ are not available. As a surrogate, empirical statistics $\log{\left(\frac{\hat{p}(x_i=1|y=1)}{\hat{p}(x_i=0|y=1)} / \frac{\hat{p}(x_i=1|y=1)}{\hat{p}(x_i=0|y=1)}\right)}$ based on marginals, can be computed. We report them in Fig.~\ref{fig: calibration -- lor} as plain grey lines, with separate estimates on training and test data to give a sense of the intra-dataset variability. Note that correlations among covariates can introduce a mismatch between these quantities and log odds ratios. Still, they provide a gauge for the relevance of inferred $\log{OR_i}=\eta_i$ and their uncertainty. 

The approach based on prevalence bias modelling displays a consistent behaviour (w.r.t.~the sign, magnitude and uncertainty surrounding an association) across order-of-magnitude changes in the assumed prevalence ($0.1$, $0.01$ or $0.001$). This is not so for the importance weighted approach – so much so that the assumed prevalence is a better predictor of the magnitude of the inferred $\eta_i$ than the covariate index ($1\leq i\leq 7$).\\

\begin{figure}
\includegraphics[width=\columnwidth]{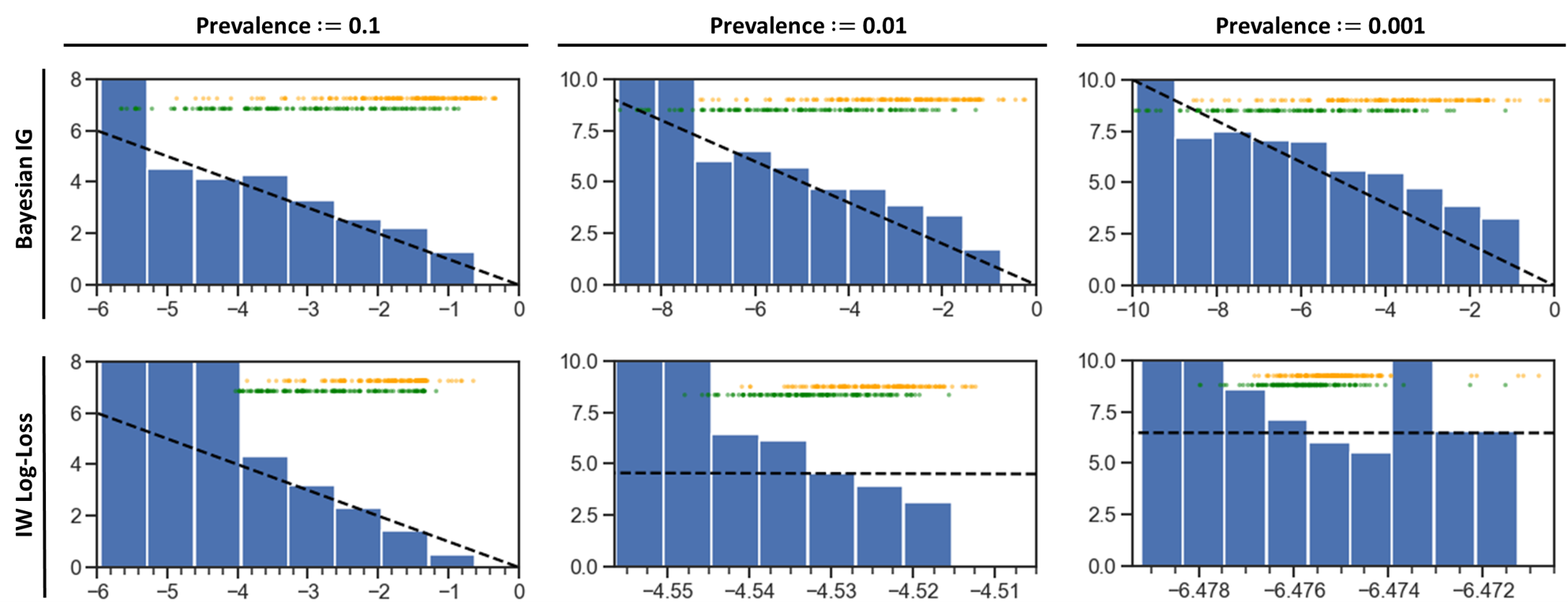}
\caption{Calibration of probabilities on the training data. Each bar plot represents the frequency $\hat{f}_{\text{tr}}$ of real positives in the probability bin, where along the $x$-axis $p_{\text{tr}}$ is the expected likelihood of malignancy. $\hat{f}_{\text{tr}}$ weighs training samples with a prevalence correction $\beta(y)\ccoloneqq p_\calY(y)/\hat{p}_{\text{tr}}(y)$ so as to be an unbiased estimate on the true population. Both axes are in log scale ($\exp{-3}\simeq 0.05$, $\exp{-6}\simeq 2.5\cdot 10^{-3}$, $\exp{-9}\simeq 1.2 \cdot 10^{-4}$), the $x$-axis shows $\log{p_{\text{tr}}}$ and the $y$-axis $-\log{\hat{f}_{\text{tr}}}$. The dashed black line manifests the ideal calibration $\hat{f}_{\text{tr}}=p_{\text{tr}}$. Predicted malignancy probabilities for all $302$ samples are overlaid near the top of each plot (real negatives $\equiv$ green; real positives $\equiv$ orange; the $y$-axis location is indifferent).} 
\label{fig: calibration -- training}
\end{figure}

\begin{figure*}
\includegraphics[width=\textwidth]{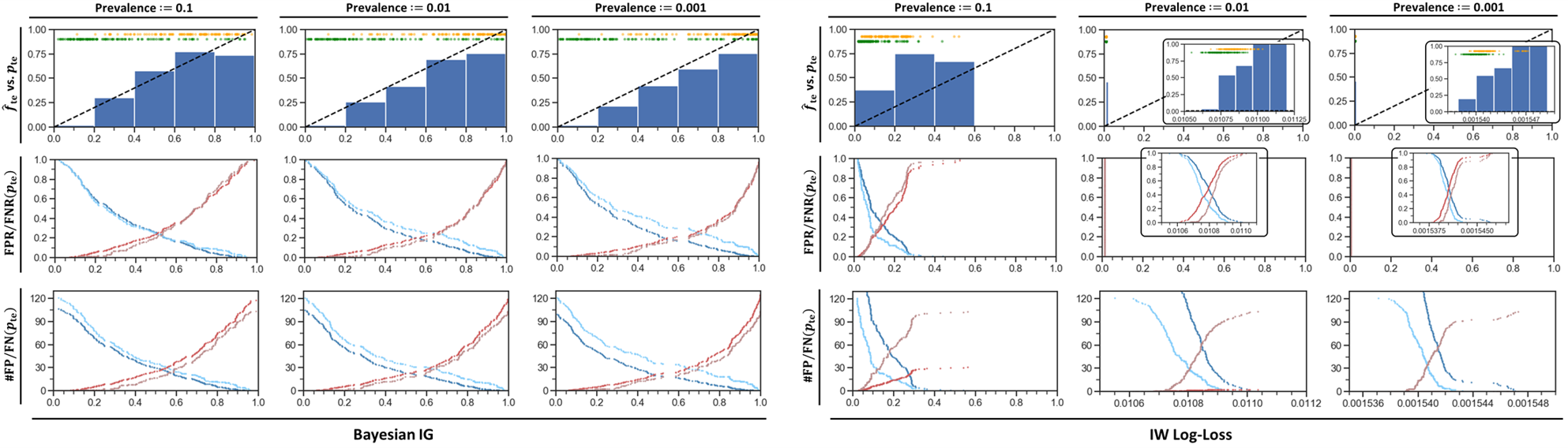}
\caption{Calibration of predictive probabilities on the held-out fold, for prediction rules built from the Bayesian prevalence bias model (IG), and from the importance weighted (IW) log-loss, at three assumed malignancy prevalences ($0.1$, $0.01$, $0.001$). Bar plots: frequency $\hat{f}_{\text{te}}$ of real positives in the probability bin, where along the $x$-axis $p_{\text{te}}$ is the predicted probability of malignancy. The dashed black line manifests the ideal calibration. Predicted malignancy probabilities for all $227$ samples are overlaid near the top of each plot (real negatives $\equiv$ green; real positives $\equiv$ orange; the $y$-axis location is indifferent). Middle row plots: expected (dark blue) vs.~empirical (light blue) FPR, respectively expected (red) vs.~empirical (brown) FNR at all malignancy probability thresholds $p_{\text{te}}$ (see main text). Good overlap means good calibration. Bottom row: Same with the (unnormalized) FP/FN counts.
} 
\label{fig: calibration -- test}
\end{figure*}

\textbf{Calibration of predictive probabilities.} We investigate two aspects of probabilistic calibration. Firstly, how inference reconciles the need to account for prevalence and the goal to separate positives and negatives – thus we look at the learnt probabilities $p(y_n|x_n,\eta)$ for training data (Fig.~\ref{fig: calibration -- training}). Secondly, we look at the calibration of predictive probabilities on test data (Fig.~\ref{fig: calibration -- test}). This anticipates on a key point formally described next in section~\ref{sec: test-time}, that the test sample (\eg, the held-out fold) may itself have prevalence bias, independent of the optimal regime for which the classifier was constructed – a scenario that a principled approach based on statistical modelling is able to account for.   

In Fig.~\ref{fig: calibration -- training}, the behaviour of the Bayesian IG-based and IW-based approaches strikingly differ, despite both approaches respecting the prevalence specification (the model marginal $p(y\eq 1|\eta)\simeq p_\calY(y\eq 1)$ closely matches the specified prevalence). The former makes use of the full range of likelihood values $p(y|x,\eta)$ up to values close to $1$, or $0$ in log-scale, even at small malignancy prevalences. At prevalence $0.001$ say, log-likelihood values span $10$ orders of magnitude, as the algorithm attempts to reconcile the low marginal probability 
with 
confident predictions for clear-cut real positives. Contrast with the IW-based approach, for which the log-likelihood values cluster close to the prevalence value, except for infinitesimal variations that yield better discrimination; resulting in poor calibration.

Whereas Fig.~\ref{fig: calibration -- training} illustrates how the Bayesian IG inference scheme used for training gives incentive to learn a well-calibrated likelihood w.r.t.~to the true population for the assumed prevalence, Fig.~\ref{fig: calibration -- test} looks at the calibration of the predictive posterior $p(y_\ast|x_\ast,X,Y)$ on the test-time distribution. For this example the held-out fold consists of a roughly balanced set of $122$ benign samples and $105$ malignant samples, which does not match the assumed prevalence. The proposed approach achieves the seemingly antinomic goals to make well-calibrated predictions for this test sample as well as for the general population, thanks to the Bayesian framework introducing an additional step in the computation of the predictive posterior in the presence of test-time prevalence bias (section~\ref{sec: test-time}). Specifically, the framework estimates the unknown test-time prevalence on the fly jointly from its own predictions, refined accordingly. 

The adequate calibration of the predictive posterior can be leveraged for optimal decision making in the sense of arbitrary loss functions, as argued in section~\ref{sec: optimal policies}. Imagine an asymmetric cost of false positives and false negatives, so that one wants fine-grained control of false positive and negative rates $FPR$ and $FNR$.  
The probability of a false positive $p_\theta(FP)$, respectively of a false negative $p_\theta(FN)$ at a given decision threshold $\theta$, for a given sample $x$, noting $p_{\text{te}}$ for short instead of $p(y=1|x,X,Y)$, is given by:
\begin{align}
p_\theta(FP)&=\left\{\begin{array}{ll}
    0 &\text{if } p_{\text{te}}<\theta \\
    1-p_{\text{te}} &\text{otherwise}
  \end{array} \right. \, , \\ 
p_\theta(FN)&=\left\{\begin{array}{ll}
    0 &\text{if } p_{\text{te}}\geq\theta \\
    p_{\text{te}} &\text{otherwise}
  \end{array} \right. \, .
\end{align}
Hence estimates of the expected FPR and FNR for the test distribution, at any threshold $\theta$, can be computed from the test sample of $227$ data points. First sort the $p_{\text{te}}$ in ascending order (let $p_{\text{te},n}$ be the $n$th value). Let $\hat{N}_0\triangleq \sum_n (1-p_{\text{te},n})$ estimate the expected number of negatives in the sample, and $\hat{N}_0(\theta)\triangleq\sum_{n\leq n(\theta)}(1-p_{\text{te},n})$ the expected number of negatives such that $p_{\text{te}}\leq p_{\text{te},n(\theta)}$. Then $1-\hat{N}_0(\theta)/\hat{N}_0$ estimates $FPR$ at threshold $\theta$, if $n(\theta)$ stands for the largest $n$ such that $p_{\text{te},n} < \theta$. The reasoning is similar for $FNR$. 

If ground truth labels are known, expected values $FPR(\theta)$ and $FNR(\theta)$ (or $FP(\theta)$ and $FN(\theta)$) obtained from the predictive posterior can be compared to the empirical values obtained from the actual count of false positives and negatives at threshold $\theta$. The middle and bottom row in Fig.~\ref{fig: calibration -- test} shows that the expected and empirical values have reasonable agreement for the Bayesian IG approach. In other words, one could implement a decision rule based on a specified trade-off between $FP$ and $FN$ counts, from the predictive posterior probabilities, without availability of ground truth for the target population.\\

\begin{table}\centering
\ra{.8}
\small
\tabcolsep=0.14cm
\begin{tabular*}{\columnwidth}{rcccccccc}\toprule

& ELL & $\text{Acc}_{HO}$ & {TNR}& {TPR} & $\text{NPV}_{HO}$& $\text{PPV}_{HO}$ & AUC\\ \midrule
pr. $=10^{-1}$ \\	
{IG} & $-0.307$ & $0.77$ & $0.71$ & $0.84$& $0.84$& $0.72$& $0.83$& \\ 
{IW} & $-0.125$& $0.55$ & $1.0$ & $0.02$& $0.54$& $1.0$& $0.81$& \\[0.2cm]
pr. $=10^{-2}$ \\	
{IG} & $-0.373$ & $0.77$ & $0.72$ & $0.83$& $0.83$& $0.72$& $0.83$ \\ 
{IW} & $-0.015$ & $0.54$ & $1.0$ & $0.0$& $1.0$& NaN & $0.81$ \\[0.2cm]
pr. $=10^{-3}$ \\	
{IG} & $-0.443$ & $0.77$ & $0.70$ & $0.84$ & $0.84$ & $0.71$ & $0.83$ \\ 
{IW} & $-0.008$ & $0.54$ & $1.0$ & $0.0$ & $1.0$ & NaN & $0.79$ \\

\bottomrule
\end{tabular*}
\label{table: calibration -- performance}
\caption{Performance summary on held-out data for the Bayesian approach to prevalence-bias (IG) vs.~importance weighted log-loss (IW). $\text{Acc}_{HO}$: accuracy. TPR / NPR: positive / negative rates. $\text{PPV}_{HO}$ / $\text{NPV}_{HO}$: positive / negative predictive values.}
\end{table}

\textbf{Predictor performance.} For completeness several performance metrics are reported in Table~\ref{table: calibration -- performance}. The better calibration of the Bayesian modelling approach does not come at a cost in predictive performance. In addition its performance is stable across the range of prevalence hypotheses. One metric w.r.t.~which importance weighting performs better is the expected log-likelihood (ELL) w.r.t.~the true population. This is no surprise since in the IW approach, the true population ELL is the metric optimized during training (see also Fig.~\ref{fig: calibration -- training}). The gaps in performance observed for IW with regard to other metrics is largely bridged if one allows for recalibration of the decision threshold between positives and negatives, as reflected in the AUC. The fact that results equally support multiple hypotheses and multiple models is a stark reminder of the limitations of the ``training-test dataset split, cross-validation'' paradigm to assess the correctness and usefulness of a model as a representation of the real-world.

\section{Prevalence bias at validation and test time}
\label{sec: test-time}

We now focus on test-time predictions on data unseen at training time (whether from a validation set, from a benchmark test set, or from a target real-world population). The optimal Bayesian prediction for a datapoint $x_\ast$ 
depends on the test population, 
and specifically on whether that distribution is also biased.
For instance, validation data may be a hold-out subset of the biased training dataset; 
or upon deployment, the target population for the model may be a select subgroup from the general population (\eg, symptomatic people).

Consider the two following cases: the data is sampled from the true population, for which the natural causal insights of Fig.~\ref{fig: graphical models}(A) hold; or it is sampled with artificial class prevalence as per Fig.~\ref{fig: graphical models}(B). 
The first scenario was already described in section~\ref{sec: Bayesian risk}. For the second, the expression for the predictive posterior changes. Algorithm~\ref{alg: test-time} summarizes the computations as pseudo-code.\\

\begin{algorithm}
  \KwIn{Trained main and auxiliary models $NN_{\hat{w}}$, $q_{\hat{\psi}}$.}
  \KwOut{Predictive probabilities $q_{x_\ast,X,Y}(y_\ast)$ for the test data $x_\ast \!\in\! X_B$}
  	\If{prevalence bias}{
		forward trained $\hat{w}$ through trained auxiliary $q_{\hat{\psi}}$ for log-marginal estimates $p(y_\ast|\hat{w})$
	}
	\For{minibatch $X_B$ in test data loader}{
		forward $X_B$ through the trained model $NN_{\hat{w}}$ for sample log-likelihoods $p(y_\ast|x_\ast,\hat{w})$\;	
		\uIf{prevalence bias}{	
			compute the prediction $q_{x_\ast,X,Y}(y_\ast)$ via Eq.~\eqref{eq: likelihood with selection}
		}
		\Else{	
			set $q_{x_\ast,X,Y}(y_\ast)\coloneqq p(y_\ast|x_\ast,\hat{w})$
		}
	}
 \caption{Test-time overview, bias-free or with test-time prevalence bias}
\label{alg: test-time}
\end{algorithm}

\noindent
\textbf{Without test-time bias.} The predictive posterior $p(y_\ast|x_\ast,X,Y)$ expands as $\int_w p(y_\ast|x_\ast,w)p(w|X,Y)dw$. It can be estimated by closed form or Monte Carlo integration with an approximate parameter posterior $q(w)\!\simeq\! p(w|X,Y)$. For the case of pointwise inference, $q(w)$ collapses to a point estimate $\hat{w}$ and one retrieves the likelihood estimate $p(y_\ast|x_\ast,\hat{w})$. This is the standard deep learning recipe of running the test point $x_\ast$ through the trained network $NN_{\hat{w}}(x_\ast)$ to get the probabilistic outcome $y_\ast$. (The same applies in presence of covariate bias).\\

\noindent
\textbf{With test-time bias.} Suppose label frequencies $\tilde{p}(y_\ast)$ in some hold-out set $\calD_{out}$ are defined by the researcher and arbitrary, possibly differing both from the apparent training prevalence and the true prevalence. If one is allowed to inform the prediction with these frequencies, the optimal prediction should be adjusted. A Bayes risk can again be defined, where the loss is now averaged w.r.t.~$\calD_{out}$. 
The predictive posterior writes differently 
owing to different assumptions on the hold-out distribution (namely, selection bias). From~\ref{sec: test-time selection} the likelihood $p(y_\ast|x_\ast,w)$ in the integrand is replaced by:
\begin{equation}
p_{s_\ast=1}(y_\ast|x_\ast,w) \propto 
\frac{p(y_\ast|x_\ast,w)}{p(y_\ast|w)} \cdot \tilde{p}(y_\ast)\, ,
\label{eq: likelihood with selection}
\end{equation}
where we let $s_\ast\eq1$ explicitly denote a selection process.  
The predictive posterior is $\int_w p_{s_\ast=1}(y_\ast|x_\ast,w) p(w|X,Y)dw$ and it reduces down to $p_{s_\ast=1}(y_\ast|x_\ast,\hat{w})$ for pointwise inference of $w\!\coloneqq\! \hat{w}$. Thus the DL recipe is to run the test point through the trained network $NN_{\hat{w}}(x_\ast)$ for $p(y_\ast|x_\ast,\hat{w})$ and through the trained auxiliary network $q_{\hat{\psi}}(\hat{w})$ for the marginal $p(y_\ast|\hat{w})$, from which to compute the conditional likelihood of Eq.~\eqref{eq: likelihood with selection}.

What if the biased test-time distribution $\tilde{p}(y_\ast)$ is unknown? Say one takes part in a challenge, where the organiser designs a benchmark with a certain balance of class labels that is not communicated to the participant. The Bayesian framework gives us the chance to jointly infer the hidden distribution of labels along with the label values. The derivation is reported in~\ref{sec: test-time selection}. 
 
\section{Performance metrics}
\label{sec: Performance metrics}

Benchmarking based on a single metric (\eg, accuracy) is likely to give an incomplete and skewed picture of performance. This is crucial in medical imaging, even more so in retrospective studies or when a precise specification of the clinical context is missing. The section motivates a range of metrics that are subsequently used in the case studies. 
We split these metrics in three overlapping subgroups:~(1) expected risks;~(2) summary statistics derived from a confusion matrix; and~(3) performance metrics (\eg, AUC) and summary curves (\eg, ROC) whose definition implicitly allows for a test-time ``surgery'' whereby an operating point (\eg cut-off probability) can be moved around to generate a family of predictions with different trade-offs.

\begin{figure*}[t]
\includegraphics[width=\textwidth]{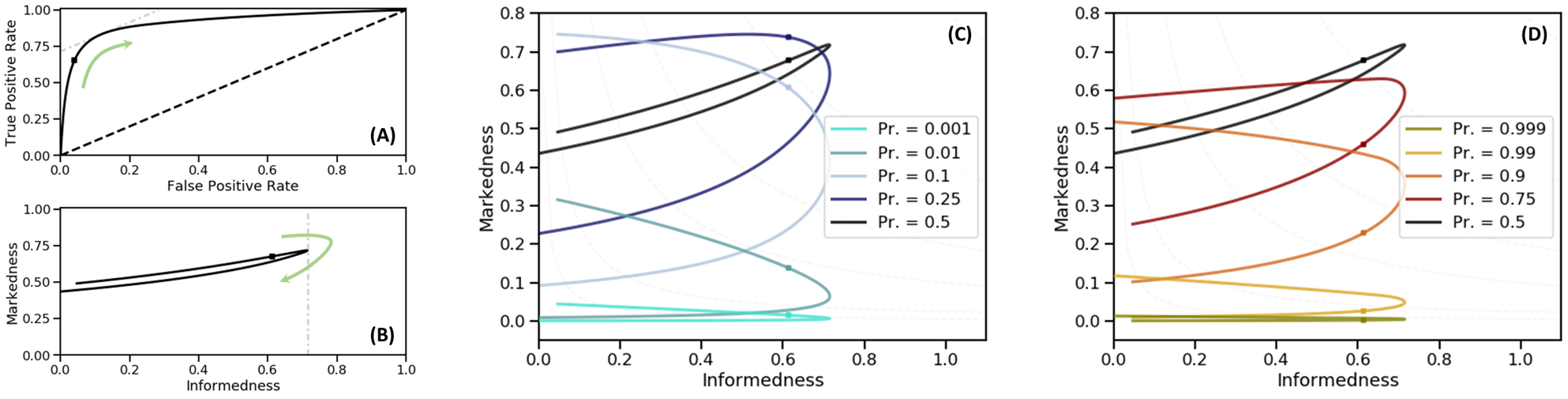}
\caption{ROC and IM curves: impact of varying the true prevalence on the prediction performance of a reference classifier. (A) ROC curve for a mock binary classifier. The square marks an arbitrary operating point along the curve, corresponding to some cut-off probability. Dashed-black line: chance-level line corresponding to $I=0$. Dash-dot grey line: isoline $I=I_{max}$ tangent to the point of maximum informedness. (B) Corresponding Informedness-Markedness curve at prevalence $0.5$, with the tangent now a $y$-axis aligned line at $I_{max}$. Curved green arrow: corresponding directions of travel on ROC and IM curves. What if the test-time prevalence of real positives ($t=1$) changes, but the classifier is not retrained nor the operating point adjusted? (C) Effect of a decrease in real positive prevalence on the IM curve. (D) Same for an increase in prevalence. The ROC curve is unchanged, as well as the informedness and the $x$-value of an operating point on the IM curve. Notice from the location of the square how the upper and lower branches of the IM curve are switched at low vs.~high prevalence. Moreover the maximum of informedness gradually switches from a maximum of markedness to a minimum of markedness, from balanced to extremely imbalanced prevalences, forcing an $I,M$ trade-off. Isolines of $MCC=\sqrt(I\cdot M)$ dashed and grey in the background of (C) and (D).} \label{fig: IM curve 101}
\end{figure*}

\subsection{Expected true population and hold-out risks}
\label{sec: expected risks}

These are statistics derived from Bayesian or frequentist risks as defined in section~\ref{sec: Bayesian risk}. They reward sensible calibration of predicted probabilities (confident predictions when right, and unconfident mispredictions, \ie~mild errors). In practice they are evaluated on a held-out fold $\hat{\calD}_{out}$. When $\hat{\calD}_{out}$ is itself subject to prevalence bias, this leads to two scores depending on whether we view the test dataset as the definitive benchmark or as a proxy to evaluate the true population risk.\\

\textit{True population risk}. A natural application-agnostic metric is the expected predictive risk of Eq.~\eqref{eq: expected predictive risk}:
\begin{equation}
-{\calR}_{\calD^\ast}[q_{x_\ast,X,Y}] = 
\mathbb{E}_{(x_\ast,y_\ast)\sim\calD^\ast}\left[ \log{q_{x_\ast,X,Y}}(y_\ast) \right] \, ,
\label{eq: expected predictive risk}
\end{equation}
where $\calD^\ast\!\triangleq\! \calD_{w^\ast}$ is the true population. 
It encourages optimality w.r.t.~the true prevalence, 
but can be estimated from $\hat{\calD}_{out}$ by importance weighting. Whenever the prediction rule is of the form $q_{x_\ast,X,Y}(y_\ast) \triangleq p(y_\ast|x_\ast,\hat{w})$ for some estimator $\hat{w}$ of $w$, Eq.~\eqref{eq: expected predictive risk} collapses to the expected log-likelihood of Eq.~\eqref{eq: expected log-likelihood}:
\begin{equation}
-{\calR}_{\calD^\ast}[\hat{w}] = \mathbb{E}_{(x_\ast,y_\ast)\sim\calD_{w^\ast}}\left[ \log{p(y_\ast|x_\ast,\hat{w})} \right] \, .
\label{eq: expected log-likelihood}
\end{equation}
This is the case for the proposed approach under MAP inference ($\hat{w}\coloneqq\hat{w}_{MAP}$) as well as for the importance weighted estimator. In practice Eq.~\eqref{eq: expected log-likelihood} is estimated on $\hat{\calD}_{out}$ leading to Eq.~\eqref{eq: expected log-likelihood estimate}, using corrective weights $\beta(y)\coloneqq \beta^{(2)}(y)$ of Eq.~\eqref{eq: corrective weights}:
\begin{equation}
-\hat{\calR}_{\calD^\ast}[\hat{w}] = \text{avg}_{(x_\ast,y_\ast)\in\hat{\calD}_{out}}\left[ \beta(y)\log{p(y_\ast|x_\ast,\hat{w})} \right] \, .
\label{eq: expected log-likelihood estimate}
\end{equation}
Since Eq.~\eqref{eq: expected log-likelihood estimate} is an estimate from a finite sample, an additive correction $k \,\sigma_{\hat{\calD}_{out}}[\hat{\calR}_{\calD^\ast}]$ proportional to the standard deviation of the finite sample estimator $\hat{\calR}_{\calD^\ast}$ is incorporated\footnote{We choose $k\coloneqq 2$ based on the one-sided Cantelli-Bienaym{\'e}-Chebychev inequality.
\ref{sec: variance of estimators} gives the estimate of the standard deviation $\sigma_{\hat{\calD}_{out}}[\hat{\calR}_{\calD^\ast}]$}.\\

\textit{Hold-out risk}. Alternatively, let us take the held-out data as a benchmark in itself, instead of reweighting the risk to match the true population statistics. On the other hand, if the apparent hold-out prevalence differs from the true prevalence, we allow the prediction $q_{x_\ast,X,Y}(y_\ast)$ to make use of this knowledge. 
This yields a hold-out risk:
\begin{equation}
-\hat{\calR}_{HO}[q_{x_\ast,X,Y}] = \text{avg}_{(x_\ast,y_\ast)\in\hat{\calD}_{out}}\left[\log{q_{x_\ast,X,Y}(y_\ast)}\right] \, .
\label{eq: prevalence corrected hold-out risk}
\end{equation}
For instance in the proposed Bayesian approach under MAP inference, predictions take the form $q_{x_\ast,X,Y}(y_\ast)=p_{s_\ast=1}(y_\ast|x_\ast,\hat{w})$ with $\hat{w}\coloneqq\hat{w}_{MAP}(X,Y)$. See Eq.~\eqref{eq: likelihood with selection} or~\ref{sec: test-time selection} for the computation of the quantity.\\

\textit{Hold-out log-likelihood}. In the same spirit, when the apparent hold-out prevalence differs from the true prevalence, the expected log-likelihood on the held-out set resolves to an alternative form, see section~\ref{sec: test-time} and~\ref{sec: test-time selection}. Dropping constants of $\hat{w}$ we get the hold-out log-likelihood:
\begin{equation}
\text{LL}_{HO}[\hat{w}] = \text{avg}_{(x_\ast,y_\ast)\in\hat{\calD}_{out}}\left[\log{\frac{p(y_{\ast}|x_{\ast},\hat{w})}{p(y_{\ast}|\hat{w})}}\right] \, ,
\label{eq: prevalence corrected hold-out LL}
\end{equation}
which reads as an average information gain. Using the marginal computation framework of section~\ref{sec: marginal computations}, which applies regardless of the architecture and of the training loss, Eq.~\eqref{eq: prevalence corrected hold-out LL} can always be computed (including for models that do not use the proposed Bayesian approach to bias correction).

\subsection{Summary statistics derived from the confusion matrix}
\label{sec: confusion matrix}

The topic is addressed thoroughly in the literature \eg,~\cite{powers2011evaluation}. 
Let $\hat{y}_n$ (resp $t_n$) the predicted (resp. real) label for sample $n$. The \textit{accuracy} is perhaps the most widely acknowledged scalar summary statistics of classification performance. Its most striking limitation appears in settings with large class imbalance, where the score becomes overwhelmed by a single class with large prevalence (\eg for rare diseases, always predicting healthy leads to close to optimal accuracy). To address this defect one turns towards prevalence invariant statistics, namely the \textit{true class rates} $TR_y\triangleq (\sum_{n:t_n=y} \delta_{t_n}(\hat{y}_n))/(\sum_{n:t_n=y} 1)$, letting $\delta_{y}(y')$ be $1$ if $y'=y$, $0$ otherwise. 
The class rates describe how likely one is to receive a correct diagnostic ($\hat{y}$) for a known condition ($t$). 

Because they are invariant to prevalence, class rates only give a partial perspective. Indeed one may ask how likely they are, if diagnosed with $\hat{y}$, to actually have $t=\hat{y}$. This leads to the prevalence dependent statistics $PV_y\triangleq (\sum_{n:\hat{y}_n=y} \delta_{t_n}(\hat{y}_n))/(\sum_{n:\hat{y}_n=y} 1)$ known as \textit{predictive values}. 

For a summary we turn to a pair of statistics that encapsulate both aspects: the \textit{informedness} $I$ and the \textit{markedness} $M$, given in the binary case $\calY=\{0,1\}$ by:
\begin{equation}
I\triangleq \left(\sum_{y\in\calY} {TR}_y\right) - 1\, , \quad M\triangleq  \left(\sum_{y\in\calY} {PV}_y\right) - 1\, .
\label{eq: informedness and markedness}
\end{equation}
Unlike accuracy, $I$ and $M$ can be written symmetrically w.r.t.~the gain of a correct prediction and the cost of a mistake: $I\eq TPR-FPR$ and $M\eq PPV-(1-NPV)$. In that sense the informedness says how informed the predictor $\hat{y}$ is by the condition $t$, compared to chance. The markedness says how marked the condition $t$ is by the predictor $\hat{y}$ compared to chance. Chance level is $I=0$ (resp. $M\eq 0$)\footnote{This interpretation is retained in the multiclass extension ($|\calY|\!>\!2$), see \cite{powers2011evaluation}, but 
informedness is no longer prevalence independent.}, with $I, M\leq 1$. In the binary case informedness and markedness summarize predictive performance without skew towards any one class. 
The informedness $I$ and the \textit{balanced accuracy} $BA$ are equivalent up to renormalization. The pair $(TPR, FPR)$ along with knowledge of the prevalence and of the total count, is sufficient to rebuild the confusion matrix.

Finally the Matthews Correlation Coefficient $MCC$ is sometimes advocated as an informative scalar summary of predictive performance~\cite{chicco2020advantages}. In the binary case one shows that $MCC=\pm\sqrt(I\cdot M)$ is the geometric mean of informedness and markedness~\cite{powers2011evaluation} (positive above chance level and negative below). 

\subsection{ROC curve, AUC score and IM curve}
\label{sec: summary curves}

The remaining metrics evaluate the whole family of classifiers obtained by moving the cut-off point deciding label assignments, instead of just the argmax choice $\hat{y}\triangleq \argmax_y p(y|x,\hat{w})$. PR and ROC curves~\cite{hanley1982meaning} give a visual summary thereof. In the binary case, ROC curves plot the TPR as a function of FPR (Fig.~\ref{fig: IM curve 101}(A)). This is a complete summary via prevalence invariant quantities. 
The ROC curve helps identify operating points that achieve optimal trade-off for a specified cost of type I and type II errors. 
The Area Under the Curve (AUC) is also an informative summary scalar. As a drawback, ROC curves do not visually express much about predictive values, which vary significantly with the operating point. 

To visualize the impact of prevalence on the predictive power, one can also plot Informedness-Markedness (IM) curves, which convey both prevalence insensitive information ($x$-axis: $I$), and prevalence-based context ($y$-axis: M). IM curves allow to visualize the sensitivity of the model, at different operating points, to a change of the true prevalence (Fig.~\ref{fig: IM curve 101}). The change in profile shows that no operating point is ideal under all values of the prevalence.

\begin{figure*}[t]
\includegraphics[width=\textwidth]{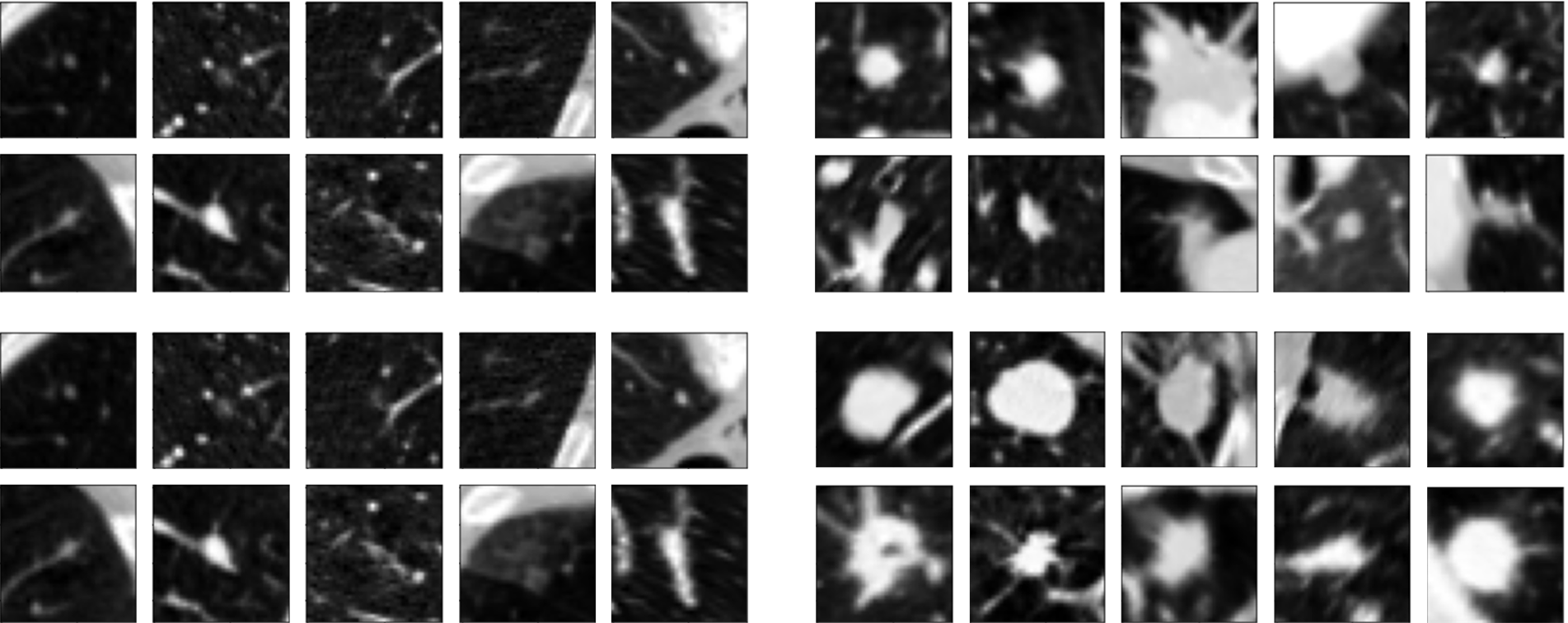}
\caption{Which nodules are benign, which ones are malignant? Each vignette is a $2$D slice from a low-dose CT scan of the lungs, cropped to a small square region centered on a nodule. The top two \textit{rows} correspond to benign nodules, the bottom two to malignant nodules.}
\label{fig: nodule examples}
\end{figure*}

\section{Deep learning case study: nodule malignancy prediction from CT imaging}
\label{sec: deep learning case study}

\textbf{LIDC-IDRI dataset.} The Lung Image Database Consortium and Image Database Resource Initiative (LIDC/IDRI)~\cite{armato2011lung} data includes over a thousand scans with one or more pinpointed nodules and corresponding annotations by multiple raters (typically $3$, $4$). The subjective malignancy score ranges from 1 (benign) to 5 (malignant), with $3$ indicating high uncertainty from the raters (in what follows it is used as a malignancy threshold when binarized labels are expected). 
Experiments consist of predicting the nodule malignancy from small patches (Fig.~\ref{fig: nodule examples}) extracted around the nodules (\eg, $64\times 64 \times 64$ $mm^3$). We consider two variants of the task, to binary classification (benign/malignant) and multilabel classification (subjective rating prediction). We use two variants of the dataset: (1) a dataset of $1407$ patches with binarized ground truth and a class imbalance of one to three in favor of benign nodules ($1065$ benign, $342$ malignant), for which nodules with an average rating of $3$ are excluded; and~(2) a dataset of $1086$ patches (marginal label distributions $\sim 0.075,0.2,0.45,0.2,0.075$) for which the raters' votes serve as a fuzzy ground truth.\\

\textbf{Architectures.} Two variants of the deep learning architectures described in \cite{ciompi2017towards} are used. Triplets of orthogonal viewplanes (dimension $32\times 32$) are extracted at random from the $3$D patch, yielding a collection of $s$ views (here, $s=9$). Each $2$D view is passed through a singleview architecture to extract an $m$-dimensional (\eg, $m=256$) feature vector, with shared weights across views. The features are then pooled (min, max, avg elementwise) to derive an $m$-dimensional feature vector for the stack of views. Finally a fully-connected layer outputs logits that are fed to a link function (\eg, softmax likelihood). 

Because of the relatively small size of the datasets, we use low-level visual layers pretrained on vgg16 (we retain the two first conv+relu blocks of the pretrained model, and convert the first block to operate on grayscale images). The low-level visual module returns a $64$-channel output image for any input view, which is then fed to the main singleview model. 

In the first variant (ConvNet), the main singleview architecture consists in a series of $2$D strided convolutional layers (stride $2$, replacing the pooling layers in \cite{ciompi2017towards}), with ReLu activations and dropout ($p=0.1$). In the second variant, convolutional layers are replaced with inception blocks. 

We observed similar trends across various architectures, from single-view fully convolutional classifiers to multi-view, multiresolution attention models. The two selected models are a trade-off between speed of experimentation for $k$-fold cross-validation and performance.\\

\subsection{Binary classification (benign vs.~malignant)}

\begin{table*}\centering
\ra{.8}
\small
\tabcolsep=0.14cm
\begin{tabular*}{0.9\textwidth}{rcccccccccccc}\toprule

& NELL & $\text{NLL}_{HO}$ & $\text{Rsk}_{HO}$ & $\text{Acc}_{HO}$ & {TNR}& {TPR} & $\text{NPV}_{HO}$& $\text{PPV}_{HO}$ & AUC & I & $\text{M}_{HO}$ & $\text{MCC}_{HO}$\\ \midrule
$\text{prev.}=0.5$ \\	
{IG} & $\mathbf{0.494}$ & $\mathbf{0.391}$ & $\mathbf{-0.325}$ & $\mathbf{0.86}$ & $\mathbf{0.88}$ & $0.78$ & ${0.92}$ & $\mathbf{0.68}$ & $\mathbf{0.89}$& $\mathbf{0.66}$ & $\mathbf{0.60}$ & $\mathbf{0.63}$ \\ 
{IW} & ${0.516}$ & ${0.418}$ & $-0.278$ & ${0.85}$ & ${0.86}$& $\mathbf{0.80}$ & $\mathbf{0.93}$ & $0.65$ & $\mathbf{0.89}$ & $\mathbf{0.66}$ & ${0.58}$ & ${0.62}$ \\[0.2cm]
$\text{prev.}=0.25$ \\	
{IG} & ${0.400}$ & ${0.416}$ & $\mathbf{-0.217}$ & ${0.84}$ & $0.85$ & $\mathbf{0.80}$ & $\mathbf{0.93}$ & ${0.64}$ & $\mathbf{0.89}$ & $\mathbf{0.66}$ & $0.57$ & $0.61$ \\ 
{IW} & $\mathbf{0.399}$ & $\mathbf{0.338}$ & $\mathbf{-0.217}$ & $\mathbf{0.87}$ & $\mathbf{0.93}$ & ${0.68}$ & $0.90$ & $\mathbf{0.78}$& $\mathbf{0.89}$ & ${0.62}$ & $\mathbf{0.68}$ & $\mathbf{0.65}$ \\[0.2cm]
$\text{prev.}=0.1$ \\	
{IG} & ${0.270}$ & ${0.515}$ & $\mathbf{-0.301}$ & ${0.80}$ & $0.78$ & $\mathbf{0.86}$ & $\mathbf{0.95}$ & ${0.56}$& $\mathbf{0.89}$& $\mathbf{0.64}$ & $0.51$ & $0.57$ \\ 
{IW} & $\mathbf{0.249}$ & $\mathbf{0.359}$ & $-0.272$ & $\mathbf{0.86}$ & $\mathbf{0.97}$ & ${0.52}$ & $0.86$ & $\mathbf{0.84}$ & $0.88$ & $0.48$ & $\mathbf{0.70}$ & $\mathbf{0.58}$ \\[0.2cm]
$\text{prev.}=0.01$ \\	
{IG} & ${0.085}$ & $\mathbf{0.565}$ & $\mathbf{-0.561}$ & $\mathbf{0.78}$ & $0.75$ & $\mathbf{0.87}$ & $\mathbf{0.95}$ & $0.53$ & $\mathbf{0.89}$ & $\mathbf{0.62}$ & $0.47$ & $\mathbf{0.54}$ \\ 
{IW} & $\mathbf{0.054}$ & $0.651$ & $-0.465$ & $\mathbf{0.78}$ & $\mathbf{0.99}$ & ${0.12}$ & $0.78$ & $\mathbf{0.94}$ & ${0.87}$ & ${0.12}$ & $\mathbf{0.72}$ & ${0.29}$ \\[0.2cm]
$\text{prev.}=0.001$ \\	
{IG} & ${0.058}$ & $\mathbf{0.604}$ & $\mathbf{-0.651}$ & ${0.75}$ & ${0.71}$ & $\mathbf{0.87}$ & $\mathbf{0.95}$ & $\mathbf{0.50}$ & $\mathbf{0.88}$ & $\mathbf{0.59}$ & $\mathbf{0.44}$ & $\mathbf{0.51}$ \\ 
{IW} & $\mathbf{0.008}$ & $1.138$ & $-0.527$ & $\mathbf{0.76}$ & $\mathbf{1.0}$ & $0.0$ & $0.76$ & NaN & 0.86 & $0.0$ & NaN & NaN\\

\bottomrule
\end{tabular*}
\label{table: binary deep learning -- performance}
\caption{Performance summary on held-out data for the Bayesian approach to prevalence-bias (IG) vs.~importance weighted log-loss (IW), averaged across the three folds. NELL: negative expected log-likelihood estimate for the true population. $\text{NLL}_{HO}$: negative hold-out log-likelihood. $\text{Rsk}_{HO}$: hold-out risk. $\text{Acc}_{HO}$: accuracy. TPR / NPR: positive / negative rates. $\text{PPV}_{HO}$ / $\text{NPV}_{HO}$: positive / negative predictive values. AUC: Area Under the (ROC) Curve. I / M: informedness / markedness. MCC: Matthews Correlation Coefficient.}
\end{table*} 

We predict a binary label corresponding to benign or malignant nodules. The $3$-fold experiment is repeated at five assumed true prevalences, \ie for a true malignancy probability of $0.5$, $0.25$, $0.1$, $0.01$ or $0.001$, and for two different architectures. Table~\ref{table: binary deep learning -- performance} reports all performance metrics averaged across all three folds in the $3$-fold cross validation, for the ConvNet archictecture. Note that each minibatch during training is balanced, \ie data samples are sampled equally among benign and malignant nodules (sampling with rebalancing). \ref{sec: additional results} reports similar results for the InceptionNet, and when training with minibatches sampled uniformly in the training dataset (sampling without rebalancing). 

\begin{figure}[t]
\includegraphics[width=\columnwidth]{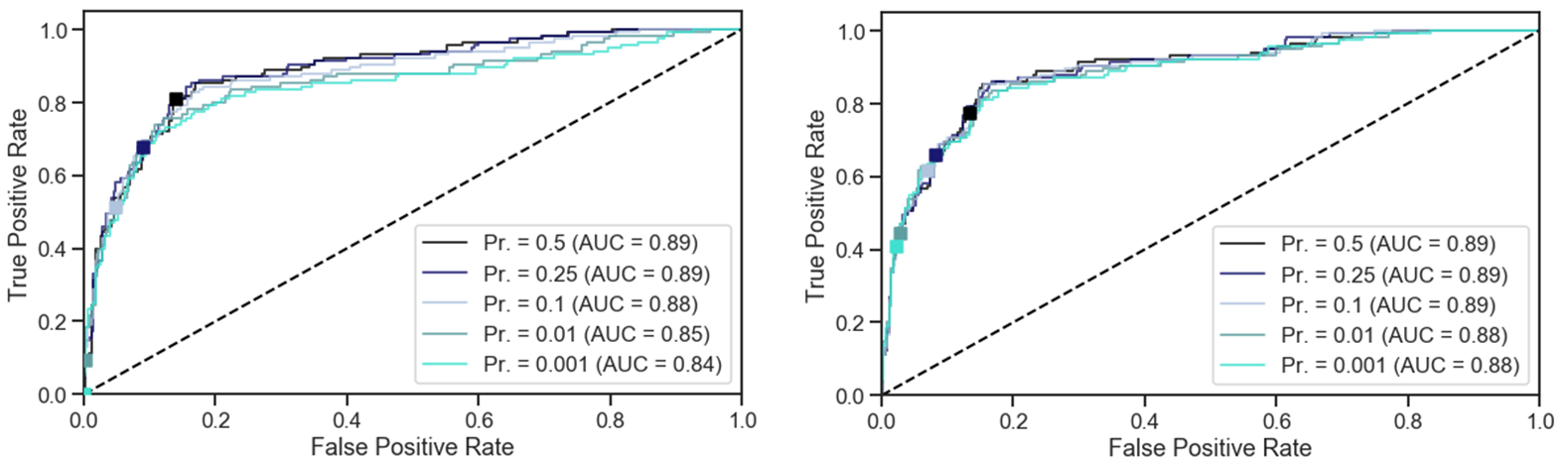}
\caption{ROC curves, for the importance weighted log-loss (left) vs. the Bayesian approach to prevalence bias (right). For each curve, the natural operating point (malignancy probability threshold of $0.5$) is marked with a square.} \label{fig: deep learning - binary roc curves}
\end{figure}

For highly imbalanced prevalences, the prediction of the importance weighted scheme becomes trivial (always benign), which is reflected in all metrics. The Bayesian scheme remains well calibrated even in this scenario. Fig.~\ref{fig: deep learning - binary roc curves} shows that despite a slight drop in performance at imbalanced prevalences, the importance weighted scheme is not necessarily poor. However it becomes poorly calibrated so that the probabilistic predictions can not be used, unlike for the Bayesian scheme. In other words, the importance weighted scheme is unable to cope with the effects of prevalence on prediction.

\subsection{Subjective rating prediction (multiclass)}

We predict a malignancy score ranging between $1$ and $5$. There are three experiments with different values for the assumed true prevalence of each label. In the first experiment, the observed dataset prevalence is taken as the true prevalence for each label ($\sim 0.075,0.2,0.45,0.2,0.075$). In the second experiment, a prevalence of $0.9$ is assigned to the label $y\eq 1$, the remaining $0.1$ being spread equally across the remaining labels. The last experiment proceeds in the same manner, but with a prevalence of $0.999$ assigned to $y\eq 1$. 

\begin{figure}[t]
\includegraphics[width=\columnwidth]{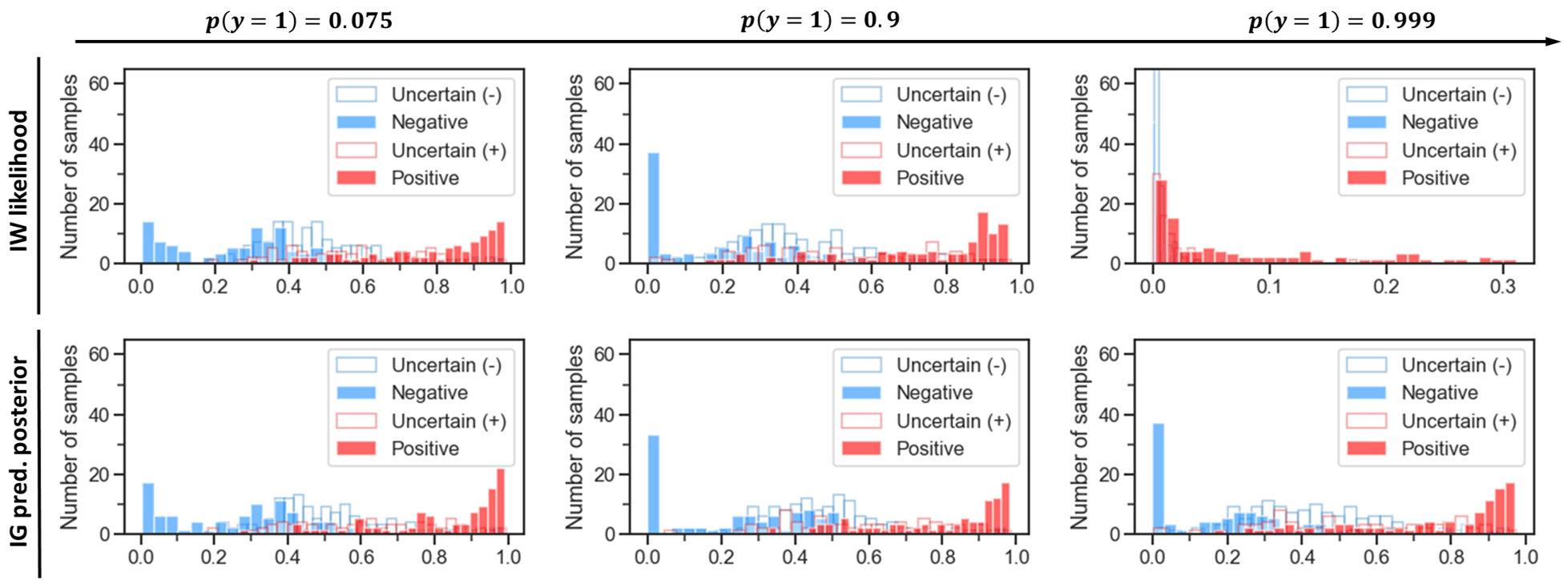}
\caption{Calibration of the prediction rule for different values of the true prevalence, for the Bayesian approach to prevalence bias (IG, bottom row) vs. the importance weighted log-loss (IW, top row). The plots report the predicted probability of malignancy for each sample (benign $\equiv$ negative, malignant $\equiv$ positive) in the test dataset. Left to right: varying the assumed true prevalence.} \label{fig: deep learning - multiclass binarized histogram}
\end{figure}

Because the malignancy scores are fundamentally subjective, part of the evaluation uses binarized labels instead, benign vs.~malignant. A score below $3$ translates to a benign nodule whereas a score above $3$ is malignant. Since the ground truth is fuzzy, with multiple annotators potentially giving different scores, the binarized ground truth is itself fuzzy. For each datum, votes are first translated into a fuzzy vote frequency for each label between $1$ and $5$, with frequencies summing to $1$ across labels. For instance, a data point with $3$ votes for label $2$ and $2$ votes for label $3$ translates to the fuzzy uplet $(0, 0.6, 0.4, 0, 0)$. The uplet is binarized to (benign, malignant)~frequencies by aggregating vote frequencies on each side of the threshold label $3$ (possibly summing below $1$), for instance $(0.6,0)$ in the previous case. Data points with more than $0.5$ of frequency for the label $3$ are called uncertain. Fig.~\ref{fig: deep learning - multiclass binarized histogram} reports the histogram of predicted malignancy probabilities across the test dataset for one of the three folds, distinguishing between true malignant (for which more than $0.5$ of target probability is assigned to labels $4$ and $5$), true benign (more than $0.5$ of target probability assigned to labels $1$ and $2$), uncertain benign and uncertain malignant. The behaviours of the importance weighted and Bayesian schemes are similar when the prevalence in the test dataset corresponds to the true prevalence (Fig.~\ref{fig: deep learning - multiclass binarized histogram}, first column), but they differ drastically when the assumed true prevalence of the label $1$ is gradually increased to $1$. The Bayesian scheme remains well calibrated unlike the importance weighted scheme. Table~\ref{table: multiclass deep learning -- performance} reports all performance metrics averaged across all three folds in the $3$-fold cross validation. The first five metrics are computed over the multilabel ground truth and the remaining seven over the binarized ground truth. The Bayesian scheme becomes clearly superior to the importance weighted scheme across all metrics when the dataset prevalence is heavily imbalanced compared to the assumed true prevalence. 

\begin{table*}\centering
\ra{.8}
\small
\tabcolsep=0.14cm
\begin{tabular*}{0.9\textwidth}{rcccccccccccc}\toprule

& NELL & $\text{NLL}_{HO}$ & $\text{Rsk}_{HO}$ & $\text{OOA}_{HO}$ & $\text{Acc}_{HO}$ & {TNR}& {TPR} & $\text{NPV}_{HO}$& $\text{PPV}_{HO}$ & AUC & I & $\text{M}_{HO}$\\ \midrule
$p(y=1)=0.075$ \\	
{IG} & $\mathbf{1.079}$ & $1.092$ & $\mathbf{-0.380}$ & $0.96$ & $0.86$ & $0.82$ & $\mathbf{0.89}$ & $\mathbf{0.90}$ & $0.82$& $\mathbf{0.94}$ & $0.71$ & $0.72$ \\ 
{IW} & ${1.093}$ & $\mathbf{1.029}$ & $-0.364$ & $\mathbf{0.97}$ & $\mathbf{0.87}$& $\mathbf{0.85}$ & $0.88$ & $0.89$& $\mathbf{0.85}$ & $\mathbf{0.94}$ & $\mathbf{0.73}$ & $\mathbf{0.74}$ \\[0.2cm]
$p(y=1)=0.9$ \\	
{IG} & ${0.461}$ & $\mathbf{1.132}$ & $\mathbf{-2.157}$ & $\mathbf{0.94}$ & $0.81$ & $0.77$ & $\mathbf{0.87}$& $\mathbf{0.88}$& $0.79$& $\mathbf{0.93}$ & $0.64$ & $0.67$ \\ 
{IW} & $\mathbf{0.436}$ & $1.536$ & $-1.793$ & $0.71$ & $\mathbf{0.85}$ & $\mathbf{0.92}$ & $0.78$ & $0.82$& $\mathbf{0.90}$ & $\mathbf{0.93}$ & $\mathbf{0.70}$ & $\mathbf{0.72}$ \\[0.2cm]
$p(y=1)=0.999$ \\	
{IG} & ${0.303}$ & $\mathbf{1.456}$ & $\mathbf{-4.192}$ & $\mathbf{0.85}$ & $\mathbf{0.81}$ & $0.83$ & $\mathbf{0.80}$ & $\mathbf{0.83}$ & $\mathbf{0.83}$ & $\mathbf{0.92}$ & $\mathbf{0.63}$ & $\mathbf{0.66}$ \\ 
{IW} & $\mathbf{0.020}$ & $5.195$ & $-2.325$ & $0.31$ & $0.56$ & $\mathbf{0.99}$ & $0.07$ & $0.55$ & NaN & $0.87$ & $0.06$ & NaN\\

\bottomrule
\end{tabular*}
\label{table: multiclass deep learning -- performance}
\caption{Performance summary on held-out data for the Bayesian approach to prevalence-bias (IG) vs.~importance weighted log-loss (IW), averaged across the three folds, for three assumed true prevalence levels. The prevalence $p(y=1)=0.075$ coincides with the dataset prevalence. NELL: negative expected log-likelihood estimate for the true population. $\text{NLL}_{HO}$: negative hold-out log-likelihood. $\text{Rsk}_{HO}$: hold-out risk. $\text{OOA}_{HO}$: one-off hold-out accuracy, counting the prediction as true if the predicted label is no more than one off from the true label (\eg, predicted label $2$ for a ground truth label of $3$). Values are computed over the multilabel ground truth for these metrics, as opposed to values for the following metrics that are derived from the binarized ground truth. $\text{Acc}_{HO}$: accuracy. TPR / NPR: positive / negative rates. $\text{PPV}_{HO}$ / $\text{NPV}_{HO}$: positive / negative predictive values. AUC: Area Under the (ROC) Curve. I / M: informedness / markedness.}
\end{table*}
\section{Discussion and conclusion}
\label{sec: discussion}

\subsection{Is the approach scalable?}

Yes. 
The closed-form derivations result in a 
computationally inexpensive implementation. The improbable prospect of marginalizing out an \textit{arbitrarily high-dimensional} input $x$ in a \textit{backpropable manner} w.r.t.~an \textit{arbitrarily high-dimensional} space of parameters $w$ is reduced to jointly training a logistic regressor, 
and backpropagating via a small pre-implemented custom backward routine. 

\subsection{What if the true prevalence is unknown?}

It is unlikely that this can be circumvented in principle. 
The true prevalence $p_\calY(y) \eq \int p(y|x,w^\ast)p_\calX(x)dx$ depends on two factors: the association $p(y|x,w^\ast)$ between explanatory factors and outcomes, and the distribution $p_\calX(x)$. In the prevalence bias scenario, both the apparent label distribution $\tilde{p}(y)$ and the resulting apparent input distribution $\tilde{p}_\calX(x)\eq \int p(x|y,w^\ast)\tilde{p}_\calY(y)dy$ are biased. If the true prevalence and the true input distribution are unknown, optimal prediction in the sense of this work may well be impossible, as key information is missing. 

If one has knowledge about $p_\calX(x)$ rather than $p_\calY(y)$, it can be used to recover an estimate of the prevalence $p_\calY(y)$ via the relationship $p_\calX(x)\eq \mathbb{E}_{p_\calY(y)}[p_{\calD'}(x|y)]$ (see~\cite{borgwardt2006integrating,huang2007correcting}); or to get an estimator for the marginal $p(y|w)\eq \mathbb{E}_{p_\calX(x)}[p(y|x,w)]$. 

On the other hand, it is always possible to train several models that reflect weak assumptions about the real prevalence. Say, the prevalence is known to be no more than $1$ in $10$. One would train models for $p_\calY(y\eq 1)\coloneqq 0.1, 0.01$, etc.~to get a sense of the sensitivity or robustness under varying prevalences. Ensembling schemes can also be designed to combine competing models.

\subsection{Causal vs.~structural dependencies, sample selection}

How can the direction of causality (Fig.~\ref{fig: graphical models}) be apparently reversed for training data compared to the test-time population? 
The customary causal interpretation of the {generative model} (\ie $A \rightarrow B$ means that $A$ causes $B$) implicitly refers to causation \textit{with regard to the sampling process}, which may be at odds with the natural causal intuition that ``the explanatory factors cause the outcome''. Fig.~\ref{fig: graphical models}(b) merely expresses that training data was sampled controlling the outcome, in anticausal fashion. A subtlety that cannot be fully developed here, is that selection mechanisms (when acting on child variables) make it possible for the causal graph and the structural graph to be irreconcilable. Since Bayesian inference directly relies on the \textit{structural} graph, we have favoured structural semantics. 
Of course causality, in particular the distinction between causation and association, plays a key role to assess the scope and applicability of the model as real-world circumstances change. Pragmatically though, causal insights are application-specific, sometimes subjective and therefore beyond the scope of this paper. 
Rather the paper offers a complete mathematical and computational solution for inference and prediction in presence of sampling bias under the assumption of label-dependent selection. 

\begin{figure}
\centering
\includegraphics[width=\columnwidth]{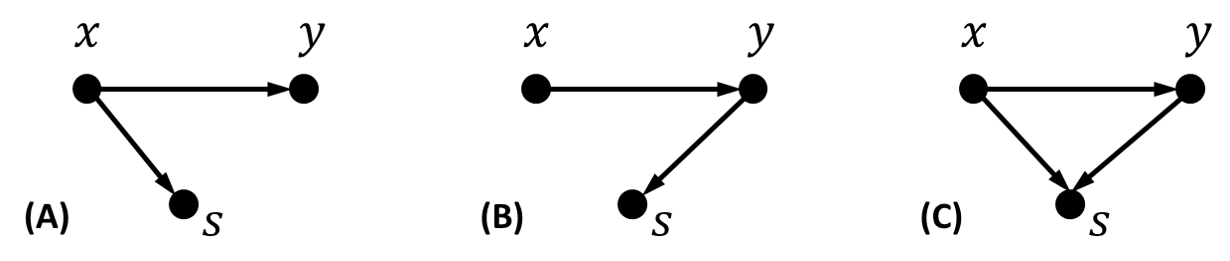}
\caption{Generative models of sample selection: (a) feature-based selection, (b) outcome-based selection, (c) both. We de-emphasize whether variables are observed or not, as well as keeping the dependency of $y$ on $w$ implicit.}
\label{fig: sample selection models}
\end{figure}

Further insights can be gained from the point of view of sample selection. Following \eg,~\cite{storkey2009training} 
the sample selection mechanism can be made manifest via a selection variable $s$, which acts as a filter. Draws from the true population are accepted (resp. rejected) if $s\eq 1$ (resp. $s\eq 0$). The acceptance-reject mechanism decides whether the sample can ever be observed ($s=1$) or not ($s=0$). This may for instance model hospital admission, in which case one can think of $s$ in terms of symptoms. The selection process may be conditioned on $x$, $y$ or both (Fig.~\ref{fig: sample selection models}). 
The training data is collected among observable subjects by sampling from $p(x,y|\, s\eq1)$. The general population is drawn from $p(x,y)$ without conditioning on the selection variable. 
Under model~Fig.~\ref{fig: sample selection models}(B), one retrieves label-dependent sampling bias as discussed in the paper (\ref{sec: Equivalence sample selection -- prevalence bias}). Sticking to the hospital analogy, one may also want to use the learnt model to do predictions on subjects admitted to the hospital rather than on the general population (\ref{sec: test-time selection}). 

In contrast, Fig.~\ref{fig: sample selection models}(A) closely relates to prospective studies and public health policy, where (say) one investigates the effect of life style ($x$) on some outcome ($y$). The sample distribution $p_{\calD'}(x)$ can differ from the general population (the population distribution $p(x)$ itself may shift due to public health campaigns). As the marginal distribution of $y$ depends on $p(x)$, the covariate \textit{bias} trickles down as prevalence \textit{shift}. Nevertheless a quick inspection of structural dependencies shows that this scenario is treated identically to the bias-free case in the Bayesian paradigm\footnote{A frequentist formalism may suggest a different estimator through importance weighting, see \eg~\cite{huang2007correcting}.}. 

To sum up, one decides whether it is more accurate to model their training set as generated while controlling outcomes ($y$) or covariates ($x$). In the former case, the present work applies. Of course the Bayesian Information Gain estimator may find practical extensions to settings where the generative assumptions are partly violated.

\section*{Acknowledgements}

This research has received funding from the European Research Council (ERC) under the European Union's Horizon 2020 research and innovation programme (grant agreement No 757173, project MIRA, ERC-2017-STG). LL is funded through the EPSRC (EP/P023509/1).

\bibliographystyle{model2-names.bst}\biboptions{authoryear}
\interlinepenalty=10000
\bibliography{refs}

\appendix
\section{Generalized minibatch estimates of the loss}
\label{sec: generalized minibatch estimates}

Consider a contribution of the training data to the overall loss of the form $\calL_{data}(X,Y)\triangleq \sum_{n\leq N} \calL_w(x_n,y_n)$, \ie a sum of individual sample contributions for all points in the full training dataset $F$. If a minibatch $B$ of size $n_B$ is drawn i.i.d.~from $F$, the sum of sample contributions $\sum_{n\in B} \calL_w(x_n,y_n)$ for all points in the minibatch is unbiased, up to a renormalization $N/n_B$. However if the statistics of the minibatch differ from those of the training data, we should account for the discrepancy so as not to risk counting some points several times on average over an epoch. 
Specifically we allow for arbitrary policies to choose label counts $n_B(y)$ in the minibatch (rather than assuming that they are drawn uniformly at random), after which data points are sampled uniformly at random given the label.
In that setting the overall contributions should be aggregated and corrected on a per-class basis to retrieve an unbiased estimate of the data loss on $F$. This yields:
\begin{equation}
\calL_{data}(X,Y)\simeq \hat{\calL}_{data}(X,Y) \triangleq \frac{N}{n_B}\sum_{n\in B} \omega(y_n) \cdot \calL_w(x_n,y_n)\, ,
\label{eq: loss minibatch estimate -- appendix}
\end{equation}
where the corrective weights $\omega(y)$ can be set to either one of the two values $\omega^{(1)}(y)$ or $\omega^{(2)}(y)$ from Eq.~\eqref{eq: minibatch corrective weights}:
\begin{equation}
\omega^{(1)}(y) \coloneqq \frac{p_{F}(y)}{\tilde{p}(y)}\, , \quad 
\omega^{(2)}(y) \coloneqq \frac{n_B}{n_B(y)} p_{F}(y) \, .
\label{eq: minibatch corrective weights}
\end{equation}
$p_F(y)\triangleq N_F(y)/N$ is the empirical frequency of label $y$ in the full dataset $F$. $\omega^{(2)}(y)$ uses the empirical statistics $n_B(y)$ of the minibatch $B$. It is defined if these label counts $n_B(y)$ are almost surely all non-zero. $\omega^{(1)}$ instead uses the expected value of minibatch label frequencies averaged over the (possibly random) policy of label count selection, $\tilde{p}(y)\!\triangleq\! \mathbb{E}_{N_B}[n_B(y)/n_B]$, where $N_B\triangleq[n_B(y), y\!\in\!\calY]$ are the label counts (that sum to a total of $n_B$). $\omega^{(1)}(y)$ is defined under the relaxed assumption that all $\tilde{p}(y)$ are non-zero, at the cost that the resulting unbiased estimator potentially has more variance. 

These estimates mirror closely those put forward in~\ref{sec: Appendix -- minibatch estimates}, 
\ref{sec: variance of estimators}, where it is developed more at length. $\omega(y)$ plays a similar role to $\beta(y)$, simply replacing the target distribution $p_\calY(y)\leftarrow p_{F}(y)$. 
The tools developed in~\ref{sec: variance of estimators} can then be applied to study the variance of $\hat{\calL}_{data}(X,Y)$ as an estimator of $\calL_{data}(X,Y)$ and/or to design adaptive minibatch strategies based on variance minimizing minibatch estimates.

In the special case where $\calL_w(x,y)\coloneqq \calL_w^{BC}(x,y)$ is the bias-corrected loss, $\calL_w^{BC}(x,y)\triangleq\log{p(y|x,w)}-\log{p(y|w)}$, we can also expand Eq.~\eqref{eq: minibatch corrective weights} as:
\begin{equation}
\begin{split}
\hat{\calL}_{data}(X,Y) = \frac{N}{n_B}&\sum_{n\in B} \omega(y_n) \log{p(y_n|x_n,w)} \\
\, & - \sum_{y\in\calY} N_F(y) \log{p(y|w)} \, . 
\end{split}
\label{eq: bias-free minibatch estimate -- appendix}
\end{equation}
The estimate of Eq.~\eqref{eq: bias-free minibatch estimate -- appendix} is equivalent to Eq.~\eqref{eq: loss minibatch estimate -- appendix} for $\omega(y)\!\coloneqq\!\omega^{(2)}(y)$ (exploiting a simplification for the last term) but gives a slight variant with lesser variance for $\omega(y)\!\coloneqq\!\omega^{(1)}(y)$.

\section{Estimates of the marginal and its gradient}
\label{sec: Appendix -- minibatch estimates}

Noting that $x \perp\!\!\!\perp w$ for the true population generative model, we get:
\begin{align}
p(y'|w) &= \textstyle \int_x p(y'|x,w) p(x) dx \, , \\
\, &= \textstyle \int_{x} p(y'|x,w) \left(\int_{y} p(x|w',y)p(y|w') dy\right) dx \, , \label{eq: x ind w'} \\
\, &= \textstyle \int_{x} p(y'|x,w) \left(\int_{y} p(x|w^\ast,y)p(y|w^\ast) dy\right) dx \, , \label{eq: w' to wast} \\
\, &= \int_{y} p_\calY(y) \left(\int_x p(y'|x,w) \cdot p_{\calD'}(x|y) dx\right) dy \label{eq: final 2} \, , \\
\, &= \int_{x,y} \left(p(y'|x,w) \frac{p_\calY(y)}{\tilde{p}(y)}\right) \cdot p_{\calD'}(x,y)  d(x,y) \label{eq: final 1} \, .
\end{align} 
Eq.~\eqref{eq: x ind w'} holds for any value $w'$ by independence $x \perp\!\!\!\perp w$, and in particular for the true value $w_\ast$ (Eq.~\eqref{eq: w' to wast}). Eq.~\eqref{eq: final 2} uses $p_\calY(y)\!\triangleq\! p(y|w_\ast)$ and $p_{\calD'}(x|y)\!\triangleq\! p(x|w_\ast,y)$ from the generative model. For Eq.~\eqref{eq: final 1}, multiply and divide by $\tilde{p}(y)$ and note that $p_{\calD'}(x,y)\!\triangleq\! p_{\calD'}(x|y)\tilde{p}(y)$. Eq.~\eqref{eq: final 1}, Eq.~\eqref{eq: final 2} are turned into the empirical estimates of Eq.~\eqref{eq: empirical marginal estimate}, Eq.~\eqref{eq: corrective weights} from the main text, as follows. The notation $p_{\calD'}$ emphasizes the part that is approximated stochastically, using the minibatch sample distribution hence collapsing the integrals into finite sums. $p_{\calD'}(x|y)$ is replaced by $1/n_B(y)\cdot\sum_{n\in B:y_n\eq y} \delta_{x_n}(x)$, for Eq.~\eqref{eq: final 2} to become:
\begin{align}
p(y'|w) &\simeq \sum_{y\in\calY} \overbrace{p_\calY(y)\,\,\frac{1}{n_B(y)}}^{\triangleq \beta^{(2)}(y)/n_B}\,\cdot\sum_{\substack{n\in B:\\y_n= y}} p(y'|x_n,w)\, , \label{eq: final 1 -- discrete} \\
\, & = \frac{1}{n_B}\,\sum_{y\in\calY} \sum_{\substack{n\in B:\\y_n= y}} \,\, \underbrace{\beta^{(2)}(y)}_{\beta^{(2)}(y_n)} p(y'|x_n,w)\, ,
\end{align} 
and we finally merge the two sums into a single one over ${n\in B}$ to get the first estimator. Eq.~\eqref{eq: final 1 -- discrete} is valid whenever the minibatch has at least $1$ sample from each class. If instead we replace $p_{\calD'}(x,y)$ by $1/n_B\cdot\sum_{n\in B} \delta_{(x_n,y_n)}(x,y)$ in Eq.~\eqref{eq: final 1}, we get the other estimator:
\begin{equation}
p(y'|w)\simeq \frac{1}{n_B}\, \sum_{n\in B} \overbrace{\frac{p_\calY(y_n)}{\tilde{p}(y_n)}}^{\triangleq \beta^{(1)}(y_n)} p(y'|x_n,w)\, .
\end{equation} 

The distribution $\calD'$ of samples can vary from minibatch to minibatch without affecting the validity of the derivations, and the strategy outlined in the main text for the estimation of the marginal remains valid. 

To get an estimate of the gradient $\nabla_w \log{p(y|w)}$ from the minibatch, first note that :
\begin{equation}
p(y|w) \cdot \nabla_w \log{p(y|w)}\eq \nabla_w p(y|w)\, .
\label{eq: gradient of the log}
\end{equation} 
Plug the unbiased minibatch estimate $\nabla_w \hat{p}_B(y;w)$ in place of $\nabla_w p(y|w)$, and the auxiliary network estimate $q_\psi(w)(y)$ (section~\ref{sec: marginal computations},~\ref{sec: KL auxiliary training}) in place of $p(y|w)$ to get:
\begin{align}
\nabla_w \log{p(y|w)} &\simeq \frac{1}{q_\psi(w)(y)}\cdot \nabla_w \hat{p}_B(y;w) \, , \label{eq: gradient estimate proof -- a}\\
\, & = \frac{\hat{p}_B(y;w)}{q_\psi(w)(y)}\cdot\nabla_w \log{\hat{p}_B(y;w)}\, . \label{eq: gradient estimate proof -- b}
\end{align}
Eq.~\eqref{eq: gradient estimate proof -- a} circumvents a minibatch approximation through the $\log$ non-linearity: $\nabla_w \log{p(y|w)} \not\simeq \nabla_w \log{\hat{p}_B(y;w)}$. Notice the factor in front of the right-hand side in Eq.~\eqref{eq: gradient estimate proof -- b} which acts as a correction. $q_\psi(w)(y)$ is the output of the auxiliary network and is based on the whole dataset, whereas $\hat{p}_B(y;w)$ is a minibatch estimate. 

\subsection{Proof of unbiasedness}

By differentiating Eq.~\eqref{eq: final 1} or Eq.~\eqref{eq: final 2} one gets respectively:
\begin{align}
\nabla_w {p(y|w)} &=\mathbb{E}_{\calD'}\left[\beta^{(1)}(y)\nabla_w p(y'|x,w)\right]\, , \label{eq: final 1 -- differentiated} \\
\, &=\mathbb{E}_{y\sim p_\calY}\left[\mathbb{E}_{x|y}\left[\nabla_w p(y'|x,w)\right]\right] \, . \label{eq: final 2 -- differentiated}
\end{align}

Replacing the expectation(s) by the minibatch average gives unbiased estimators. 
From similar arguments as in the previous paragraph it leads exactly to 
\begin{equation}
\frac{1}{n_B}\sum_{n\in B}\beta(y_n)\cdot \nabla_w p(y'|x_n,w) =\nabla_w \hat{p}_B(y';w)\, ,
\end{equation} 
with $\beta=\beta^{(1)}$ when starting from Eq.~\eqref{eq: final 1 -- differentiated}, $\beta=\beta^{(2)}$ when starting from Eq.~\eqref{eq: final 2 -- differentiated}; and the equality with the RHS holds provided that $\hat{p}_B(y';w)$ is computed from $\beta^{(1)}$, resp. $\beta^{(2)}$. This justifies the claim that $\nabla_w \hat{p}_B(y';w)$ is an unbiased estimator of $\nabla_w p(y'|w)$. In turn this also justifies that $1/p(y|w)\cdot\nabla_w \hat{p}_B(y';w)$ is unbiased as an estimator of $\nabla_w \log{p(y|w)}$, looking back at Eq.~\eqref{eq: gradient of the log}. 

Eq.~\eqref{eq: gradient estimate proof -- a},\eqref{eq: gradient estimate proof -- b} are not unbiased but close to as soon as the auxiliary network approximation $q_\psi(w)(y')\!\simeq \! p(y'|w)$, which is informed by the full batch, is accurate. 

The gradient of the final loss function depends linearly on terms $\nabla_w \log{p(y'|w)}$, $y'\!\in\!\calY$. Therefore plugging any unbiased (or close to) estimate of these quantities (Eq.~\eqref{eq: gradient estimate proof -- b}) results in an unbiased estimate of the corresponding part in the gradient of the loss.

\subsection{Computational logic with automatic differentiation}
\label{sec: Computational logic with automatic differentiation}

The quantities $\log{\hat{p}_B(y';w)}$ are output via the numerically robust logsumexp trick. Recalling Eq.~\eqref{eq: empirical marginal estimate}, Eq.~\eqref{eq: corrective weights} and denoting $\gamma(y)\!\triangleq\! \log{(\beta(y)/n_B)}$ for convenience:
\begin{equation}
\log{\hat{p}_B(y';w)} = \text{logsumexp}_{n\in B}\left(\vphantom{\gamma(y_n)+\log{p(y'|x_n,w)}}\right.
\gamma(y_n)+\underbrace{\log{p(y'|x_n,w)}}_{\triangleq f_n(w)(y')}
\left. \vphantom{\gamma(y_n)+\log{p(y'|x_n,w)}}\right) \, . \label{eq: logsumexp trick}
\end{equation}

The $f_n(w)$ are the log-likelihoods of the samples. We assume they are readily available and backpropable through the computational graph. In vectorized form with $\bm{f}(w)\triangleq[f_n(w)(y), y\in\calY, n\in B]$ and $\bm{\hat{p}}_B(w)\triangleq [\hat{p}_B(y;w), y\in\calY]$:
\begin{align}
\log{\bm{\hat{p}}_B(w)} &=  \text{logsumexp}\left(\bm{\gamma}+\bm{f}(w)\right) \, , \label{eq: full batch implementation}\\
\, & \triangleq F(\bm{\gamma}, \bm{f}(w)) \, , \label{eq: log-marginal dependence on w}
\end{align}
where Eq.~\eqref{eq: log-marginal dependence on w} stresses how $\log{\bm{\hat{p}}_B}$ only depends on $w$ through $\bm{f}(w)$. For a full batch implementation when it fits in memory, one directly implements Eq.~\eqref{eq: full batch implementation} and backpropagates through it (without need for an auxiliary network). For a minibatch implementation, we make a few more comments. Firstly recall that the loss function depends linearly on $\log{\bm{p}_{|w}}\triangleq [\log{p(y|w)},y\in\calY]$, specifically via the contribution:
\begin{equation}
N_B^T \log{\bm{p}_{|w}} \, , \label{eq: true loss contribution}
\end{equation} 
where $N_B\triangleq [n_B(y), y\in\calY]$. Therefore the gradient of the loss w.r.t.~$\log{\bm{p}_{|w}}$ doesn't actually depend on the values $\log{\bm{p}_{|w}}$, so that any other value could in fact be passed. Secondly differentiating any one of the $|\calY|$ stacked components $F=[F_y, y\in\calY]$ in Eq.~\eqref{eq: log-marginal dependence on w} gives:
\begin{equation}
\nabla_w \log{{\hat{p}}_B(y;w)} = J_w\bm{f}^T \, \nabla_{\bm{f}} F_y \, , 
\end{equation} 
via the chain rule, where $J_w$ stands for the Jacobian. Therefore by linearity $\nabla_w (\bm{\alpha}\odot_y\log{\bm{\hat{p}}_B(w)}) = J_w\bm{f}^T \, (\bm{\alpha}\odot_y\nabla_{\bm{f}} F)$ for any constant $\bm{\alpha}\triangleq [\alpha(y), y\!\in\!\calY]$ of $w$, where $\odot_y$ denotes elementwise multiplication in $y$ (with broadcasting if necessary). In particular starting from Eq.~\eqref{eq: gradient estimate proof -- b}:
\begin{align}
\nabla_w \log{\bm{p}_{|w}} &\simeq \frac{\hat{\bm{p}}_B(w)}{q_\psi(w)}\odot_y\nabla_w \log{\bm{\hat{p}}_B(w)}\, , \\
\, & = \left[\nabla_w \left(\frac{\bm{\hat{p}}_B(w')}{q_\psi(w')} \odot_y\log{\bm{\hat{p}}_B(w)} \right)\right]_{w'\coloneqq w}\, , \label{eq: gradient for implementation 1}\\
\, & = \left[J_w\bm{f}^T \, \left(\frac{\bm{\hat{p}}_B(w')}{q_\psi(w')}\odot_y\nabla_{\bm{f}} F\right)\right]_{w'\coloneqq w}\, , \\
\, & = J_w\bm{f}^T \, \left(\frac{\bm{\hat{p}}_B(w)}{q_\psi(w)}\odot_y\nabla_{\bm{f}} F\right)\, . \label{eq: gradient for implementation 2}
\end{align}

Put together this suggests two implementations and we favoured the latter. Both approaches compute the main network output, \ie the log-likelihoods $\bm{f}(w)$; and the auxiliary network output, \ie the approximate log-marginals $\log{q_\psi(w)}$. Then compute $\log{\bm{\hat{p}}_B(w)}$ (Eq.~\eqref{eq: logsumexp trick}).\\

\textit{Implementation 1.} Compute $\bm{\alpha}\coloneqq {\bm{\hat{p}}_B(w)}/{q_\psi(w)}$ and detach $\bm{\alpha}$ from the computational graph. In the true loss contribution of Eq.~\eqref{eq: true loss contribution} replace $\log{\bm{p}_{|w}}$ by $\bm{\alpha} \odot_y\log{\bm{\hat{p}}_B(w)}$. The value of the loss will not be correct, but the backpropagated gradients will be correct, as per Eq.~\eqref{eq: gradient for implementation 1}.\\

\textit{Implementation 2.} Implement the log-marginal as a dummy autodifferentiable function with inputs $\log{\bm{\hat{p}}_B}$, $\log{q_\psi}$ and with a custom backward routine. The forward routine outputs $\log{q_\psi}$, which is fed to Eq.~\eqref{eq: true loss contribution}. The backward retrieves the output gradient $\bm{g}$, multiplies pointwise to get $\exp(\log{\bm{\hat{p}_B}}-\log{q_\psi}) \odot_y \bm{g}$, and passes this as the input gradient on input $\log{\bm{\hat{p}}_B}$. This is valid from Eq.~\eqref{eq: gradient for implementation 2}. \\

A slightly different approach 
can be followed based on an expansion of $\nabla_w \log{\bm{p}_{|w}}$ and Eq.~\eqref{eq: gradient estimate proof -- a} directly in term of the log-likelihoods $\bm{f}$:
\begin{equation}
\text{avg}_{n\in B} \left( \bm{\beta} \odot_n \exp{\left(\bm{f}-\log{q_\psi}\right)} \odot_{n,y} \nabla_w \bm{f} \right) \, ,
\end{equation}
where $\bm{\beta}\triangleq [\beta(y_n), n\in B]$ and broadcasting is implied whenever required. Exploiting the linearity of operators involved in the chain rule yields an approach similar to implementation 2. This numerical variant bypasses the (\textit{pytorch} optimized) logsumexp computation of $\log{\bm{\hat{p}}_B}$ from Eq.~\eqref{eq: logsumexp trick}, but the computational load of the (high-level \textit{pytorch}) custom backward is slightly higher and it does not exploit redundancies with Eq.~\eqref{eq: ELL - approximation 1}.\\

\textit{Implementation 3.} Implement an auto-differentiable function with inputs $\bm{f}$, $\log{q_\psi}$ and $\bm{\beta}$, and a custom backward routine. In the forward routine output $\log{q_\psi}$, which is fed to Eq.~\eqref{eq: true loss contribution}. The custom backward retrieves the output gradient $\bm{g}$, computes $\exp{(\bm{\gamma}+\bm{f}-\log{q_\psi})} \odot_y \bm{g}$ (using broadcasting conventions), and passes this as the input gradient on input log-likelihoods $\bm{f}$. \\

\section{Optimality of the predictive posterior and derivation of the posterior}
\label{sec: proofs}

The probabilistic decision rule $p(\cdot|x_\ast,X,Y)$ minimizes the Bayesian predictive risk $\calR_{\text{Bayes}}= \mathbb{E}_{w\sim p(w)}[\calR_{w}[q_{x_\ast,X,Y}]]$, with $\calR_{w}[q_{x_\ast,X,Y}]=\mathbb{E}_{(X,Y)}\mathbb{E}_{(x_\ast,y_\ast)}[\calE(q_{x_\ast,X,Y},y_\ast)]$ for the logarithmic loss function $\calE(q,y)\coloneqq-\log{q(y)}$. Indeed merging the triple expectation as a single expectation w.r.t.~the product distribution, $\calR_{\text{Bayes}}$ immediately rewrites as 
\begin{align}
\calR_{\text{Bayes}}&=-\mathbb{E}_{p(y_\ast,x_\ast,X,Y,w)}[\log{q_{x_\ast,X,Y}(y_\ast)}]\, , \\
&= -\mathbb{E}_{p(x_\ast,X,Y)}\left[\textstyle \int \log{q_{x_\ast,X,Y}(y_\ast)}\, p(y_\ast|x_\ast,X,Y)dy_\ast\right]\, , \label{eq: Bayes risk -- posterior introduction}\\
&= \mathbb{E}_{p(x_\ast,X,Y)}\left[\text{KL}[p(\cdot|x_\ast,X,Y)\Vert q_{x_\ast,X,Y}(\cdot)]\right] + \text{cst.}
\end{align}
The result $q_{x_\ast,X,Y}\eq p(\cdot|x_\ast,X,Y)$ follows from the properties of the Kullbach-Leibler divergence. Eq.~\eqref{eq: Bayes risk -- posterior introduction} introduces the posterior predictive distribution $p(y_\ast|x_\ast,X,Y)\triangleq \int_w p(y_\ast|w,x_\ast)p(w|X,Y) \, dw$. It relies in turn on the posterior distribution $p(w|X,Y)\triangleq p(Y|X,w)p(w)/p(X,Y)$ of parameters $w$, defined as soon as $p(X,Y)> 0$.\\

\noindent
\textbf{Actions and optimal policies.} In the general case, one looks for a decision policy $d_{X,Y}:x\in\calX\mapsto d_{X,Y}(x)\in\calA$ in some action space $\calA$, optimal w.r.t.~a loss $\calE:(a,y)\in\calA\times\calY\rightarrow \calE(a,y)\in\R$ for the Bayes risk $\calR_{\text{Bayes}}[d_{X,Y}]= \mathbb{E}_{w\sim p(w)}[\calR_{w}[d_{X,Y}]]$, where $\calR_{w}[d_{X,Y}]=\mathbb{E}_{(X,Y)}\mathbb{E}_{(x_\ast,y_\ast)}[\calE(d_{X,Y}(x_\ast),y_\ast)]$. 

By the same argument as above,
\begin{align}
\calR_{\text{Bayes}}&=\mathbb{E}_{p(y_\ast,x_\ast,X,Y,w)}\left[\calE(d_{X,Y}(x_\ast),y_\ast)\right]\, , \\
&= \mathbb{E}_{p(x_\ast,X,Y)}\left[\vphantom{\textstyle\int \calE(d_{X,Y}(x_\ast),y_\ast)}\right. 
\underbrace{\textstyle\int \calE(d_{X,Y}(x_\ast),y_\ast) p(y_\ast|x_\ast,X,Y)dy_\ast}_{\triangleq \calR[d_{X,Y}(x_\ast)]}
\left. \vphantom{\textstyle\int \calE(d_{X,Y}(x_\ast),y_\ast)}\right]\, ,
\end{align}
which decouples into (a sum/expectation of) separate minimization problems over the optimal $d_{X,Y}(x_\ast)\in\calA$ for each possible $X,Y,x_\ast$ with risk $\calR[d_{X,Y}(x_\ast)]$. Eq.~\eqref{eq: optimal policy} gives the optimal policy:
\begin{equation}
d_{X,Y}^\ast(x_\ast) = \argmin_{a\in\calA} \int_\calY \calE(a,y_\ast) p(y_\ast|x_\ast,X,Y)dy_\ast \, ,
\label{eq: optimal policy}
\end{equation}
which as announced can be computed given the knowledge of the predictive posterior $p(\cdot |x_\ast,X,Y)$.\\

\noindent
\textbf{Examples.} Let $\calA\eq\calY$ so that the decision rule $d_{X,Y}(x_\ast)\eq\hat{y}$ is about a choice of label. Let $\calE(\hat{y},y_\ast)\ccoloneqq\delta_{y_\ast}(\hat{y})$, meaning any error has an identical cost of $1$. Then, 
\begin{align}
d_{X,Y}^\ast(x_\ast) &=\argmin_{y} 1-p(y|x_\ast,X,Y)\, , \\
\, &= \argmax_{y} p(y|x_\ast,X,Y)\, .
\end{align}
Let $\calE(\hat{y},y_\ast)$ be arbitrary instead, but $\calY=\{0,1\}$ be a binary label set. Let $p_\ast\coloneqq p(y_\ast=1|x_\ast,X,Y)$ for short. Then,
\begin{align}
d_{X,Y}^\ast(x_\ast) & = \argmin_{\hat{y}} p\cdot\calE(\hat{y},1)+(1-p)\cdot\calE(\hat{y},0)\, , \\
\, & = \argmin_{\hat{y}} p\left(\calE(\hat{y},1)-\calE(\hat{y},0)\right) + \calE(\hat{y},0) \, ,
\end{align}
so that: 
\begin{equation}
d_{X,Y}^\ast(x_\ast)= \left\{\begin{array}{ll}
    1 &\text{if } p_\ast\geq \theta \\
    0 &\text{otherwise}
  \end{array} \right. \, ,
\end{equation}
with $1/\theta=1+\frac{\calE(0,1)-\calE(1,1)}{\calE(1,0)-\calE(0,0)}$. Without loss of generality, one may constrain $\calE(1,1)\coloneqq -\calE(0,1)$ and $\calE(0,0)\coloneqq-\calE(1,0)$, so that $1/\theta=1+{\calE(0,1)}/{\calE(1,0)}$. \\

\noindent
\textbf{Proof of Eq.~\eqref{eq: bias-corrected posterior}.} The posterior can be expressed as the ratio $p(w|X,Y)=p(X,Y,w)/p(X,Y)$  of the joint probability and evidence. The latter is a constant of $w$. The tilde notation denotes distributions under the generative model of training data. Rewriting the joint distribution we get:
\begin{align}
p(w|X,Y) &\propto \tilde{p}(X|Y,w)\tilde{p}(Y|w)\tilde{p}(w) \, , \\
\, &\propto \tilde{p}(X|Y,w)\tilde{p}(Y)p(w) \, , \label{eq: Y ind w} \\
\, &\propto p(w)\cdot \textstyle \prod_n \tilde{p}(x_n|y_n,w) \, , \label{eq: iid}\\
\, &\propto p(w)\cdot \textstyle \prod_n p(x_n|y_n,w) \, , \label{eq: same conditional}\\
\, &\propto p(w) \cdot \prod_n \frac{p(y_n|x_n,w)p(x_n)}{p(y_n|w)} \, , \label{eq: apparently not idiot proof} \\
\, &\propto p(w) \cdot \prod_n \frac{p(y_n|x_n,w)}{p(y_n|w)} \, .
\end{align}
Eq.~\eqref{eq: Y ind w} uses $\tilde{p}(w)\eq p(w)$ and the independence $Y\!\perp\!\!\!\perp \!w$ in 
Fig.~\ref{fig: graphical models}(b). Eq.~\eqref{eq: iid} follows by i.i.d., $x_n{\perp\!\!\!\perp} x_{-n},y_{-n}\enspace|w$. The conditional $\tilde{p}(x_n|y_n,w)$ is unchanged in the label-based sampling, equal to $p(x_n|y_n,w)$, hence Eq.~\eqref{eq: same conditional}. Eq.~\eqref{eq: apparently not idiot proof} results from the application of Bayes' rule and the independence $x\!\perp\!\!\!\perp \!w$ in the true population's generative model. The last line and Eq.~\eqref{eq: bias-corrected posterior} ensue after dropping the $p(x_n)$ as constants of $w$. From the above we also see that for variational inference, the ELBO and its various usual expressions still hold.

\section{Training the auxiliary network $q_\psi(w)$}
\label{sec: KL auxiliary training}

The aim is to approximate the marginal $p(y|w)$ via an auxiliary neural network $q_\psi: w\mapsto q_\psi(w)$ with trainable parameters $\psi$, where $q_\psi(w): y\in\calY \mapsto q_\psi(w)(y)$ assigns a probability to every outcome $y\in\calY$. Recall that the first-order dependency $\nabla_w \log{q_\psi(w)}$ between $w$ and $q_\psi(w)$ is \textit{never} used. The network is only expected to accurately approximate the $0$th order quantity $p(y|w)$ in a neighborhood of the current value $w$ of interest during the optimization process. 

For MAP or MLE inference of $w$, the estimate of $p(y|\hat{w})$ at the current value $\hat{w}$ of the parameters must be accurate. For variational inference, the estimate $q_\psi(w)$ of $p(\cdot|w)$ should be accurate in the range of probable values of the current variational posterior estimate $q(w)$. 

This is achieved by minimizing the Kullback-Leibler divergence $KL[p(\cdot|w)||q_\psi(w)]$ w.r.t.~$\psi$ during the optimization process, concurrently with the main network optimization:
\begin{align}
KL[p(\cdot|w)||q_\psi(w)] &= \int_{\calY} p(y|w) \log{\frac{p(y|w)}{q_\psi(w)(y)}} dy \, , \label{eq: KL divergence definition} \\
\, &= -\int_{\calY} p(y|w) \log{q_\psi(w)(y)} \, dy + \text{cst.} \, , \label{eq: ELL}
\end{align}
where the entropy of $p(\cdot | w)$ can be treated as a constant (of $\psi$). KL divergence is not symmetric in its arguments and the choice of Eq.~\eqref{eq: KL divergence definition} is not analogous to the one used in variational inference. Rather it is analogous to that of Expectation Propagation or that of Bayesian utility. 
The motivation stems from the availability of unbiased minibatch estimates $\hat{p}_B(y;w)$ of $p(y|w)$ as per \ref{sec: Appendix -- minibatch estimates}. Plugging into Eq.~\eqref{eq: ELL} yields a minibatch estimate of the KL loss for the auxiliary network:
\begin{equation}
-\int_{\calY} \hat{p}_B(y;w)\log{q_\psi(w)(y)} \, dy \, , \label{eq: ELL - approximation 1}
\end{equation}
Taking note of the linearity of Eq.~\eqref{eq: ELL} w.r.t.~$p(\cdot|w)$, Eq.~\eqref{eq: ELL - approximation 1} 
is in fact an unbiased estimate of the KL loss.
The log-marginals $\log{q_\psi(w)(y)}$ needed in Eq.~\eqref{eq: ELL - approximation 1} are returned by the auxiliary network. 

The minibatch training loss of Eq.~\eqref{eq: ELL - approximation 1} can be interpreted as a ``soft'' variant of the negative log-likelihood (NLL) loss with soft targets $\hat{p}_B(y;w)$. The network weights $\psi$ are updated during each iteration by backpropagation, concurrently with the update of the main network parameters. One has to take care to detach $\hat{p}_B(y;w)$ from the computational graph prior to computing the auxiliary loss, and to detach $q_\psi(w)(y)$ from the computational graph prior to computing the main network's loss so that no spurious backpropagation occurs on either $\psi$ nor $w$ from the other loss' backward pass. 

In the context of variational inference (VI), the loss Eq.~\eqref{eq: ELL} is taken in expectation over the variational posterior $q(w)$:
\begin{equation}
 -\left\langle \int_{\calY} p(y|w)\log{q_\psi(w)(y)} \, dy \right\rangle_{q(w)} \, . \label{eq: ELL -- VI}
\end{equation}
In VI for DL, the mainstream approach to training $q(w)$ couples the so called reparametrization trick with stochastic backpropagation. Parameters $w_k$, $k=1\cdots K$ are sampled from $q(w)$ in an autodifferentiable manner and used to evaluate relevant integrals by Monte Carlo integration. Hence to minimize Eq.~\eqref{eq: ELL -- VI} one may reuse the samples $w_k$, compute Eq.~\eqref{eq: ELL - approximation 1} for each $w_k$ and use the empirical average as the loss, before backpropagating from the loss to $\psi$ as usual.

\subsection{Auxiliary network architecture and practical considerations}
\label{sec: auxiliary network architectures}

We set a large learning rate for the auxiliary network parameters $\psi$ compared to the main network's parameters $w$. The learning rate of the former is typically set to $10$ times that of the latter. 
A single update of the auxiliary network is performed with every update of the main network.

Since $0$th order accuracy is sufficient for the auxiliary network, we favour minimalistic architectures 
such that~(a) $q_\psi$ is lightweight despite taking as input a high dimensional object (the parameters $w$ of $NN_w$);~(b) $\psi$ is easily trainable;~(c) large learning rates can be used. 
We have opted for the two following architectures for their simplicity and suitable performance.\\

\noindent
\textit{The constant prediction.} The parameters $\psi\!\coloneqq\!\bm{\eta}$ of the auxiliary network are logits $\bm{\eta}\triangleq [\eta_1\cdots \eta_{\calY}]$ from which the log-marginal probabilities $\log{q_\psi(w)}\!\triangleq\! \text{logsoftmax}(\bm{\eta})$ are computed. The output is then a constant of $w$. This is sufficient provided that one does MAP/MLE inference of network weights $w$, that the log-marginals smoothly vary w.r.t.~$w$ or that the learnable $\bm{\eta}$ are adjusted quickly enough.\\

\noindent
\textit{The logistic regressor.} The auxiliary network is a linear layer plus a $|\calY|$-dimensional bias vector $\bm{b}$, followed by a $\text{logsoftmax}$ activation. The architecture uses a custom implementation of the linear layer. It combines a $P\times |\calY|$ weight matrix $\rmA$ ($P$ the dimension of $w$) and a $P$-dimensional offset vector $o$; so that given input $w$, it outputs $\rmA^T (w-{o})$. Combined with the bias vector and the non-linearity this leads to a prediction $q_\psi(w)\triangleq \text{logsoftmax}(\bm{\eta}(w))$ with logits $\bm{\eta}(w)=\rmA^T (w-{o}) + \bm{b}$. 

The redundancy between offset $o$ and bias $\bm{b}$ is intended. This enables $\bm{b}$ to capture the ``global trend'' for the log-marginal values, while the linear part adjusts for the ``local trend'' around some parameter value $o$. Offsets $o$ are expected to move towards some average of probable values $w$ visited during stochastic updates. Learning rates are set for $b$ and $o$ in regard to these intuitions. $b$ has the largest learning rate by at least an order of magnitude, so that it adjusts fastest during the first epoch(s). $o$ has the virtual dimensionality of $w$, and its learning rate mimics that of $w$. This allows $o$ to ``track'' $w$. $o$ can be initialized to the value of $w$. $A$ is randomly initialized to small values (or zeros). We use regularizing priors all throughout experiments, 
and indeed for both the main and auxiliary networks. For the auxiliary network, we place a sparsity-inducing prior on $A$, as sparsity is known to encourage robustness in the $|P|\gg |\calY|$ setting. It also breaks the redundancy between $o$ and $\bm{b}$.

\section{Variance of estimators and minibatch design}
\label{sec: variance of estimators}

The approach described in the article is compatible with most minibatch designs: \ref{sec: generalized minibatch estimates} and \ref{sec: Appendix -- minibatch estimates} guarantee that minibatch estimates are unbiased regardless of the class balance. This prompts the question of whether there is a minibatch design superior to others. We give a few elements of answer from the perspective of minimizing the variance of minibatch estimates. 

Many of the quantities to be estimated from minibatches are of the form $\mathbb{E}_{\calD}[g(x,y)]\ttriangleq \bar{g}$ for some function $g(x,y)$ implicitly allowed to depend on $w$ and other variables as needed. For instance one retrieves the marginal $\bar{g}\eq p(y'|w)$ when setting $g(x,y)\!\coloneqq\! p(y'|x,w)$ (in that instance $g(x,y)$ does not depend on $y$; instead it depends implicitly on a free variable $y'\in\calY$). The frequentist prediction risk $\bar{g}\eq \mathbb{E}_{\calD}[\log{p(y|x,w)}]$ is obtained when setting $g(x,y)\!\coloneqq\! \log{p(y|x,w)}$. The gradient of these quantities w.r.t.~model parameters $w$ is obtained when taking $g(x,y)\ccoloneqq \nabla_w p(y'|x,w)$, resp.~$g(x,y)\!\coloneqq\! \nabla_w \log{p(y|x,w)}$.

Therefore one may ask whether there are ways to design minibatches that guarantees a good approximation quality for minibatch estimators. We have already encountered estimators for $\bar{g}$ in specific subcases in previous sections. Namely,
\begin{equation}
\hat{g}^{(i)} \triangleq \frac{1}{n_B} \sum_{n\in B} \beta^{(i)}(y_n) g(x_n,y_n)\, ,
\end{equation} 
defines for $i=1,2$ two unbiased estimators of $\bar{g}$. The mean square error $\mathbb{E}_{X_B,Y_B}[(\hat{g}^{(i)}-\bar{g})^2]$ of these estimators, where the expectation is taken w.r.t.~the minibatch sample $(X_B,Y_B)$, is then exactly the variance $\text{Var}_{X_B,Y_B}[\hat{g}^{(i)}]$, which suggests to look closely at these variances.

It turns out that $\text{Var}_{X_B,Y_B}[\hat{g}^{(i)}]$ can be expressed in terms of readily computable quantities. All details aside, these expressions suggest a few simple rules of thumb for the case where the number of labels is small compared to the minibatch size, $|\calY|\ll n_B$. Firstly, $\hat{g}^{(2)}$ is likely to have smaller variance than $\hat{g}^{(1)}$ and should be preferred. Secondly, the optimal minibatch design w.r.t.~the variance (\eg, of gradients) is then always a deterministic strategy with fixed label counts, rather than randomly sampled labels. Thirdly, after deciding label counts $N_B=[n_B(y),y\in\calY]$, the $n_B(y)$ samples $x_n$ for each class $y$ must of course be sampled i.i.d. among those samples of class $y$ in the training dataset.

Under the three conditions above, the variance $\text{Var}_{X_B,Y_B}[\hat{g}^{(2)}]$ is given by Eq.~\eqref{eq: g2 variance} 
\begin{equation}
\text{Var}_{X_B,Y_B}\left[\hat{g}^{(2)}\right] = 
\frac{1}{n_B}\sum_{y\in\calY}\hat{p}_\calY(y)\, \text{Var}_{x|y}\left[ g(x,y) \frac{p_\calY(y)}{\hat{p}_\calY(y)} \right] \, ,
\label{eq: g2 variance}
\end{equation}
where $\hat{p}_\calY(y)\triangleq n_B(y)/n_B$ stands for the empirical frequency of labels in the minibatch of fixed size $n_B$. Furthermore, the variance can be estimated from validation data. For this it suffices to get an estimate $\hat{v}_y$ of the conditional variance $\text{Var}_{x|y}[ g(x,y) ]$ for all $y\in\calY$. Let $B_V$ a validation batch with $n_{B_V}(y)\geq 2$ for all $y$. Eq.~\eqref{eq: estimate of the conditional variance} gives the unbiased sample variance estimator: 
\begin{equation}
\hat{v}_y \triangleq \frac{1}{n_{B_V}(y)-1}\cdot \sum_{\substack{n\in B_V:\\y_n=y}}\left(g(x_n,y_n) - \hat{g}_y(X_{B_V},Y_{B_V})\right)^2 \, ,
\label{eq: estimate of the conditional variance}
\end{equation}
where $\hat{g}_y(X_{B_V},Y_{B_V})\!\triangleq\! 1/n_{B_V}(y) \cdot\sum_{n:y_n=y}g(x_n,y_n)$ is an estimate of the conditional expectation $\mathbb{E}_{x|y}[g(x,y)]$.

If hardware constrains minibatch sizes $n_B$ not to be large compared to the number $|\calY|$ of labels, alternative estimators such as $g^{(1)}$ become more relevant.

\section{Additional results}
\label{sec: additional results}

Table~\ref{table: binary deep learning -- InceptionNet performance} reports the performance metrics for the binary classification experiment of section~\ref{sec: deep learning case study}, when using an alternative architecture with inception blocks. As with the dropout ConvNet, minibatches are sampled with rebalancing, \ie with equal number of benign and malignant examples. Table~\ref{table: binary deep learning -- ConvNet performance without rebalancing} shows similar results with the ConvNet when sampling minibatches uniformly among the training batch without rebalancing.

\begin{table*}\centering
\ra{.8}
\small
\tabcolsep=0.14cm
\begin{tabular*}{0.9\textwidth}{rcccccccccccc}\toprule

& NELL & $\text{NLL}_{HO}$ & $\text{Rsk}_{HO}$ & $\text{Acc}_{HO}$ & {TNR}& {TPR} & $\text{NPV}_{HO}$& $\text{PPV}_{HO}$ & AUC & I & $\text{M}_{HO}$ & $\text{MCC}_{HO}$\\ \midrule
$\text{prev.}=0.5$ \\	
{IG} & ${0.516}$ & $\mathbf{0.374}$ & $\mathbf{-0.348}$ & $\mathbf{0.87}$ & $\mathbf{0.90}$ & $0.76$ & ${0.92}$ & $\mathbf{0.72}$ & $\mathbf{0.89}$& $\mathbf{0.66}$ & $\mathbf{0.64}$ & $\mathbf{0.65}$ \\ 
{IW} & $\mathbf{0.510}$ & ${0.405}$ & $-0.291$ & ${0.84}$ & ${0.86}$& $\mathbf{0.79}$ & $\mathbf{0.93}$ & $0.64$ & $\mathbf{0.89}$ & ${0.65}$ & ${0.57}$ & ${0.61}$ \\[0.2cm]
$\text{prev.}=0.25$ \\	
{IG} & ${0.413}$ & ${0.397}$ & $\mathbf{-0.211}$ & $\mathbf{0.86}$ & $0.88$ & $\mathbf{0.76}$ & $\mathbf{0.92}$ & ${0.68}$ & ${0.88}$ & $\mathbf{0.65}$ & $0.60$ & $\mathbf{0.63}$ \\ 
{IW} & $\mathbf{0.409}$ & $\mathbf{0.347}$ & ${-0.207}$ & $\mathbf{0.86}$ & $\mathbf{0.92}$ & ${0.67}$ & $0.89$ & $\mathbf{0.73}$& $\mathbf{0.89}$ & ${0.60}$ & $\mathbf{0.63}$ & ${0.61}$ \\[0.2cm]
$\text{prev.}=0.1$ \\	
{IG} & ${0.268}$ & ${0.464}$ & $\mathbf{-0.277}$ & ${0.83}$ & $0.83$ & $\mathbf{0.82}$ & $\mathbf{0.94}$ & ${0.61}$& ${0.88}$& $\mathbf{0.65}$ & $0.55$ & $\mathbf{0.60}$ \\ 
{IW} & $\mathbf{0.255}$ & $\mathbf{0.387}$ & $-0.254$ & $\mathbf{0.85}$ & $\mathbf{0.96}$ & ${0.50}$ & $0.86$ & $\mathbf{0.83}$ & $\mathbf{0.89}$ & $0.47$ & $\mathbf{0.69}$ & ${0.56}$ \\[0.2cm]
$\text{prev.}=0.01$ \\	
{IG} & ${0.093}$ & $\mathbf{0.569}$ & $\mathbf{-0.543}$ & $\mathbf{0.78}$ & $0.76$ & $\mathbf{0.85}$ & $\mathbf{0.94}$ & $0.54$ & ${0.88}$ & $\mathbf{0.61}$ & $0.48$ & $\mathbf{0.54}$ \\ 
{IW} & $\mathbf{0.046}$ & $0.689$ & $-0.451$ & $\mathbf{0.78}$ & $\mathbf{1.0}$ & ${0.10}$ & $0.78$ & $\mathbf{1.0}$ & $\mathbf{0.90}$ & ${0.10}$ & $\mathbf{0.78}$ & ${0.26}$ \\[0.2cm]
$\text{prev.}=0.001$ \\	
{IG} & ${0.086}$ & $\mathbf{0.610}$ & $\mathbf{-0.625}$ & $\mathbf{0.78}$ & ${0.76}$ & $\mathbf{0.85}$ & $\mathbf{0.94}$ & $\mathbf{0.53}$ & $\mathbf{0.88}$ & $\mathbf{0.61}$ & $\mathbf{0.47}$ & $\mathbf{0.54}$ \\ 
{IW} & $\mathbf{0.007}$ & $1.198	$ & $-0.486$ & ${0.76}$ & $\mathbf{1.0}$ & $0.0$ & $0.76$ & NaN & $\mathbf{0.88}$ & $0.0$ & NaN & NaN\\

\bottomrule
\end{tabular*}
\label{table: binary deep learning -- InceptionNet performance}
\caption{Performance summary on held-out data for the Bayesian approach to prevalence-bias (IG) vs.~importance weighted log-loss (IW), averaged across the three folds, for the InceptionNet architecture. NELL: negative expected log-likelihood estimate for the true population. $\text{NLL}_{HO}$: negative hold-out log-likelihood. $\text{Rsk}_{HO}$: hold-out risk. $\text{Acc}_{HO}$: accuracy. TPR / NPR: positive / negative rates. $\text{PPV}_{HO}$ / $\text{NPV}_{HO}$: positive / negative predictive values. AUC: Area Under the (ROC) Curve. I / M: informedness / markedness. MCC: Matthews Correlation Coefficient.}
\end{table*} 

\begin{table*}\centering
\ra{.8}
\small
\tabcolsep=0.14cm
\begin{tabular*}{0.9\textwidth}{rcccccccccccc}\toprule

& NELL & $\text{NLL}_{HO}$ & $\text{Rsk}_{HO}$ & $\text{Acc}_{HO}$ & {TNR}& {TPR} & $\text{NPV}_{HO}$& $\text{PPV}_{HO}$ & AUC & I & $\text{M}_{HO}$ & $\text{MCC}_{HO}$\\ \midrule
$\text{prev.}=0.5$ \\	
{IG} & $\mathbf{0.492}$ & $\mathbf{0.397}$ & $\mathbf{-0.318}$ & $\mathbf{0.85}$ & $\mathbf{0.87}$ & $\mathbf{0.79}$ & $\mathbf{0.93}$ & $\mathbf{0.66}$ & $\mathbf{0.89}$& $\mathbf{0.66}$ & $\mathbf{0.59}$ & $\mathbf{0.62}$ \\ 
{IW} & ${0.498}$ & ${0.410}$ & $-0.293$ & ${0.84}$ & ${0.86}$& $\mathbf{0.79}$ & $\mathbf{0.93}$ & $0.64$ & $\mathbf{0.89}$ & ${0.65}$ & ${0.57}$ & ${0.61}$ \\[0.2cm]
$\text{prev.}=0.25$ \\	
{IG} & ${0.397}$ & ${0.419}$ & ${-0.219}$ & ${0.84}$ & $0.86$ & $\mathbf{0.80}$ & $\mathbf{0.93}$ & ${0.64}$ & $\mathbf{0.89}$ & $\mathbf{0.65}$ & $0.57$ & $0.61$ \\ 
{IW} & $\mathbf{0.396}$ & $\mathbf{0.335}$ & $\mathbf{-0.220}$ & $\mathbf{0.87}$ & $\mathbf{0.93}$ & ${0.68}$ & $0.90$ & $\mathbf{0.78}$& $\mathbf{0.89}$ & ${0.62}$ & $\mathbf{0.68}$ & $\mathbf{0.65}$ \\[0.2cm]
$\text{prev.}=0.1$ \\	
{IG} & ${0.262}$ & ${0.498}$ & $\mathbf{-0.300}$ & ${0.81}$ & $0.80$ & $\mathbf{0.85}$ & $\mathbf{0.95}$ & ${0.58}$& $\mathbf{0.89}$& $\mathbf{0.65}$ & $0.53$ & $0.59$ \\ 
{IW} & $\mathbf{0.251}$ & $\mathbf{0.348}$ & $-0.287$ & $\mathbf{0.87}$ & $\mathbf{0.97}$ & ${0.55}$ & $0.87$ & $\mathbf{0.84}$ & $\mathbf{0.89}$ & $0.52$ & $\mathbf{0.72}$ & $\mathbf{0.61}$ \\[0.2cm]
$\text{prev.}=0.01$ \\	
{IG} & ${0.073}$ & $\mathbf{0.578}$ & $\mathbf{-0.573}$ & ${0.76}$ & $0.72$ & $\mathbf{0.87}$ & $\mathbf{0.95}$ & $0.50$ & $\mathbf{0.88}$ & $\mathbf{0.59}$ & $0.44$ & $\mathbf{0.51}$ \\ 
{IW} & $\mathbf{0.053}$ & $0.653$ & $-0.467$ & $\mathbf{0.78}$ & $\mathbf{0.99}$ & ${0.11}$ & $0.78$ & $\mathbf{0.93}$ & $\mathbf{0.87}$ & ${0.10}$ & $\mathbf{0.71}$ & ${0.26}$ \\[0.2cm]
$\text{prev.}=0.001$ \\	
{IG} & ${0.042}$ & $\mathbf{0.602}$ & $\mathbf{-0.661}$ & ${0.75}$ & ${0.70}$ & $\mathbf{0.88}$ & $\mathbf{0.95}$ & $\mathbf{0.49}$ & $\mathbf{0.88}$ & $\mathbf{0.58}$ & $\mathbf{0.44}$ & $\mathbf{0.50}$ \\ 
{IW} & $\mathbf{0.008}$ & $1.128$ & $-0.540$ & $\mathbf{0.76}$ & $\mathbf{1.0}$ & $0.0$ & $0.76$ & NaN & 0.86 & $0.0$ & NaN & NaN\\

\bottomrule
\end{tabular*}
\label{table: binary deep learning -- ConvNet performance without rebalancing}
\caption{Performance summary on held-out data for the Bayesian approach to prevalence-bias (IG) vs.~importance weighted log-loss (IW), averaged across the three folds, for the ConvNet architecture, when sampling minibatches uniformly without rebalancing. NELL: negative expected log-likelihood estimate for the true population. $\text{NLL}_{HO}$: negative hold-out log-likelihood. $\text{Rsk}_{HO}$: hold-out risk. $\text{Acc}_{HO}$: accuracy. TPR / NPR: positive / negative rates. $\text{PPV}_{HO}$ / $\text{NPV}_{HO}$: positive / negative predictive values. AUC: Area Under the (ROC) Curve. I / M: informedness / markedness. MCC: Matthews Correlation Coefficient.}
\end{table*} 

\section{Prevalence bias as sample selection}
\label{sec: prevalence bias as sample selection}

\subsection{From sample selection to prevalence bias}
\label{sec: Equivalence sample selection -- prevalence bias}

\begin{figure}
\centering
\includegraphics[width=\columnwidth]{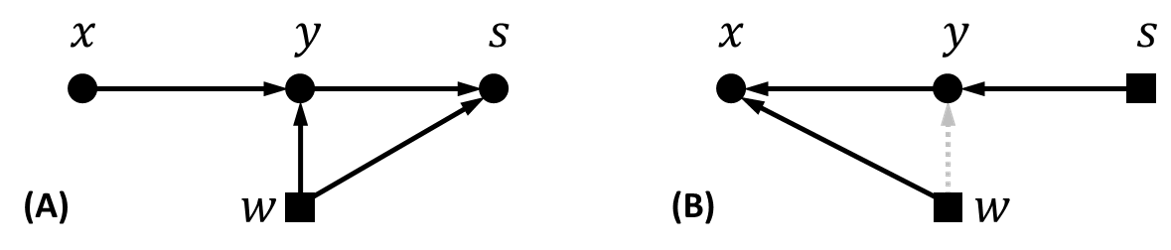}
\caption{Sample selection mechanisms in the form of~(A) can give rise to the prevalence bias model discussed in this work. 
The general population is sampled from $p(x,y|w^\ast)$. The training data is obtained by sampling from the conditional $p(x,y|s\eq 1, w^\ast)$. 
Seen as a structural graphical model, this yields (B). Square nodes signal conditioning. The dashed grey link signals a conditional dependency that is absent for $s=1$ since $p(y|\, s\eq 1, w)=\tilde{p}(y)$ but in general not for $s\eq 0$.}
\label{fig: prevalence bias as sample selection}
\end{figure}

Sample selection introduces an acceptance-reject mechanism that decides whether the sample can ever be observed ($s=1$) or not ($s=0$). To derive the label-dependent sampling bias, let the selection variable $s$ be conditioned on $y$ and $w$. A dependence on $w$ is a priori allowed (we will see shortly that in fact it cannot be removed). This leads to Fig.~\ref{fig: prevalence bias as sample selection}(A) for the generative model.
The joint distribution expands in several ways:
\begin{align}
p(y,x,w,s) & = p(y|x,w)p(x)p(w)p(s|y,w) \, , \label{eq: sample selection joint -- from the graph}\\
\, & = p(x|y,w)p(y|w)p(w)p(s|y,w) \, , \label{eq: sample selection joint -- from Bayes rule}\\
\, & = p(x,y|s,w)p(s|w)p(w) \, . \label{eq: sample selection joint -- always}
\end{align}
Eq.~\eqref{eq: sample selection joint -- from the graph} directly translates the graph assumptions. Eq.~\eqref{eq: sample selection joint -- from Bayes rule} follows from Bayes' rule. Eq.~\eqref{eq: sample selection joint -- always} holds for any joint distribution. Equating Eq.~\eqref{eq: sample selection joint -- from Bayes rule} and Eq.~\eqref{eq: sample selection joint -- always} and simplifying\footnote{Alternatively, $p(x,y|s,w)\eq p(x|y,w,s)p(y|s,w)$ and remark from the graph that $x\pperp s \,\,|y,w$.}, we find that:
\begin{equation}
p(x,y|s,w) = p(x|y,w)p(y|s,w)
\label{eq: sample selection -- generative conditional}
\end{equation}
In the sample selection model, training data is collected among observable subjects $s\eq 1$ by sampling from $p(x,y|\,s\eq 1,w)$.  
In particular, the relationship with prevalence bias becomes clear if one is allowed to choose $p(y|\,s\eq 1, w)\!\coloneqq\! \tilde{p}(y)$, where $\tilde{p}(y)$ is the distribution of labels specified for data collection. This is always possible and corresponds to fixing a choice of conditional $p(s|y,w)$ in the factorization Eq.~\eqref{eq: sample selection joint -- from the graph}. Indeed from Bayes' rule and by expanding $p(s|w)$ as an integral:
\begin{equation}
\frac{p(s|y,w)p(y|w)}{\int_\calY p(s|y,w) p(y|w)dy} = p(y|s,w)\, ,
\end{equation}
and since $\tilde{p}(y)$ sums to $1$, Eq.~\eqref{eq: sample selection -- prevalence bias constraint}: 
\begin{equation}
\frac{p(s\eq 1 \,|y,w)p(y|w)}{\int_\calY p(s\eq 1\, |y,w) p(y|w)dy} = \tilde{p}(y)
\label{eq: sample selection -- prevalence bias constraint}
\end{equation}
always admits solutions\footnote{$p(y|w)$, $y\in\calY$ is not allowed to cancel for any value of $w$. Although many models readily prevent this from happening, one can always choose a prior $p(w)$ to assign probability $0$ to this subset. $\tilde{p}(y)$ is not allowed to cancel for any $y$ either.}, for which the desired result holds. An example solution is $p(s\eq 1 \,|y,w)\coloneqq \alpha(w)\cdot\tilde{p}(y)/p(y|w)$, $p(s\eq 0 \,|y,w)\coloneqq 1-\alpha(w)\tilde{p}(y)/p(y|w)$ with $\alpha(w)\triangleq \min_y p(y|w)/\tilde{p}(y)$. It is easily checked from Eq.~\eqref{eq: sample selection -- prevalence bias constraint} that no solution can be found in the form $p(s\eq 1|y)$ unless $p(y|w)$ is a constant of $w$, hence the dependence on $w$ cannot be dropped.

Fig.~\ref{fig: prevalence bias as sample selection}(B) represents the generative process with sample selection. Keeping the selection variable $s$ implicit, we recover the viewpoint adopted for simplicity in Fig.~\ref{fig: graphical models} and the main text. More rigorously in decision theoretic terms, define as per section~\ref{sec: Bayesian risk} the optimal prediction rule as minimizing the Bayesian risk of Eq.~\eqref{eq: Bayes prediction risk -- selection}:
\begin{equation}
\calR_{\text{Bayes}}[q_{x_\ast,X,Y}(y)]  \triangleq 
\mathbb{E}_{w\sim p(w)}\!\left[
	\calR_w[q_{x_\ast,X,Y}]
\right] \, ,
\label{eq: Bayes prediction risk -- selection}
\end{equation}
with $\calR_w$ defined by Eq.~\eqref{eq: frequentist risk -- selection}:
\begin{equation}
-\calR_w[q_{x_\ast,X,Y}]  \triangleq \mathbb{E}_{(X,Y)\sim\calD'_w}\!\left[
								\mathbb{E}_{(x_\ast,y_\ast)\sim\calD_w}[\log{q_{x_\ast,X,Y}(y_\ast)}]
						  \right]\, .
\label{eq: frequentist risk -- selection}
\end{equation}
Here $(X,Y)\!\sim\!\calD'_w$ refers to sampling the training data i.i.d.~ from the conditional $p(x,y|\,s\eq 1,w)$, so that the expectation is w.r.t. the (conditioned) joint distribution $p(X,Y|\,S\eq 1,w)$:
\begin{align}
p(X,Y|\,S\eq 1,w) &= \prod_{n=1}^{N} p(x_n,y_n|\,s_n\eq 1,w)\, , \\
\, &= \prod_{n=1}^{N} p(x_n|y_n,w) \tilde{p}(y_n) \, ,
\label{eq: sample selection -- training data likelihood}
\end{align}
where $S\eq 1$ is shorthand for $\{s_n\eq1, \,n\eq 1\cdots N\}$ and the last line uses Eq.~\eqref{eq: sample selection -- generative conditional}. $(x_\ast,y_\ast)\sim\calD_w$ refers to the generative process at test time, so that the expectation is w.r.t. the distribution:
\begin{equation}
p(x_\ast,y_\ast|w)=p(y_\ast|x_\ast,w)p(x_\ast)\, .
\label{eq: sample selection -- test data likelihood}
\end{equation}
The derivations of the optimal prediction rule mirror those of~\ref{sec: proofs}. The Bayesian risk is an integral w.r.t.~$(X,Y)$, $w$ and $(x_\ast,y_\ast)$. The integrand is the product of $\log{q_{x_\ast,X,Y}(y_\ast)}$, of Eq.~\eqref{eq: sample selection -- training data likelihood}, of $p(w)$ and of Eq.~\eqref{eq: sample selection -- test data likelihood}:
\begin{equation}
p(x_\ast,y_\ast|w) p(X,Y|\,S\eq 1,w) p(w) \,\log{q_{x_\ast,X,Y}(y_\ast)}\, .
\label{eq: integrand of Bayes risk with selection}
\end{equation}
Define the Bayesian posterior for $w$ under $S\eq1$, 
$p_{S=1}(w|X,Y)$, as the probability distribution proportional to the product of the prior $p(w)$ and of the likelihood of Eq.~\eqref{eq: sample selection -- training data likelihood}:
\begin{equation}
p_{S= 1}(w|X,Y)\triangleq \frac{1}{Z_{X,Y}}\cdot p(X,Y|\,S\eq 1,w) p(w)\, ,
\label{sec: sample selection -- posterior}
\end{equation}
where $Z_{X,Y}$ normalizes the posterior. Define the predictive posterior under $S\eq1$, $p_{S=1}(y_\ast|x_\ast,X,Y)$, by Eq.~\eqref{sec: sample selection -- predictive posterior}:
\begin{equation}
p_{S=1}(y_\ast|x_\ast,X,Y)\triangleq \int p(y_\ast|x_\ast,w)p_{S=1}(w|X,Y)dw\, .
\label{sec: sample selection -- predictive posterior}
\end{equation}
The predictive posterior minimizes the Bayes risk, which justifies calling it so. Indeed taking $\log{q_{x_\ast,X,Y}(y_\ast)}$ 
out of the integral w.r.t. $w$ in the expression of the Bayesian risk:
\begin{equation}
-\int p_{S=1}(y_\ast|x_\ast,X,Y)p(x_\ast)Z_{X,Y}\log{q_{x_\ast,X,Y}(y_\ast)}d(X,Y)d(x_\ast,y_\ast) \, .
\label{eq: sample selection -- risk 1}
\end{equation}
Eq.~\eqref{eq: sample selection -- risk 1} decouples over every value of $X,Y,x_\ast$ leaving us to maximize decoupled problems of the form of Eq.~\eqref{eq: sample selection -- risk 2} w.r.t. distributions $q_{x_\ast,X,Y}(y_\ast)$ over $\calY$:
\begin{equation}
\int p_{S=1}(y_\ast|x_\ast,X,Y)\log{q_{x_\ast,X,Y}(y_\ast)}dy_\ast \, .
\label{eq: sample selection -- risk 2}
\end{equation}
Thus $q_{x_\ast,X,Y}(y_\ast)\eq p_{S=1}(y_\ast|x_\ast,X,Y)$. As announced, the posterior $p_{S=1}(w|X,Y)$ of Eq.~\eqref{sec: sample selection -- posterior} corresponds to the one of~\ref{sec: proofs}, Eq.~\eqref{eq: same conditional}, the predictive posterior is exactly the one of section~\ref{sec: Bayesian analysis}, Eq.~\eqref{eq: predictive posterior}, and Fig.~\ref{fig: prevalence bias as sample selection} makes the link with Fig.~\ref{fig: graphical models}. 

The fundamental insight is that training data is sampled \textit{controlling} model parameters $w$ (controlled by ``laws of nature'') and the label distribution (controlled by selection mechanisms). The choice to make these constraints manifest or to present upfront the resulting structural model of the training dataset is a matter of convenience. 

\subsection{When the test-time population is drawn with selection}
\label{sec: test-time selection}

Sometimes the population of interest may be drawn in a similar manner as the training data, both being drawn with label-based selection mechanisms. The optimal prediction rule in this case is the (conditional) predictive posterior:
\begin{equation}
p(y_\ast|\,s_\ast\eq 1, x_\ast, X,Y) = \int p(y_\ast|\,s_\ast\eq 1, x_\ast, w) p(w|X,Y)dw\, ,
\label{eq: conditional predictive posterior with selection}
\end{equation}
where:
\begin{align}
p(y_\ast|\,s_\ast\eq 1, x_\ast, w) &= \frac{p(x_\ast|y_\ast,w)\tilde{p}(y_\ast)}{\int p(x_\ast|y_\ast,w)\tilde{p}(y_\ast) dy} \, , \\
& = \frac{p(y_\ast|x_\ast,w)\tilde{p}(y_\ast)/p(y_\ast|w)}{\int p(y_\ast|x_\ast,w)\tilde{p}(y_\ast)/p(y_\ast|w) dy} \, .
\label{eq: conditional predictive likelihood with selection}
\end{align}
Eq.~\eqref{eq: conditional predictive likelihood with selection} holds as long as the population of interest is drawn with label-based selection, regardless of whether the label distribution $\tilde{p}(y_\ast)$ differs from the distribution $\tilde{p}_Y(Y)$ of the training labels. It can be understood as the (normalized) product of a likelihood $p(x_\ast|y_\ast,w)\propto p(y_\ast|x_\ast,w)/p(y_\ast|w)$ by the test specific prior $\tilde{p}(y_\ast)$.\\

\textit{Unknown test-time prevalence}. What if the test-time prevalence is unknown? This corresponds to a typical ``challenge'' scenario where the organiser decides on the prevalence $\tilde{p}(y_\ast)$ of labels in the benchmark, but does not communicate it to the participants. $N_{te}$ data points $x_{n,\ast}$ are provided and a prediction $y_{n,\ast}$ for each of the $N_{te}$ data points is expected. The Bayesian treatment can actually be extended to infer information about the hidden probability distribution $\tilde{p}(y_\ast)$. Denote $X_\ast\triangleq(x_{1,\ast}\cdots x_{N_{te},\ast})$ (and likewise $Y_\ast$) the ordered collection of test points, to be contrasted with training data $X,Y$. The question is formalized and answered as follows. 

In full generality the hidden label sampling distribution $\tilde{p}_{\bm{\pi}}(Y_\ast)$ of the organiser could be a joint distribution over $Y_\ast$ (parametrized by unknown parameters $\bm{\pi}$). For instance the organiser could decide on an exact label count (equal counts across labels, say). The generative model of data leads to the following distribution for test data:
\begin{align}
p(X_\ast,Y_\ast|S_\ast=1, w, \bm{\pi}) & = \tilde{p}_{\bm{\pi}}(Y_\ast)\cdot\prod_{n\in N_{te}} p(x_{n,\ast}|y_{n,\ast},w) \, ,\\
\, & \propto \tilde{p}_{\bm{\pi}}(Y_\ast)\cdot\prod_{n\in N_{te}} \frac{p(y_{n,\ast}|x_{n,\ast},w)}{p(y_{n,\ast} | w)} \, ,
\end{align}
Henceforth as an example and with the intent to avoid combinatorial problems, we restrict our attention to i.i.d.~draws according to a categorical distribution $\tilde{p}_{\bm{\pi}}(Y_\ast)\coloneqq \prod_n p(y_{n,\ast}|\bm{\pi})$, with $\bm{\pi}=(\pi_y, y\in\calY)$ describing the (unknown) label probabilities.

Due to the independence between training and test data given $w$, we can adopt a test-centric view in which the training data only enters the problem description via an updated prior on $w$, a.k.a.~the posterior $p(w|X,Y)$. In other words the posterior $p(w|X,Y)$ incorporating information about the training data replaces the original prior $p(w)$. 

If one assumes that the size of the test dataset $N_{te}\ll N$ is sufficiently small compared to that of the training dataset, so that the resulting posterior $p(w|X,Y,X_\ast)$ on $w$ is not too different from $p(w|X,Y)$, it dispenses from a test-time round of optimization on $w$ and allows to reuse the trained architecture as is. The inference then focuses on $Y_\ast$, and in the present case $\bm{\pi}$.

We endow $\bm{\pi}$ with a conjugate Dirichlet prior $p(\bm{\pi})=\calD(\bm{\pi}|\bm{\alpha}_0)$, where $\bm{\alpha}_0=(\alpha_{y,0}, y\!\in\!\calY)>0$. A flat prior over $\bm{\pi}$ in the $|\calY|$-simplex corresponds to the choice $\alpha_{y,0}=\alpha_0=1$ for all $y\in\calY$. The joint distribution $p(X_\ast,Y_\ast,w,\bm{\pi}|S_\ast\eq 1, X,Y)$ has its logarithm write as:
\begin{equation}
\begin{split}
\sum_{n\in N_{te}} &\log{\frac{p(y_{n,\ast}|x_{n,\ast},w)}{p(y_{n,\ast}|w)}} + \log{p(y_{n,\ast}|\bm{\pi})} \\
\, & + \log{p(w|X,Y)} + \log{p(\bm{\pi})} + \text{cst} \, .
\end{split}
\end{equation}
Proceeding within the variational Bayesian framework with a variational family $q(Y_\ast,w,\bm{\pi})\triangleq q_{Y_\ast}(Y_\ast|w)q_{\bm{\pi}}(\bm{\pi}|w)q_w(w)$, 
one gets that $q_{Y_\ast}(Y_\ast|w)=\prod_n q_{n}(y_{n,\ast}|w)$ factorizes over data points, and that $q_{\bm{\pi}}(\bm{\pi}|w)=\calD(\bm{\pi}|\bm{\alpha}(w))$ remains in the Dirichlet family with updated parameters $\bm{\alpha}(w)=(\alpha_1(w)\cdots \alpha_{|\calY|}(w))$. The following iterative updates greedily improve an evidence lower-bound:
\begin{align}
q_n(y_{n,\ast}|w) &\propto \frac{p(y_{n,\ast}|x_{n,\ast},w)}{p(y_{n,\ast}|w)} \cdot \tilde{{\pi}}_{y_{n,\ast}}(w) \, , \label{eq: VBI unknown test prevalence - y} \\
\alpha_y(w) & = \alpha_{y,0} + \sum_{n\in N_{te}} q_n(y_{n,\ast}=y|w)\, , \label{eq: VBI unknown test prevalence - pi}
\end{align}
where $\log{\tilde{\pi}_y}(w)\triangleq \langle\log{\pi_y}(w)\rangle_{q_{\bm{\pi}|w}}= \psi(\alpha_y(w)) - \psi(\sum_{y'\in\calY} \alpha_{y'}(w))$ and $\psi(\cdot)$ is the digamma function. As explained one may choose to fix $q_w(w)\ccoloneqq p(w|X,Y)$ rather than jointly update it. Then all updates are computationally inexpensive as the forward model $NN_w(x_\ast)$ is evaluated once and for all. Only $\tilde{{\pi}}_{y_{n,\ast}}(w)$ changes across iterations of Eq.~\eqref{eq: VBI unknown test prevalence - y}. 

Eq.~\eqref{eq: VBI unknown test prevalence - y} is the counterpart of Eq.~\eqref{eq: conditional predictive likelihood with selection} for the unknown prevalence scenario. The counterpart of $\tilde{p}(y)$ in Eq.~\eqref{eq: conditional predictive likelihood with selection} is $\tilde{\pi}_y(w)$, as an estimate of the test-time prevalence. After iterating Eq.~\eqref{eq: VBI unknown test prevalence - y} and Eq.~\eqref{eq: VBI unknown test prevalence - pi} to convergence, the approximate predictive posterior for any one of the test points is given by Eq.~\eqref{eq: conditional predictive posterior with unknown prevalence selection}:
\begin{equation}
p(y_{n,\ast}|s_\ast\eq 1, x_{n,\ast}, X, Y)=\int q_n(y_{n,\ast}|w) p(w|X,Y)dw\, ,
\label{eq: conditional predictive posterior with unknown prevalence selection}
\end{equation}
the direct counterpart of Eq.~\eqref{eq: conditional predictive posterior with selection}. In practice one relies on samples $w_k$ from $p(w|X,Y)$ both during iterations of Eq.~\eqref{eq: VBI unknown test prevalence - y},~\eqref{eq: VBI unknown test prevalence - pi} and to integrate Eq.~\eqref{eq: conditional predictive posterior with unknown prevalence selection}.

Regardless of its computational simplicity, the joint inference of test labels and of the hidden test-time prevalence remains rather ill-posed. 
The choice of prior on $\bm{\pi}$ (the choice of $\bm{\alpha}_0$) can strongly impact the fixed point reached by the iterative procedure. There is no definitive solution (for instance, the flat prior choice $\alpha_{y,0}=1$ is not necessarily effective in our experience) other than to use all available prior knowledge. We make a few remarks as guidelines below. 

If one assumes the test sample to be drawn from the true population, the optimal prediction is $p(y_{n,\ast}|x_{n,\ast},w)$, averaged w.r.t.~$w$ over the posterior $p(w|X,Y)$. If one instead assumes the distribution of the test sample to be subject to prevalence bias, with label probabilities $\pi_y\coloneqq 1/|\calY|$ known to be equiprobable (balanced), the optimal rule is ${p(y_{n,\ast}|x_{n,\ast},w)}/{p(y_{n,\ast}|w)}$ averaged over $p(w|X,Y)$. This coincides in the limit of $\alpha_{y,0}=\alpha_0\rightarrow +\infty$ with the result of the above VBI procedure. This choice of $\alpha$ encodes the following beliefs: there is prevalence bias; the data is insufficient to also determine the hidden prevalence; there is no compelling reason to assume a class was favoured above any other in the sampling. 

We have found encouraging results with the following heuristic. The value of $\bm{\alpha}_0$ is set to encode a preference for the resulting rule $q_n(y_{n,\ast}|w)$ to coincide, early in the iterative process, with the optimal prediction for the true population, $p(y_{n,\ast}|x_{n,\ast},w)$. Starting from this ``safe bet'', we wish to only update this prediction given sufficient incentive. 
Given the standard interpretation of $\alpha_{y,0}$ as a ``virtual'' observation for the label $y$, we suggest: $\alpha_{y,0}\propto p(y|w)$ such that $\text{min}_y (\alpha_{y,0}) = k$ for a small value $k$ (\eg, $k\coloneqq1$). Notice that in the limit of $k\rightarrow +\infty$, the resulting prior on $\bm{\pi}$ collapses to a pointwise mass at $\pi_y=p(y|w)$ so that $q_n(y_{n,\ast})=p(y_{n,\ast}|x_{n,\ast},w)$. For smaller values of $k$ the expectation of $\pi_y$ w.r.t.~$p(\bm{\pi})$ is $p(y|w)$ but the spread of the distribution increases, allowing for non trivial iterations Eq.~\eqref{eq: VBI unknown test prevalence - y},~\eqref{eq: VBI unknown test prevalence - pi}. 

\end{document}